\documentclass[10pt]{article} 

\usepackage[utf8]{inputenc}
\usepackage{amsmath}
\usepackage{amsfonts}
\usepackage{amssymb}
\usepackage{mathtools}
\usepackage{listings}
\usepackage{mathrsfs}
\usepackage{times}
\usepackage{float}
\usepackage{color}
\usepackage{setspace}
\usepackage{color,soul}

\usepackage{amsmath,amsfonts,amsthm,eucal}
\usepackage{enumerate}
\usepackage[letterpaper,margin=3cm]{geometry}
\usepackage{float}
\usepackage[title]{appendix}
\usepackage{color}
\usepackage{tabularx}
\usepackage{siunitx}
\usepackage[tableposition=below]{caption}
\usepackage{array}

\usepackage[
pdfstartview=XYZ,
bookmarks=true,
colorlinks=true,
linkcolor=blue,
urlcolor=blue,
citecolor=blue,
pdftex,
bookmarks=true,
linktocpage=true,   
hyperindex=true
]{hyperref}

\usepackage{lipsum}
\usepackage{lineno}

\usepackage[FIGBOTCAP,TABTOPCAP,bf,tight]{subfigure}

\usepackage{natbib}
\setcitestyle{authoryear}

\usepackage[pdftex]{graphicx}
\usepackage{epstopdf}
\usepackage{booktabs} 
\usepackage{tikz,mathpazo}
\usetikzlibrary{shapes.geometric, arrows}
\usepackage{caption}

\newcommand{\tensor}[1]{\ensuremath{\boldsymbol{#1}}}

\DeclareMathOperator{\Diver}{\nabla^{\tensor X}\cdot}

\DeclareMathOperator*{\argmin}{argmin}

\usepackage{algorithm}
\usepackage{algorithmicx}
\usepackage[noend]{algpseudocode}

\theoremstyle{remark}

\newtheorem{remark}{Remark}

\renewcommand{\vec}[1]{\ensuremath{\boldsymbol{#1}}}
\theoremstyle{definition}
\newtheorem{definition}{Definition}

\usepackage{algorithm}
\usepackage[noend]{algpseudocode}
\newcolumntype{M}[1]{>{\centering\arraybackslash}m{#1}}

\title{Geometric deep learning for computational mechanics Part I:  Anisotropic Hyperelasticity} 

\begin{document}

\author{Nikolaos N. Vlassis\thanks{Department of Civil Engineering and Engineering Mechanics, 
 Columbia University, 
 New York, NY 10027.     \textit{nnv2102@columbia.edu}  }       \and
Ran Ma\thanks{Department of Civil Engineering and Engineering Mechanics, 
 Columbia University, 
 New York, NY 10027.     \textit{rm3681@columbia.edu}    
}
\and
        WaiChing Sun\thanks{Department of Civil Engineering and Engineering Mechanics, 
 Columbia University, 
 New York, NY 10027.
  \textit{wsun@columbia.edu}  (corresponding author)   }
}

\maketitle

\begin{abstract}
This paper is the first attempt to use geometric deep learning and  Sobolev training to incorporate non-Euclidean microstructural data such that anisotropic hyperelastic material machine learning models can be trained 
in the finite deformation range. 
While traditional hyperelasticity models often incorporates homogenized measures of microstructural attributes, 
such as porosity averaged orientation of constitutes, these measures cannot reflect the
topological structures of the attributes. We fill this knowledge gap by introducing the concept of 
weighted graph as a new mean to store topological information, such as 
the connectivity of anisotropic grains in an assembles. Then, by 
leveraging a graph convolutional deep neural network architecture in the spectral domain, 
w introduce a mechanism to 
incorporate these 
non-Euclidean weighted graph data directly as input for training and for predicting the elastic responses of materials with complex microstructures. 
To ensure smoothness and prevent non-convexity of the trained stored energy functional, we 
we introduce a Sobolev training technique for neural networks such that stress measure is obtained implicitly 
from taking directional derivatives of the trained energy functional. 
By optimizing the neural network to approximate both the energy functional output and the stress measure, 
we introduce a training procedure the improves
the efficiency and the generalize the learned energy functional for different micro-structures. 
The trained hybrid neural network model is then used to generate new stored energy functional for unseen microstructures in a parametric study to predict the influence of elastic anisotropy on the nucleation and propagation of fracture in the brittle regime.
\end{abstract}

\section{Introduction}
\label{intro}

Conventional constitutive modeling efforts often rely on human interpretation of geometric descriptors of microstructures. These descriptors, such as volume fraction of void, dislocation density, degradation function, slip system orientation and shape factor are often incorporated as state variables in a system of ordinary differential equations 
that leads to the constitutive responses at a material point. 
Classical examples include the family of Gurson models in which
volume fraction of void is related to ductile fracture
 \citep*{gurson_continuum_1977,needleman_continuum_1987,zhang_complete_2000,nahshon_modification_2008,nielsen_ductile_2010},  critical state plasticity in which porosity and over-consolidation ratio dictates the plastic dilatancy and hardening law \citep*{schofield_critical_1968, borja_cam-clay_1990, manzari_critical_1997, sun_unified_2013,  liu_determining_2016, wang_identifying_2016} and crystal plasticity where activation of slip system leads to plastic deformation 
\citep*{anand_computational_1996, na_computational_2018, ma_investigating_2018}. 
In those cases, a specific subset of descriptors are often incorporated manually such that 
the most crucial deformation mechanisms for the stress-strain relationships 
are described mathematically. 

While this approach has achieved a level of success, especially for isotropic materials, materials of complex microstructures often requires more complex geometric and topological descriptors to sufficiently describe the geometrical features \citep*{jerphagnon_description_1978, sun_multiscale_2014, kuhn_stress-induced_2015}. 
The human interpretation limits the complexity of the state variables and may lead to lost opportunity of utilizing all the available information for the microstucture, which could in turn reduce the prediction quality.  A data-driven approach should be considered to discover constitutive law mechanisms when human interpretation capabilities become restrictive \citep*{kirchdoerfer2016data,eggersmann2019model,he2019physics,stoffel2019stability,bessa2017framework,liu2018data}. 
In this work, we consider the general form of a strain energy functional that reads, 
\begin{equation}
\label{eq:form} \psi = \psi(\tensor{F}, \mathbb{G}) \: \: , \: \:  \tensor{P} = \frac{\partial \psi}{\partial \tensor{F}} ,
\end{equation}
where $\mathbb{G}$ is a graph that stores the non-Euclidean data of the microstructures (e.g. crystal connectivity, grain connectivity). Specifically, we attempt to train a neural network approximator of the anisotropic stored elastic energy functional across different polycrystals with the sole extra input to describe the anisotropy being the weighted crystal connectivity graph. 

\begin{figure}[h]
\centering
\includegraphics[width=14.0cm,angle=0]{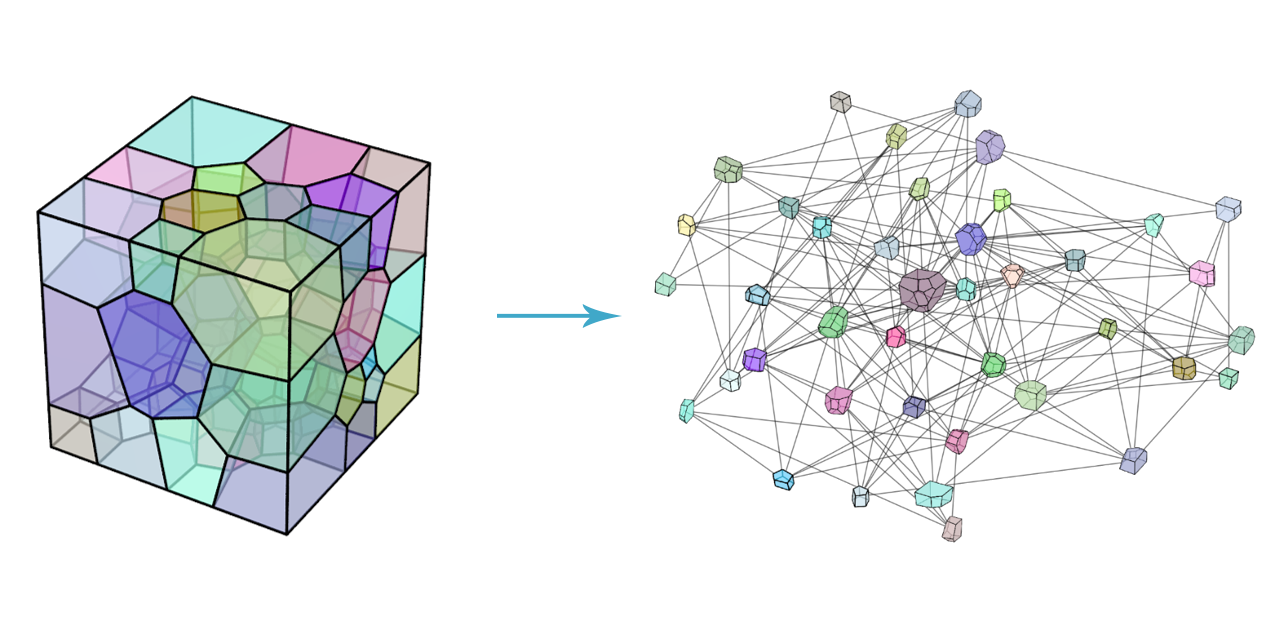}
\caption{Polycrystal interpreted as a weighted connectivity graph. The graph is undirected and weighted at the nodes.} 
\label{fig:weighted_graph}
\end{figure}

It can be difficult to directly incorporate either Euclidean or non-Euclidean data to a hand-crafted constitutive model. There have been attempts to infer information directly from scanned microstructual images using neural networks that utilize a convolutional layer architecture (CNN) \citep{lubbers_inferring_2017}. The endeavor to distill physically meaningful and interpertable features from scanned microstructural images 
stored in a Euclidean grid can be a complex and sometimes futile process.
While recent advancements in convolutional neural networks have 
provided an effective mean to extract features that lead to extraordinary superhuman performance for image classification tasks \citep{krizhevsky_imagenet_2012}, similar success has not been recorded for mechanics predictions. 
This technical barrier could be attributed to the fact that feature vectors obtained from voxelized image data are highly sensitive to the grid resolution and noise. The robustness and accuracy also exhibit strong dependence on the number of dimensions of the feature vector space and the algorithms that extract the low-dimensional representations. In some cases, over-fitting and under-fitting can both cause the trained CNN extremely vulnerable to adversarial attacks and hence not suitable for high-risk, high-regret applications. 

 As demonstrated by \citep*{frankel_predicting_2019,jones2018machine}, using images directly as additional input to our polycrystal energy functional approximator would be heavily contingent to the quality and size of the training pool. A large number of images, possibly in three dimensions, and in high enough resolution would be necessary to represent the latent features that will aid the approximator to distinguish successfully between different polycrystals. Using data in a Euclidean grid is an esoteric process that is dependent on empirical evidence that the current training sample holds adequate information to infer features useful in formulating a constitutive law. However, gathering that evidence can be a laborious process as it requires numerous trial and error runs and is weighed down by the heavy computational costs of performing filtering on Euclidean data (e.g. on high resolution 3D image voxels).

Graph representation of the data structures can provide a momentous head-start to overcome this very impediment. An example is the connectivity graph used in granular mechanics community where 
the formations and evolution of force chains are linked to macroscopic phenomena, such as shear band 
formation and failures \citep{satake1992discrete, kuhn_stress-induced_2015, sun2013multiscale, tordesillas2014micromechanics, wang2019meta, wang2019updated}.
The distinct advantage of the graph representation of data, as showcased in the previous granular mechanics studies, is 
the high-level interpretability of the data structures. 
A knowledgeable user can employ domain expertise to craft graph structures that carry crucial relational information to solve the problem at hand. 
Designing graph structures - in terms of node connectivity, node and edge weights - 
can be highly expressive and exceptionally tailored to the task at hand. 
At the same time, by concisely selecting appropriate graph weights, 
one may incorporate only the essences of micro-structural data critical for 
mechanics predictions and 
hence more interpretable, flexible, economical and efficient than 
than incorporating feature spaces inferred from 3D voxel images.
Furthermore, since one may easily rotational and transitional invariant data as weights, 
the graph approach is also advantageous for predicting constitutive responses that require frame indifference.

Currently, machine learning applications often employs two families of algorithms to take graphs as inputs, 
i.e., representation learning algorithms and graph neural networks. 
The former usually refer to unsupervised methods that convert graph data structures into formats or features that are easily comprehensible by machine learning algorithms \citep{bengio2013representation}.
The later refer to neural network algorithms that accept graphs as inputs with layer formulations that can operate directly on graph structures \citep{scarselli2008graph}. 
Representation learning on graphs shares concepts with rather popular embedding techniques on text and speech recognition \citep{mikolov_distributed_2013} to encode input in a vector format that can be utilized by common regression and classification algorithms. There has been multiple studies on encoding graph structures, spanning from the level of nodes \citep{grover_node2vec:_2016} up to the level of entire graphs \citep*{perozzi_deepwalk:_2014,narayanan_graph2vec:_2017}. Graph embedding algorithms, like DeepWalk \citep{perozzi_deepwalk:_2014}, utilize techniques such as random walks to "read" sequences of neighbouring nodes resembling reading word sequences in a sentence and encode those graph data in an unsupervised fashion. 

While these algorithms have been proven to be rather powerful and demonstrate competitive results in tasks like classification problems, they do come with disadvantages that can be limiting for use in engineering problems. Graph representation algorithms work very well on encoding the training dataset. However, 
they could be difficult to generalize and cannot accommodate dynamic data structures. 
This can be proven problematic for mechanics problems 
, where we expect a model to be as generalized as much as possible in terms of material structure variations (e.g. polycrystals, granular assemblies). Furthermore, representation learning algorithms can be difficult to  combine with another neural network architecture for a supervised learning task in a sequential manner. 
In particular, when the representation learning is performed separately and independently from
the supervised learning task that generates the 
 the energy functional approximation, there is no guarantee that 
the clustering or classifications obtained from the representative learning is physically meaningful.
Hence, the representation learning may not be capable of generating features that facilitates the energy functional prediction task in a completely unsupervised setting. 

For the above reasons, we have opted for a hybrid neural network architecture that combines an unsupervised graph convolutional neural network with a multilayer perceptron to perform the regression task of predicting an energy functional. Both branches of our suggested hybrid architecture learn simultaneously from the same back-propagation process with a common loss function tailored to the approximated function. The graph encoder part - borrowing its name from the popular autoencoder architecture \citep{vincent_extracting_2008} - learns and adjusts its weights to encode input graphs in a manner that serves the approximation task at hand. Thus, it does eliminate the obstacle of trying to coordinate the asynchronous steps of graph embedding and approximator training by parallel fitting both the graph encoder and the energy functional approximator with a common training goal (loss function).

As for notations and symbols in this current work, bold-faced letters
denote tensors (including vectors which are rank-one tensors); 
the symbol '$\cdot$' denotes a single contraction of adjacent indices of two tensors (e.g. $\vec{a} \cdot \vec{b} = a_{i}b_{i}$ or $\tensor{c}
\cdot \vec{d} = c_{ij}d_{jk}$ ); the symbol `:' denotes a double
contraction of adjacent indices of tensor of rank two or higher (
e.g. $\tensor{C} : \vec{\epsilon^{e}}$ = $C_{ijkl} \epsilon_{kl}^{e}$
); the symbol `$\otimes$' denotes a juxtaposition of two vectors
(e.g. $\vec{a} \otimes \vec{b} = a_{i}b_{j}$) or two symmetric second
order tensors (e.g. $(\vec{\alpha} \otimes \vec{\beta})_{ijkl} =
\alpha_{ij}\beta_{kl}$). Moreover, $(\tensor{\alpha}\oplus\tensor{\beta})_{ijkl} = \alpha_{jl} \beta_{ik}$ and $(\tensor{\alpha}\ominus\tensor{\beta})_{ijkl} = \alpha_{il} \beta_{jk}$. We also define identity tensors $(\tensor{I})_{ij} = \delta_{ij}$, $(\tensor{I}^4)_{ijkl} = \delta_{ik}\delta_{jl}$, and $(\tensor{I}^4_{\text{sym}})_{ijkl} = \frac{1}{2} (\delta_{ik}\delta_{jl} + \delta_{il}\delta_{kj})$, where $\delta_{ij}$ is the Kronecker delta. As for sign conventions, unless specified otherwise,
we consider the direction of the tensile stress and dilative pressure as positive.

\section{Graphs as non-Euclidean descriptors for micro-structures}

This section provides a detailed account on how to incorporate microstructural data represented by 
weighted graphs as descriptors for constitutive modeling. 
To aid readers not familiar with graph theory, 
we provide a brief review on some basic concepts of graph theory essential for understanding this research.
The essential terminologies and definitions required to construct 
the graph descriptors can be found in  Section \ref{sec:graph_theory}. Following this review, 
we establish a method to translate the topological information of microstructures into various types of graphs (Section \ref{sec:polyformation}) and explain the properties of these graphs that are critical for the 
constitutive modeling tasks (Section \ref{sec:deeplearning}). 

\subsection{Graph theory terminologies and definitions}
\label{sec:graph_theory}
  
In this section, a brief review of several terms of graph theory is provided to facilitate the illustration of the concepts in this current work. More elaborate descriptions can be found in \citep*{graham1989concrete,west2001introduction,bang2008digraphs}: \\

\begin{definition}
A \textbf{graph} is a two-tuple $\mathbb{G} = (\mathbb{V,E})$ where $\mathbb{V} = \{v_1,...,v_N\}$ is a non-empty \textbf{vertex set} (also referred to as nodes) and $\mathbb{E} \subseteq \mathbb{V} \times \mathbb{V}$ is an \textbf{edge set}. To define a graph, there exists a relation that associates each edge with two vertices (not necessarily distinct). These two vertices are called the edge's \textbf{endpoints}. The pair of endpoints can either be unordered or ordered. \label{def:graph} 
\end{definition}

\begin{definition}
 An \textbf{undirected graph} is a graph whose edge set $\mathbb{E} \subseteq \mathbb{V} \times \mathbb{V}$ connects \textit{unordered} pairs of vertices together. 
 \label{def:undirectedgraph}
 \end{definition}

\begin{definition}
A \textbf{directed graph} is a graph whose edge set $\mathbb{E} \subseteq \mathbb{V} \times \mathbb{V}$ connects \textit{ordered}  pairs of vertices together.
 \label{def:directedgraph}
  \end{definition}

\begin{definition}
A \textbf{loop} is an edge whose endpoint vertices are the same. When the all the nodes in the graph are in a loop with themselves, the graph is referred to as allowing self-loops. 
 \label{def:loop}
 \end{definition}
 
\begin{figure}[h!]
\centering
\begin{tabular}{cccc}
\includegraphics[width=3.5cm,angle=-0]{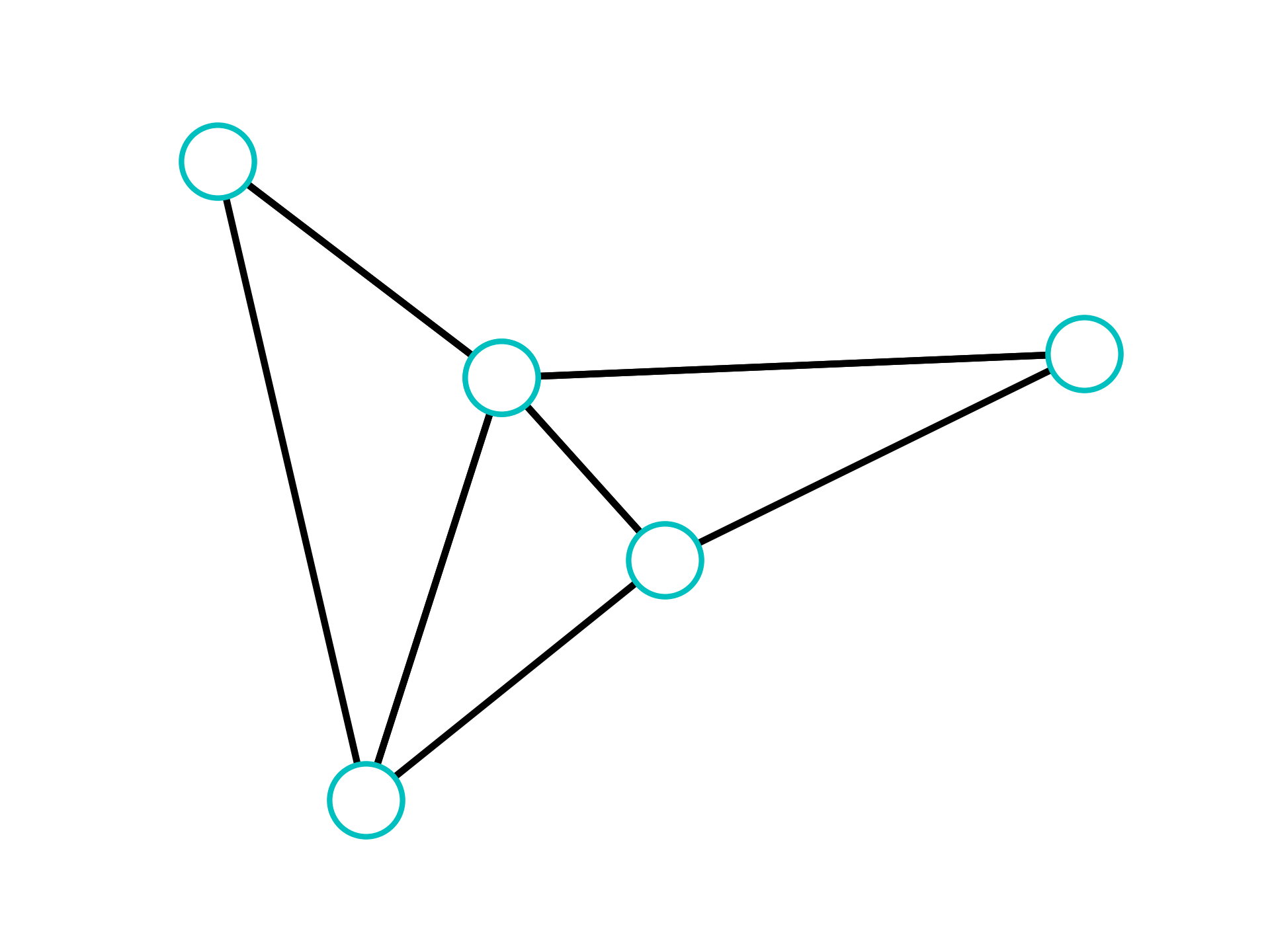}  &
\includegraphics[width=3.5cm ,angle=-0]{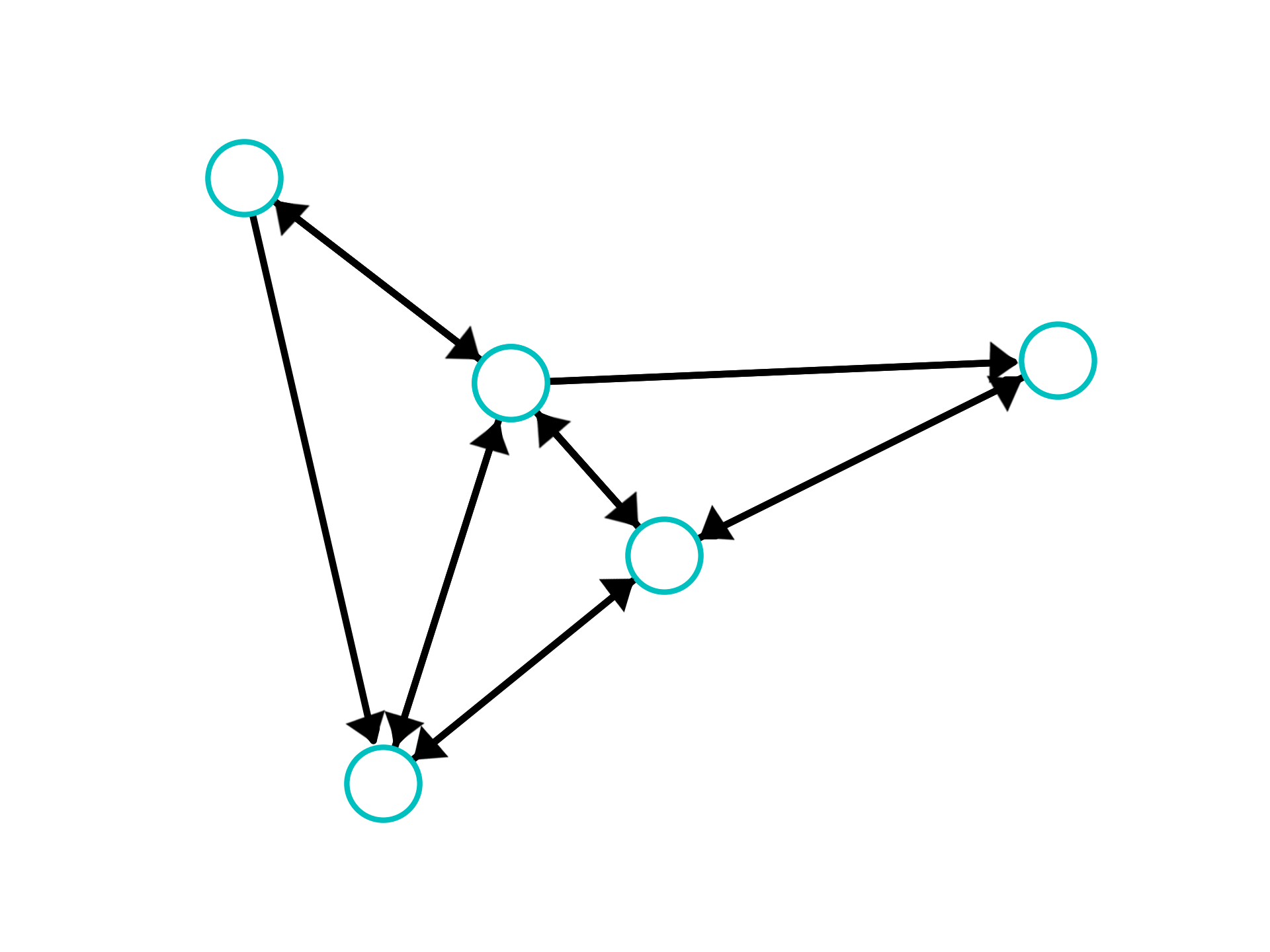}&
\includegraphics[width=3.5cm,angle=-0]{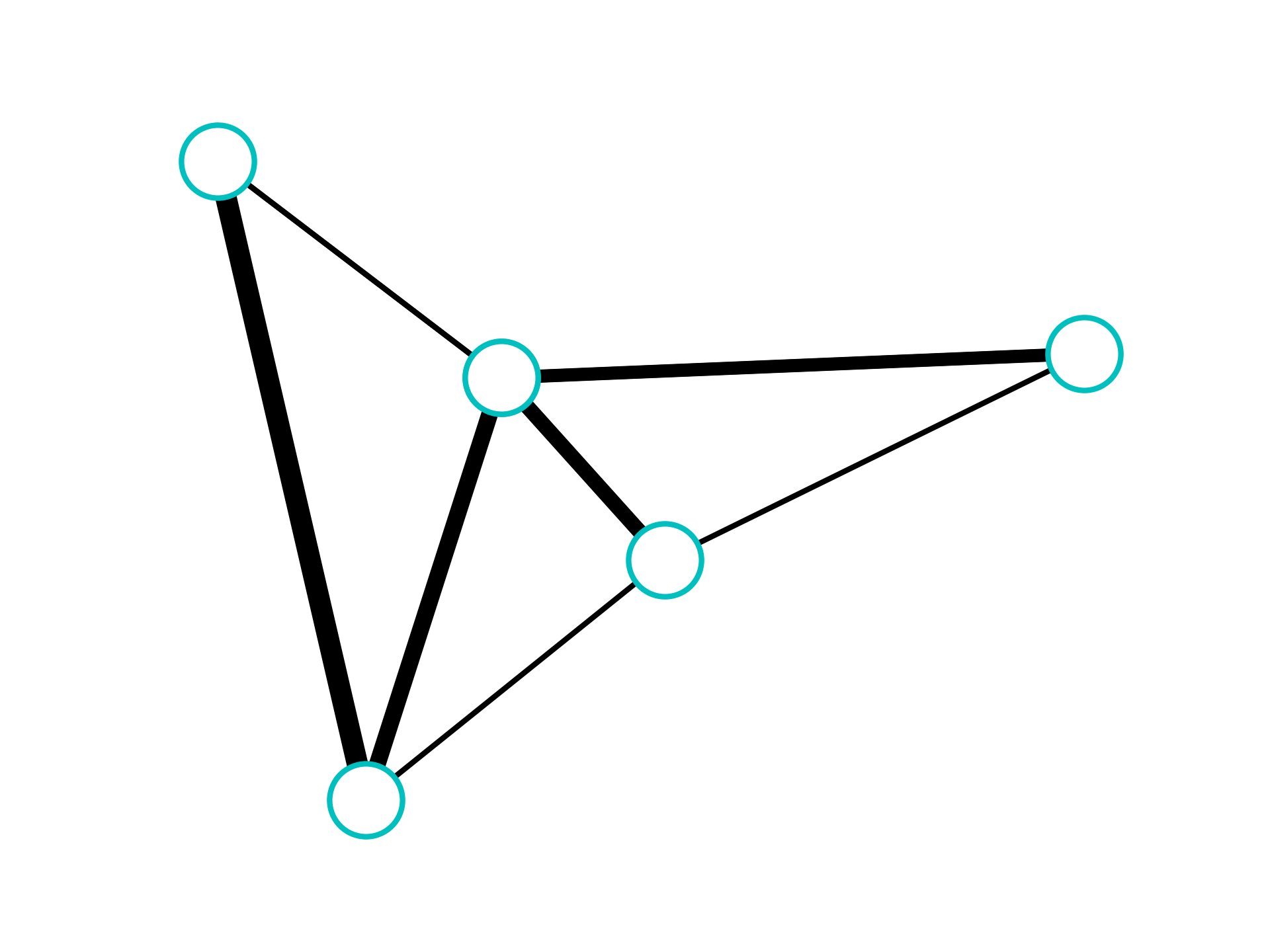}  &
\includegraphics[width=3.5cm,angle=0]{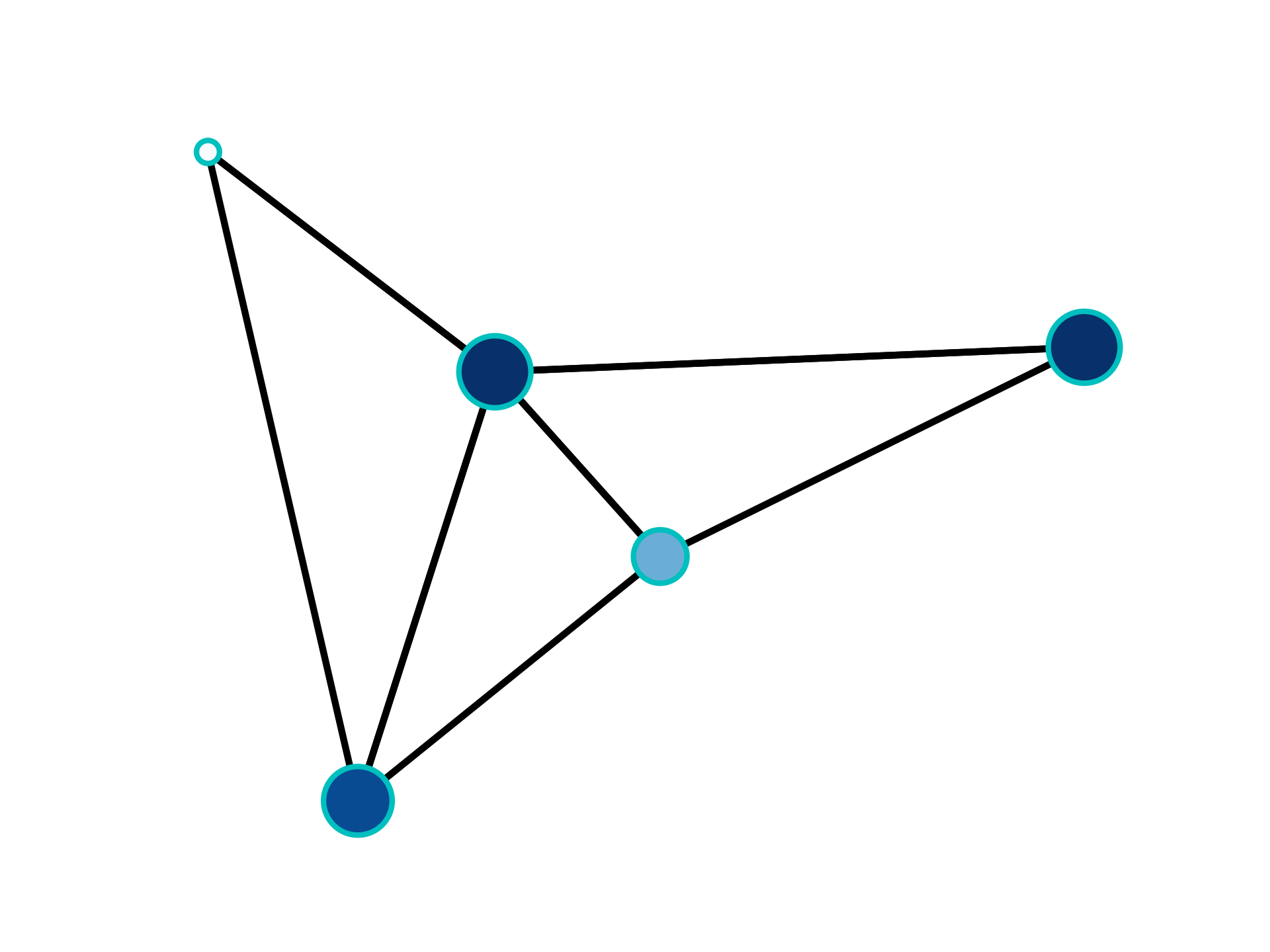}  \\
(a) & (b) & (c) & (d)
\end{tabular}{}

\caption{Different types of graphs. (a) Undirected (simple) binary graph (b) Directed binary graph (c) Edge-weighted undirected graph (d) Node-weighted undirected graph.}
\label{fig:graph_types}
\end{figure}

\begin{definition}
\textbf{Multiple edges} are edges having the same pair of endpoint vertices. 
 \label{def:multiple_edges}
 \end{definition}
 
\begin{definition}
A \textbf{simple graph} is a graph that does not have loops or multiple edges.
 \label{def:simple_graph}
 \end{definition}

\begin{definition}
Two vertices that are connected by an edge are referred to as \textbf{adjacent} or as \textbf{neighbors}. 
 \label{def:adjacent}
 \end{definition}
 
 \begin{definition}
The term \textbf{weighted graph} traditionally refers to graph that consists of edges that associate with edge-weight function $w_{ij}: \mathbb{\mathbb{E}} \rightarrow \mathbb{R}^{n} \enskip \text{with} \enskip (i,j) \in \mathbb{E}$ that maps all edges in $\mathbb{E}$ onto a set of real numbers. $n$ is the total number of edge weights and each set of edge weights can be represented by a matrix $\tensor{W}$ with components $w_{ij}$.
 \label{def:weighted_graph}
 \end{definition}

In this current work, unless otherwise stated, we will be referring to weighted graphs as graphs weighted at the vertices - each node carries information as a set of weights that quantify features of microstructures. All vertices are associated with a vertex-weight function $f_{v}: \mathbb{\mathbb{V}} \rightarrow \mathbb{R}^D  \enskip \text{with} \enskip v \in \mathbb{V}$ that maps all vertices in $\mathbb{V}$ onto a set of real numbers, where $D$ is the number of weights - features. The node weights can be represented by a $N \times D$ matrix $\tensor{X}$ with components $x_{ik}$, where the index $i \in [1,...,N]$ represents the node and the index  $k \in [1,...,D]$ represents the type of node weight - feature.\\

\begin{definition}
A graph whose edges are unweighted ($w_{\epsilon} = 1 \enskip \forall \epsilon \in \mathbb{E}$) can be called a \textbf{binary graph}. 
 \label{def:binary_graoh}
 \end{definition}
 
To facilitate the description of graph structures, several terms for representing graphs are introduced: \\

\begin{definition}
The \textbf{adjacency matrix} $\tensor{A}$ of a graph $\mathbb{G}$ is the $N \times N$ matrix in which entry $\alpha_{ij}$ is the number of edges in $\mathbb{G}$ with endpoints $\{v_i , v_j\}$. 
 \label{def:adjacency_matrix}
 \end{definition}

\begin{definition}
If the vertex $v$ is an endpoint of edge $\epsilon$, then $v$ and $\epsilon$ are \textbf{incident}. The \textbf{degree} $d$ of a vertex $v$ is the number of incident edges. The \textbf{degree matrix} $\tensor{D}$ of a graph $\mathbb{G}$ is the $N \times N$ diagonal matrix with diagonal entries $d_{ii}$ equal to the degree of vertex $v_i$.
 \label{def:degree_matrix}
 \end{definition}

\begin{definition}
The \textbf{unnormalized Laplacian operator} $\tensor{\Delta}$ is defined such that:

\begin{align}
(\tensor{\Delta}f)_i &= \sum_{j:(i,j)\in \mathbb{E}} w_{ij} (f_i - f_j) \\
& = f_i \sum_{j:(i,j)\in \mathbb{E}} w_{ij} - \sum_{j:(i,j)\in \mathbb{E}} w_{ij} f_j .
\end{align}
\label{eq:laplacian_operator}

By writing the equation above in matrix form, the unnormalized Laplacian matrix $\tensor{\Delta}$ of a graph $\mathbb{G}$ is the $N \times N$ positive semi-definite matrix defined as $\tensor{\Delta} = \tensor{D} - \tensor{W}$. 

\begin{figure}[H]
\centering
\includegraphics[width=6.0cm,angle=-0]{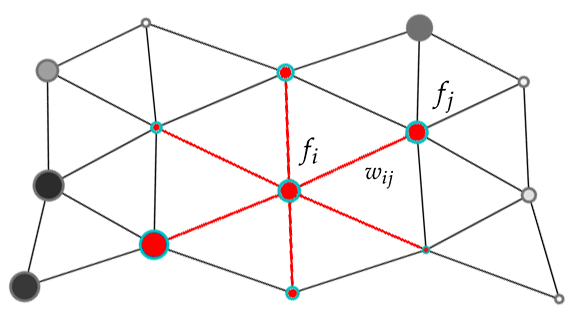}
\caption{The graph Laplacian operator $\tensor{\Delta}$ describes the difference between a value $f$ at a node and its local average.} 
\label{fig:lapl_example_graph}
\end{figure}

In this current work, binary graphs will be used, thus, the equivalent expression is used for the unnomrmalized Laplacian matrix $\tensor{L}$, defined as $\tensor{L} = \tensor{D} - \tensor{A}$ with the entries $l_{ij}$ calculated as:

\begin{equation}
\label{eq:laplacian_terms}
\l_{ij}= \left\{
\begin{array}{ll}
      d_i, & i=j \\
       -1, & i \neq j \enskip \text{and} \enskip v_i \enskip \text{is adjacent to} \enskip v_j \\
       0, &  \text{otherwise}. \\
\end{array} 
\right. 
\end{equation}
 \label{def:laplacian}
 \end{definition}

\begin{definition}
For binary graphs, the \textbf{symmetric nomrmalized Laplacian matrix} $\tensor{L^{\text{sym}}}$ of a graph $\mathbb{G}$ is the $N \times N$ matrix defined as:

\begin{equation}
\tensor{L^{\text{sym}}} = \tensor{D}^{-\frac{1}{2}} \tensor{L}\tensor{D}^{-\frac{1}{2}} = \tensor{I} - \tensor{D}^{-\frac{1}{2}} \tensor{A}\tensor{D}^{-\frac{1}{2}}.
\label{eq:graph_laplacian}
\end{equation} 

The entries $l^{\text{sym}}_{ij}$ of the matrix $\tensor{L^{\text{sym}}}$ can also be calculated as:

\begin{equation}
\label{eq:norm_laplacian_terms}
\l^{\text{sym}}_{ij}= \left\{
\begin{array}{ll}
      1, & i=j \enskip \text{and} \enskip d_i \neq 0\\
       - (d_i d_j)^{-\frac{1}{2}}, & i \neq j \enskip \text{and} \enskip v_i \enskip \text{is adjacent to} \enskip v_j \\
       0, &  \text{otherwise}. \\
\end{array} 
\right. 
\end{equation}

 \label{def:normalized_laplacian}
 \end{definition}
 
\subsection{Polycrystals represented as node-weighted undirected graphs}
\label{sec:polyformation}

Representing microstructural data as weight graphs requires pooling, a down-sampling procedure to converts 
field data of a specified domain into low-dimensional features that preserve the important information. 
One of the most intuitive pooling is to infer the grain connectivity graph from an micro-CT image \citep{jaquet2013estimation, wang2016identifying} or realization of micro-structures generated from software packages such as Neper or Cubit \citep{quey_large-scale_2011, salinger2016albany}. 
In this work, we treat each individual crystal grain as as node or vertex in a graph, and create an edge for each 
in-contact grain pair. The sets of the nodes and edges, $\mathbb{B}$ and $\mathbb{E}$ collectively forms as a graph (cf. Def. \ref{def:graph}). Without adding any weight, this graph 
can be represented by a binary graph (cf. Def. \ref{def:binary_graoh}) of which the binary weight for each edge indicates whether the two grains are in contact, as shown in Figure \ref{fig:graph_types}.
While the unweighted graph can be used incorporated into the machine learning process, 
additional information of the microstructures can be represented by weights assigned on the nodes and edges of a graph that represents an assembles. In this current work, the database included information on the volume, the orientation (in Euler angles), the total surface area, the number of faces, the numbers of neighbors, as well as other shape descriptors (convexity, equivalent diameter, etc) for every crystal in the polycrystals - all of which could be assigned as node weights in the connectivity graph. Information was also available on the nature of contact between grains - such as the surface and the angle of contact - which could be used as weights for the edges of the graph. While this current work is solely focused on node weighted graphs, future work could employ algorithms that utilize edge weights as well to generate more robust microstructure predictors.

\begin{figure}[h!]
\centering
\begin{tabular}{M{6cm}M{6cm}}
\includegraphics[width=5cm,angle=0]{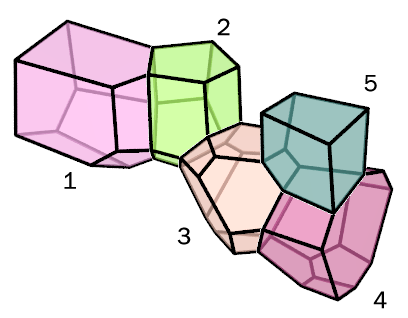} &   
\includegraphics[width=6 cm ,angle=0]{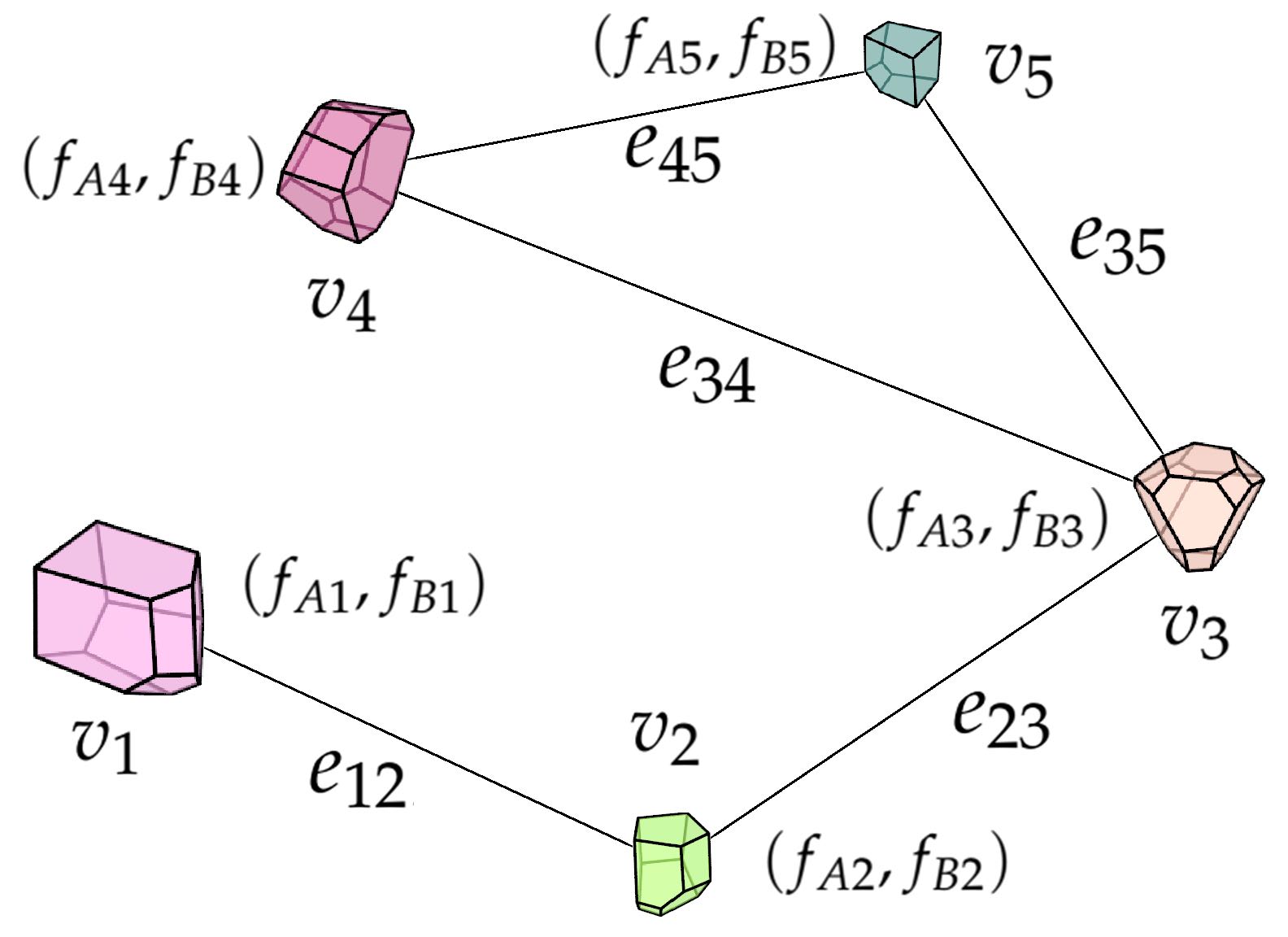} \\
(a) & (b)  \\[6pt]

\end{tabular}

\caption{A sub-sample of a polycrystal (a) represented as an undirected weighted graph (b). If two crystals in the formation share an edge, their nodes are also connected in the graph. Each node is weighted by two features $f_{A}$ and $f_{B}$.}
\label{fig:sample_graph}

\end{figure}

To demonstrate how graphs used to represent a polycrystalline assembles are generated, we introduced a simple example where an assembly consist of  5 crystals shown in Fig.~\ref{fig:sample_graph}(a) is converted into a node-weighted graph. 
 Each node of the graph represents a crystal. An edge is defined between two nodes if they are connected - share a surface. The graph is undirected meaning that there is no direction specified for the edges. The vertex set $\mathbb{V}$ and edge set $\mathbb{E}$ for this specific graph are $\mathbb{V} = \{v_1,v_2,v_3,v_4,v_5\}$ and $\mathbb{E} = \{e_{12},e_{23},e_{34},e_{35},e_{45}\}$ respectively.

An undirected graph can be represented by an adjacency matrix $\tensor{A}$ (cf. Def. \ref{def:adjacency_matrix}) that holds information for the connectivity of the nodes. The entries of the adjacency matrix $\tensor{A}$ in this case are binary - each entry of the matrix is 0 if an edge does not exist between two nodes and 1 if it does. Thus, for the example in Fig.~\ref{fig:sample_graph}, crystals 1 and 2 are connected so the entries $(1,2)$ and $(2,1)$ of the matrix $\tensor{A}$ would be 1, while crystals 1 and 3 are not so the entries $(1,3)$ and $(3,1)$ will be 0 and so on. If the graph allows self-loops, then the entries in the diagonal of the matrix are equal to 1 and the adjacency matrix with self-loops is defined as $\tensor{\hat{A}} = \tensor{A} + \tensor{I}$. The complete symmetric matrices $\tensor{A}$ and $\tensor{\hat{A}}$ for this example will be: 

\begin{minipage}{.45\linewidth}
\[\tensor{A}= \left[ \begin{array}{ccccc}
0 & 1 & 0 & 0 & 0 \\
1 & 0 & 1 & 0 & 0 \\
0 & 1 & 0 & 1 & 1 \\
0 & 0 & 1 & 0 & 1 \\
0 & 0 & 1 & 1 & 0 \end{array} \right]\] 
\end{minipage}
\begin{minipage}{.45\linewidth}

\[\tensor{\hat{A}} = \tensor{A} + \tensor{I}= \left[ \begin{array}{ccccc}
1 & 1 & 0 & 0 & 0 \\
1 & 1 & 1 & 0 & 0 \\
0 & 1 & 1 & 1 & 1 \\
0 & 0 & 1 & 1 & 1 \\
0 & 0 & 1 & 1 & 1 \end{array} \right]\]

\end{minipage}

\medskip

A diagonal degree matrix $\tensor{D}$ can also useful to describe a graph representation. The degree matrix $\tensor{D}$ only has diagonal terms that equal the number of neighbors of the node represented in that row. The diagonal terms can simply be calculated by summing all the entries in each row of the adjacency matrix. It is noted that, when self-loops are allowed, a node is a neighbor of itself, thus it must be added to the number of total neighbors for each node. The degree matrix $\tensor{D}$ for the example graph in Fig.~\ref{fig:sample_graph} would be:

\[\tensor{D}= \left[ \begin{array}{ccccc}
1 & 0 & 0 & 0 & 0 \\
0 &2 & 0 & 0 & 0 \\
0 & 0 & 3 & 0 & 0 \\
0 & 0 & 0 & 2 & 0 \\
0 & 0 & 0 & 0 & 2 \end{array} \right]\]

The polycrystal connectivity graph can be represented by its graph Laplacian matrix $\tensor{L}$ - defined as $\tensor{L} = \tensor{D} - \tensor{A}$, as well as the normalized symmetric graph Laplacian matrix $\tensor{L^{\text{sym}}} =\tensor{D}^{-\frac{1}{2}} \tensor{L}\tensor{D}^{-\frac{1}{2}}$. The two matrices for the example of Fig.~\ref{fig:sample_graph} are calculated below:

\begin{minipage}[t]{.45\linewidth}
\vspace*{0.65cm}
\[\tensor{L}= \left[ \begin{array}{ccccc}
1 & -1 & 0 & 0 & 0 \\
-1 & 2 & -1 & 0 & 0 \\
0 & -1 & 3 & -1 & -1 \\
0 & 0 & -1 & 2 & -1 \\
0 & 0 & -1 & -1 & 2 \end{array} \right]\] 
\end{minipage}
\begin{minipage}[t]{.45\linewidth}
\[\tensor{L^{\text{sym}}} =  \left[ \begin{array}{ccccc}
1 & -\frac{\sqrt{2}}{2} & 0 & 0 & 0 \\
-\frac{\sqrt{2}}{2} & 1 & -\frac{\sqrt{6}}{6} & 0 & 0 \\
0 & -\frac{\sqrt{6}}{6} & 1 & -\frac{\sqrt{6}}{6} & -\frac{\sqrt{6}}{6} \\
0 & 0 & -\frac{\sqrt{6}}{6} & 1 & -\frac{1}{2} \\
0 & 0 & -\frac{\sqrt{6}}{6} & -\frac{1}{2}  & 1 \end{array} \right]\]
\end{minipage}
\medskip

Assume that, for the example in Fig.~\ref{fig:sample_graph},there is information available for two features $A$ and $B$ for each crystal in the graph that will be used as node weights - this could be the volume of each crystal, the orientations and so on. The node weights for each feature can be described as a vector,  $\tensor{f}_A = (f_{A1},f_{A2},f_{A3},f_{A4},f_{A5})$ and $\tensor{f}_B = (f_{B1},f_{B2},f_{B3},f_{B4},f_{B5})$, such that each component of the vector corresponds to a feature of a node. The node features can all be represented in a feature matrix $\tensor{X}$ where each row corresponds to a node and each column corresponds to a feature. For the example in question, the feature matrix would be:

\[\tensor{X}= \left[ \begin{array}{cc}
f_{A1} & f_{B1} \\
f_{A2} & f_{B2} \\
f_{A3} & f_{B3}\\
f_{A4} & f_{B4}\\
f_{A5} & f_{B5}\end{array} \right]\]

While the connectivity graph appears as the most straightforward approach to pooling polycrystal microstuctural information in a non-Euclidean domain, this is not necessarily valid for other applications. While, for a polycrystal material, the connectivity graph could possibly remain constant with time, this would not be the case for a granular material (grain contacts). Another graph descriptor should be constructed that would evolve with time. For the flow modelling of a porous material, other graph descriptors could be more important (pore space, flow network).

\section{Deep learning on graphs}
\label{sec:deeplearning}

Machine learning often involves algorithms designed to statistically estimate highly complex functions by learning from data. Some common applications in machine learning are those of regression and classification. A \textbf{regression} algorithm attempts to make predictions of a numerical value provided some input data. A \textbf{classification} algorithm attempts to assign a label to an input and place it to one or multiple classes / categories that it belongs to. Classification tasks can be \textbf{supervised}, if information for the true labels of the inputs are available during the learning process. Classification tasks can also be \textbf{unsupervised}, if the algorithm is not exposed to the true labels of the input during the learning process but attempts to infer labels for the input by learning properties of the input dataset structure. The hybrid geometric learning neural network introduced in this work performs simultaneously an unsupervised classification of polycrystal graph structures and the regression of an anisotropic elastic energy potential functional.

In the following sections, we firstly introduce several basic machine learning and deep neural network terminologies that will be encountered in this work (Section~\ref{sec:machine_learning}). We provide an overview the fundamental deep learning architecture of the multilayer-perceptron (MLP) - that will also carry the regression part of the hybrid architecture. In Section~{\ref{sec:gcn_formulation}, we introduce the novel application of the graph convolution technique that will carry out the unsupervised classification of the polycrystals. Finally, in Section~{\ref{sec:hybrid_architecture}, we introduce our hybrid architecture that combines these two architectures to perform their tasks simultaneously.

\subsection{Deep learning for regression}
\label{sec:machine_learning}

To describe a machine learning algorithm, a \textbf{dataset}, a \textbf{model}, a \textbf{loss function}, and an \textbf{optimization procedure} must be specified. The dataset refers to the total samples that are available for the training and testing of a machine learning algorithm. A \textbf{dataset} is commonly split in training, validation and testing sets. The training set will be used for the algorithm to be trained on and learned from. The validation set is used, while the learning process takes place, to evaluate the the learning procedure and optimize the learning algorithm. The testing set consists of unseen data - data exclusive from the training set - to test the algorithm's blind prediction capabilities, after the learning process is complete. A (parametric) \textbf{model} refers to the structure that holds the parameters that describe the learned function - the number of these parameters are finite and fixed before any data is observed. A \textbf{loss function} (usually also referred to as cost, error or objective function) refers to a metric that must be either minimized or maximized during learning for the learning to be successful - the values of this function drive the learning process. The \textbf{optimization procedure} refers the numerical method utilized to find the optimal parameters of the model that minimize or maximize the loss function. A more complete discussion on machine learning and neural networks can be found in, for instance, \citep{Goodfellow-et-al-2016}. 

\begin{figure}
\centering
\includegraphics[width=11.0cm,angle=0]{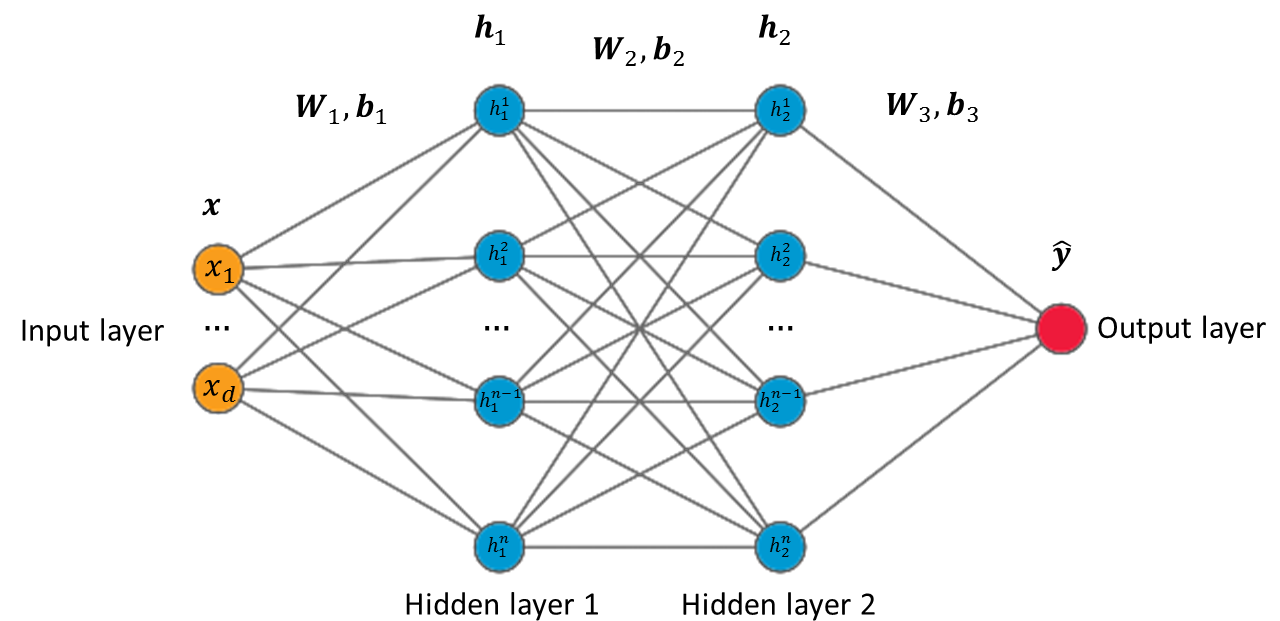}
\caption{A two-layer perceptron. The input vector \tensor{x} has $d$ features, each of the two hidden layers $h_l$ has $m$ neurons. } 
\label{fig:mlp}
\end{figure}

A subset of machine learning algorithms that can learn from high-dimensional data are the artificial neural network (ANN) and deep learning algorithms. Inspired by the structure and function of biological neural networks, ANNs can learn to perform highly complex tasks on large datasets, such as those of image, audio, and video data. One of the simplest ANN architectures would be the multilayer perceptron (MLP) or often called feed forward neural network. The formulation for the two-layer perceptron in Fig.~\ref{fig:mlp}, similar to the one that is also used in this work, is presented below as a series of matrix multiplications:

\begin{align}
\textbf{z}_1 &= \textbf{x}\textbf{W}_1 + \textbf{b}_1 \\
\textbf{h}_1 &= \sigma(\textbf{z}_1) \\
\textbf{z}_2 &= \textbf{h}_1 \textbf{W}_2 + \textbf{b}_2 \\
\textbf{h}_2 &= \sigma(\textbf{z}_2) \\
\textbf{z}_{\text{out}} &= \textbf{h}_2 \textbf{W}_3 + \textbf{b}_3 \\
\hat{\textbf{y}} &= \sigma_{\text{out}}(\textbf{z}_{\text{out}}).
\end{align}

In the above formulation, the input vector $\tensor{x}_l$ contains the features of a sample, the weight matrix $\tensor{W}_l$ contains the weights - parameters of the network, and $\tensor{b}_l$ is the bias vector for every layer. The function $\sigma$ is the chosen activation function for the hidden layers. In the current work, the MLP hidden layers have the ELU function as an activation function, defined as:

\begin{equation}
\label{eq:ELU}
\text{ELU}(\alpha)= \left\{
\begin{array}{ll}
      e^\alpha - 1, & \alpha < 0 \\
      \alpha, & \alpha \geq 0.\\ 
\end{array} 
\right. 
\end{equation}

The vector $\tensor{h}_l$ contains the activation function values for every neuron in the hidden layer. The vector $\hat{\tensor{y}}$ is the output vector of the network with linear activation $\sigma_{\text{out}}(\bullet)=(\bullet)$. 

If $\tensor{y}$ are the true function values corresponding to the inputs $\tensor{x}$, then the MLP architecture could be simplified as an approximator function $\hat{\tensor{y}} = \hat{\tensor{y}}(\tensor{x}|\tensor{W},\tensor{b})$ with inputs $\tensor{x}$ parametrized by $\tensor{W}$ and $\tensor{b}$, such that:

\begin{equation}
\tensor{W}',\tensor{b}' = \argmin_{\tensor{W},\tensor{b}}\ell(\hat{\tensor{y}}(\tensor{x}|\tensor{W},\tensor{b}),\tensor{y}) ,
\label{eq:minimize_loss}
\end{equation} 

where $\tensor{W}'$ and $\tensor{b}'$ are the optimal weights and biases of the neural network that arrive from the optimization - training process such that a defined loss function $\ell$ is minimized. The loss functions used in this work are discussed in Section~\ref{sec:sob_training}.

The fully-connected (Dense) layer that is used as the hidden layer that is used for a standard MLP architecture has the following general formulation:

\begin{equation}
\tensor{h}^{(l+1)}_{\text{dense}} = \sigma(\tensor{h}^{(l)}\tensor{W}^{(l)} + \tensor{b}^{(l)}) .
\label{eq:denselayer}
\end{equation} 

The architecture described above will constitute the energy functional regression branch of the hybrid architecture described in Section~\ref{sec:hybrid_architecture}. It is noted, as it will be discussed later in Section~\ref{sec:graph_based_model}, that this architecture would be sufficient to predict the energy functional for a single polycrystal with the strain being the sole input and the energy functional the output. To predict the complex behavior of multiple polycrystals, the hybrid architecture is introduced in the following sections.

\subsection{Graph convolution network for unsupervised classification of polycrystals}
\label{sec:gcn_formulation}

Geometric learning refers to the extension of previously established neural network techniques to graph structures and manifold-structured data. Graph Neural Networks (GNN) refers to a specific type of neural networks architectures that operate directly on graph structures. An extensive summary of different graph neural network architectures currently developed can be found in \citep{wu_comprehensive_2019}. Graph convolution networks (GCN) \citep*{defferrard_convolutional_2016,kipf_semi-supervised_2017}  are variations of graph neural networks that bear similarities with the highly popular convolutional neural network (CNN) algorithms, commonly used in image processing \citep*{lecun_gradient-based_1998,krizhevsky_imagenet_2012}. The mutual term convolutional refers to use of filter parameters that are shared over all locations in the graph similar to image processing. Graph convolution networks are designed to learn a function of features or signals in graphs $\mathbb{G} = (\mathbb{V,E})$ and they have demonstrated competitive scores at tasks of classification \citep*{kipf_semi-supervised_2017,simonovsky2017dynamic}. 

In this current work, we utilize a GCN layer implementation similar to that introduced in \citep{kipf_semi-supervised_2017}. The implementation is based on the open-source neural network library Keras  \citep{chollet2015keras} and the open-source library on graph neural networks Spektral \citep*{noauthor_spektral_nodate}. The GCN layers will be the ones that learn from the polycrystal connectivity graph information. A GCN layer accepts two inputs, a symmetric normalized graph Laplacian matrix $\tensor{L^{\text{sym}}}$ and a node feature matrix $\tensor{X}$ as described in Section~\ref{sec:graph_theory}. The matrix $\tensor{L^{\text{sym}}}$ holds the information about the graph structure. The matrix $\tensor{X}$ holds information about the features of every node in the graph - every crystal in the polycrystal. In matrix form, the GCN layer has the following structure:

\begin{equation}
\tensor{h}^{(l+1)}_{\text{GCN}} = \sigma(\tensor{L^{\text{sym}}}\tensor{h}^{(l)}\tensor{W}^{(l)} + \tensor{b}^{(l)}) .
\label{eq:GCNlayer}
\end{equation} 

In the above formulation, $\tensor{h}^l$ is the output of a layer $l$. For $l=0$, the first GCN layer of the network accepts the graph features as input such that $\tensor{h}^0 = \tensor{X}$. For $l>1$, $\tensor{h}$ represents a higher dimension representation of the graph features that are produced from the convolution function, similar to a CNN layer.
The function $\sigma$ is a non-linear activation function. In this work, the GCN layers use the Rectified Linear Unit activation function, defined as $ReLU(\bullet) = \max(0,\bullet)$. The weight matrix $\tensor{W}^l$ and bias vector $\tensor{b}^l$ are the parameters of the layer that will be optimized during training.  

The matrix $\tensor{L^{\text{sym}}}$ has dimensions $N \times N$, where $N$ is the number of nodes in the graph - crystalline grain in the polycrystal. The node feature matrix $\tensor{X}$ has dimensions of $N \times D$ where $N$ is the number of nodes in the graph and $D$ is the number of used input features (node weights). In this work, four crystal features where used as node weights (the volume and the three Euler angles for each crystal), thus, $D=4$. Unweighted graphs can be used too - in that case the feature matrix is just the identity matrix $\tensor{X}=\tensor{I}$. The matrix $\tensor{L^{\text{sym}}}$ acts as an operator on the node feature matrix $\tensor{X}$ so that, for every node, the sum of every neighbouring node features and the node itself is accounted for. In order to include the features of the node itself, the matrix $\tensor{L^{\text{sym}}}$ comes by using Equation~\ref{eq:graph_laplacian} with the binary adjacency matrix $\tensor{\hat{A}}$ allowing self-loops and the equivalent degree matrix $\tensor{D}$. Using the normalized laplacian matrix $\tensor{L^{\text{sym}}}$, instead of the adjacency matrix $\tensor{\hat{A}}$, for feature filtering remedies possible numerical instabilites and vanishing / exploding gradient issues when using the GCN layer in deep neural networks.

This type of spatial filtering can be of great use in constitutive modelling. In the case of the polycrystals, for example, the neural network model does not solely learn on the features of every crystal separately. It also learns by aggregating the features of the neighboring crystals in the graph and potentially uncover a behavior that stems from the feature correlation between different nodes. This property deems this filtering function a considerable candidate for learning on spatially heterogeneous material structures.

\subsection{Hybrid neural network architecture for simultaneous unsupervised classification and regression}
\label{sec:hybrid_architecture}
\begin{figure}[]
\centering
\includegraphics[width=15.0cm,angle=0]{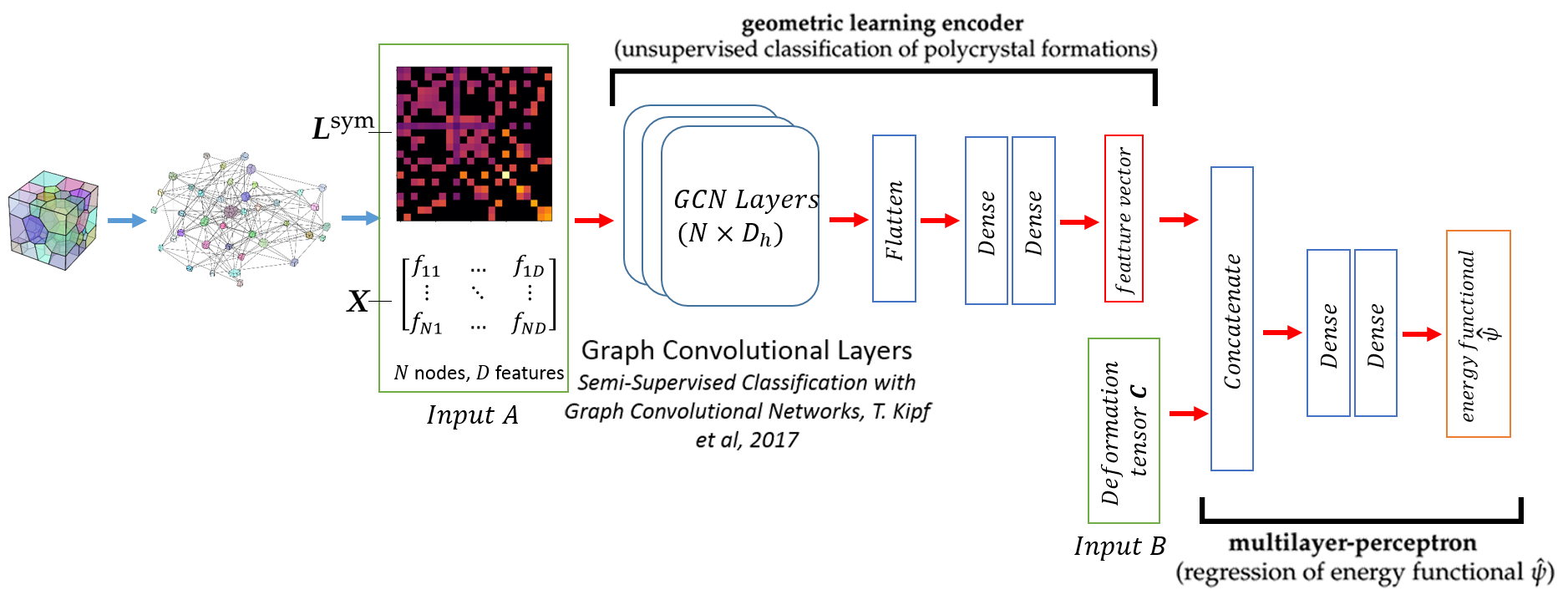}
\caption{Hybrid neural network architecture. The network is comprised of two branches - a graph convolutional encoder and a multi-layer perceptron. The first branch accepts the graph structure (normalized Laplacian $\tensor{L^{\text{sym}}}$) and graph weights (feature matrix $\tensor{X}$) (Input A) as inputs and outputs an encoded feature vector. The second branch accepts the concatenated encoded feature vector and the right Cauchy–Green deformation tensor $\tensor{C}$ in Voigt notation (Input B) as inputs and outputs the energy functional $\hat{\psi}$ prediction.\\} 
\label{fig:hybrid_architecture}
\end{figure}

The hybrid network architecture employed in this current work is designed to perform two tasks simultaneously, guided by a common objective function. The first task is the unsupervised classification of the connectivity graphs of the polycrystals. This is carried through by the first branch of the hybrid architecture that resembles that of a convolutional encoder, commonly used in image classification \citep*{lecun_gradient-based_1998,krizhevsky_imagenet_2012} and autoencoders \citep{vincent_extracting_2008}. However, the convolutional layers are now following the aforementioned GCN layer formulation. A convolutional encoder passes a complex structure (i.e images, graphs) through a series of filters to can generate a higher level representation and encode - compress the information in a structure of lower dimensions (i.e. a vector). It is common practice, for example, in image classification \citep{krizhevsky_imagenet_2012}, to pass an image through a series of stacked convolutional layers, that increase the feature space dimensionality, and then encode the information in a vector through a multilayer perceptron - a series of stacked fully connected layers. The weights of the every layer in the network are optimized using a loss function (usually categorical cross-entropy) so that the output vector matches the classification labels of the input image.

A similar concept is employed for the geometric learning encoder branch of the hybrid architecture. This branch accepts as inputs the normalized graph Laplacian and the node feature matrices. The two convolutional layers read the graph features and increase the dimensionality of the node features. These features are then and flattened and then fed to  two fully connected layers that encode the graph information in a feature vector. The encoded feature vector dense layer can have a linear activation function, similar to regression problems, or a softmax activation function with a range of 0 to 1, similar to multi-label classification problems. Both activation functions have been tested and appear to have comparable results. 

The second task performed by the hybrid network is a regression task - the prediction of the energy functional. The architecture of this branch of the network follows that of a simple feed-forward network with two hidden fully connected layers, similar to the one described in Section~\ref{sec:machine_learning}. The input of this branch is the encoded feature vector, arriving from the geometric learning encoder branch, concatenated with the second order right Cauchy–Green deformation tensor $\tensor{C}$ in Voigt vector notation. The output of this branch is the predicted energy functional $\hat{\psi}$. It is noted that in this current work, an elastic energy functional is predicted and the not history dependent behavior can be adequately mapped with feed-forward architectures. Applications of geometric learning on plastic behavior will be the object of future work and will require recurrent network architectures that can capture the material's behavior history, similar to \citep{wang_multiscale_2018}. 

The layer weights of these two branches are updated in tandem with a common back-propagation algorithm and an objective function that rewards the better energy functional and stress field predictions, using a Sobolev training procedure, described in Section~\ref{sec:sob_training}.  

While this hybrid network architecture provides a promising aspect for incorporating structural data in the form of graphs, there are still several shortcomings that should be addressed in future work. The GCN algorithm itself is not inductive - it cannot introduce new nodes and generalize in terms of the graph structure very efficiently. It is, thus, suggested that the graph structures used in training are statistically similar to each other, so that with adequate regularization the model can generalize on unseen but similar structures. This is the reason why in this current work we focus on making predictions on families of polycrystals with statistically similar crystal number distributions. Simultaneously, we implement rigorous methods of regularization on the graph encoder branch of the hybrid architecture, in the form of Dropout layers \citep{srivastava_dropout:_2014} and $L_2$ regularization. We have discovered that regularization techniques provide a competent method for combating overfitting issues, addressed later in this work. This work is a first attempt to utilizing geometric learning in material mechanics and model refinement will be considered when approaching more complex problems in the future (e.g. history dependent plasticity problems).

\section{Sobolev training for hyperelastic energy functional predictions}
\label{sec:sob_training}

In principle, forecast engines for elastic constitutive responses are trained by (1) an energy-conjugate pair of stress
and strain measures \citep*{ghaboussi_knowledge-based_1991, wang_multiscale_2018, lefik_artificial_2009}, (2) a power-conjugate pair of stress and strain rates \citep{liu_deep_2019} and (3) a pair of strain measure and Helmholtz stored energy \citep{lu_data-driven_2019, huang_predictive_2019}. 
While options (1) and (2)  can both be simple and easy to train once the proper configuration of the neural networks are determined, one critical drawback is that the resultant model may predict non-convex energy response and exhibit ad-hoc path-dependence 
\citep{borja1997coupling}.

An alternative is to introduce supervised learning that takes strain measure as input and output the stored energy functional.
This formulation leads to the so-called hyperelastic or Green-elastic material, which postulate the existence of a Helmholtz free-energy function \citep{holzapfel_new_2000}. 
 The concept of learning a free energy function as a mean to describe multi-scale materials has been previously explored  \citep*{le2015computational,teichert2019machine_a}\nocite{teichert2019machine_b}. However, without direct control of the gradient of the energy functional, the predicted stress and elastic tangential operator may not be sufficiently smooth unless the activation functions and the architecture of the neural network are carefully designed. To rectify the drawbacks of these existing options, we leverage the recent work on Sobolev training  \citep{czarnecki_sobolev_2017} in which we incorporate both the stored elastic energy functional and the derivatives (i.e. conjugate stress tensor) into the loss function such that the objective of the training is not solely minimizing the errors of the energy predictions but the discrepancy of the stress response as well.

Traditional deep learning regression algorithms aim to train a neural network to approximate a function by
minimizing the discrepancy between the predicted values and the benchmark data. 
However, the metric or norm used to measure discrepancy is often the $L_{2}$ norm, which 
does not regularize the derivative or gradients or the learned function. 
When combined with the types of activation functions that include high-frequency basis, 
the learned function may exhibit spurious oscillation and hence not suitable for training 
hyperelastic energy function that requires smoothness for the first and second derivatives.

 Sobolev training we adopted from \citet{czarnecki_sobolev_2017}  is designed to maximize the utilization of data by leveraging the available additional higher order data in the form of higher order constraints in the training objective function. 
In the Sobolev training, objective functions are constructed for minimizing the $H^K$ Sobolev norms of the corresponding Sobolev space.
Recall that a Sobolev space refers to the space of functions equipped with norm comprised of $L^p$ norms of the functions and their derivatives up to a certain order $K$. 

Since it has been shown that neural networks with the ReLU activation function (as well as functions similar to that) can be universal approximators for $C^1$ functions in a Sobolev space \citep{sonoda2017neural}, our goal here is to directly 
predict the elastic energy functional by using the Sobolev norm as loss function to train the hybrid neural network models.

This current work focuses on the prediction of an elastic stored energy functional listed in Eq. \ref{eq:form}, thus, for simplicity, the superscript $e$ (denoting elastic behavior) will be omitted for all energy, strain, stress, and stiffness scalar and tensor values herein. 
In the case of the simple MLP feed- forward network, the network can be seen as an approximator function $\hat{\psi} = \hat{\psi}(\tensor{C}|\tensor{W},\tensor{b})$ of the true energy functional $\psi$ with input the right Cauchy–Green deformation tensor $\tensor{C}$, parametrized by weights $\tensor{W}$ and biases $\tensor{b}$. In the case of the hybrid neural network architecture, the network can be seen as an approximator function $\hat{\psi} = \hat{\psi}(\mathbb{G},\tensor{C}|\tensor{W},\tensor{b})$ of the true energy functional $\psi$ with input the polycrystal connectivity graph information (as described in Fig.~\ref{fig:hybrid_architecture}) and the tensor $\tensor{C}$, parametrized by weights $\tensor{W}$ and biases $\tensor{b}$. The first training objective in Equation~\ref{eq:l2_loss} for the training samples $i \in [1,...,N]$ is modelled after an $L_2$ norm, constraining only $\psi$: 

\begin{equation}
\tensor{W}',\tensor{b}' = \argmin_{\tensor{W},\tensor{b}}\left( \frac{1}{N} \sum_{i=1}^{N} \left\lVert \psi_{i} - \hat{\psi}_{i}\right\rVert^2_2\right).
\label{eq:l2_loss}
\end{equation} 


The second training objective in Equation~\ref{eq:h1_loss} for the training samples $i \in [1,...,N]$ is modelled after an $H_1$ norm, constraining both $\psi$ and its first derivative with respect to $\tensor{C}$ - i.e. one half of the 2nd Piola Kirchhoff stress tensor $\tensor{S}$:

\begin{equation}
\tensor{W}',\tensor{b}' = \argmin_{\tensor{W},\tensor{b}}\left( \frac{1}{N} \sum_{i=1}^{N} \left\lVert \psi_{i} - \hat{\psi}_{i}\right\rVert^2_2 + \left\lVert \frac{\partial \psi_{i}}{\partial \tensor{C}} - \frac{\partial \hat{\psi}_{i}}{\partial \tensor{C}}\right\rVert^2_2 \right),
\label{eq:h1_loss}
\end{equation} 

where in the above:

\begin{equation}
\tensor{S} = 2 \frac{\partial \psi}{\partial \tensor{C}}.
\end{equation}


 It is noted that higher order objective functions can be constructed as well, such as an $H_2$ norm constraining the predicted $\hat{\psi}$, stress, and stiffness values. This would be expected to procure even more accurate $\hat{\psi}$ results, smoother stress predictions and more accurate stiffness predictions. However, since a neural network is a combination of linear functions - the second order derivative of the ReLU and its adjacent activation functions is zero, it becomes innately difficult to control the second order derivative during training, thus in this work we mainly focus on the first order Sobolev method. In case it is desired to control the behavior of the stiffness tensor, a first order Sobolev training scheme can be designed with strain as input and stress as output. The gradient of this approximated relationship would be the stiffness tensor. This experiment would also be meaningful and useful in finite element simulations.

\begin{figure}[h]
\centering
\includegraphics[width=15.0cm,angle=0]{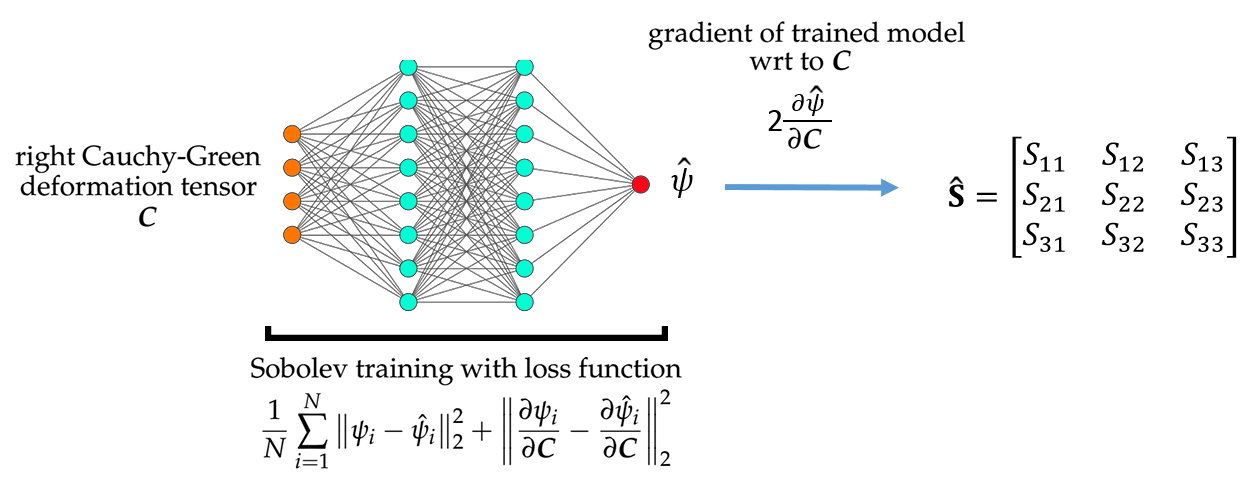}
\caption{Schematic of the training procedure of a hyperelastic material surrogate model with the right Cauchy–Green deformation tensor $\tensor{C}$ as input and the energy functional $\hat{\psi}$ as output. A Sobolev trained surrogate model will output smooth $\hat{\psi}$ predictions and the gradient of the model with respect to $\tensor{C}$ will be a valid stress tensor $\hat{\tensor{S}}$.} 
\label{fig:training_schematic}
\end{figure}

It is noted that, in this current work, the Sobolev training is implemented using the available stress information as the higher order constraint, assuring that the predicted stress tensors are accurate component-wise. In simpler terms, the $H_1$ norm constrains every single component of the second order stress tensor. It is expected that this could be handled more efficiently and elegantly by constraining the spectral decomposition of the stress tensor - the principal values and directions. It has been shown in \citep{heider_invariance_2019} that using loss functions structured to constrain tensorial values in such manner can be beneficial in mechanics-oriented problems and will be investigated in future work.

\remark{
Since the energy functional $\psi$ and the stress values are on different scales of magnitude, the prediction errors are demonstrated using a common scaled metric. For all the numerical experiments in this current work, to demonstrate the discrepancy between any predicted value ($X_{\text{pred}}$) and the equivalent true value ($X_{\text{true}}$) for a sample of size $N$, the following scaled mean squared error (scaled MSE) metric is defined:

\begin{equation}
\text{scaled} \quad MSE = \frac{1}{N} \sum_{i=1}^N \left[(\overline{X}_{\text{true}})_i -(\overline{X}_{\text{pred}})_i \right]^2  \quad \text{with} \quad \overline{X}:= \frac{X - X_{\text{min}}}{X_{\text{max}}-X_{\text{min}}} .
\label{eq:scaled_mse}
\end{equation} 

 The function mentioned above scales the values $X_{\text{pred}}$ and $X_{\text{true}}$ to be in the feature range $[0,1]$. 
}

\section{Verification exercises for checking compatibility with physical constraints}
\label{sec:verification}

While data-driven techniques, such as the neural network architectures discussed in this work, has provided unprecedented efficiency in generating constitutive laws, the consistency of these laws with well-known mechanical theory principles can be rather dubious. Generating black-box constitutive models by blindly learning from the available data is considered to be one of the pitfalls of data-driven methods . If the necessary precautions are not taken, a data-driven model while appearing to be highly accurate in replicating the behaviors discerned from the available database, it may lack the utility of a mechanically consistent law and, thus, be inappropriate to use in describing physical phenomena. In this work, we leverage the mechanical knowledge on fundamental properties of a hyperelastic constitutive laws to check and - if necessary - enforce the consistency of the approximated material models with said properties. In particular for this work, the generated neural network energy functional models are tested for their material frame indifference, isotropy (or lack of), and convexity properties. A brief discussion of these properties is presented in this section, while the verification test results are provided in Section~\ref{sec:verification_tests}.

\subsection{Material Frame Indifference}
\label{sec:mfi}
Material frame indifference or objectivity requires that the energy and stress response of a deformed elastic body remains unchanged, when rigid body motion takes place. The trained models are expected to meet the objectivity condition - i.e. the material response should not depend on the choice of the reference frame. While translation invariance is automatically ensured by describing the material response as a function of the deformation, invariance for rigid body rotations is not necessarily imposed and must be checked. The definition of material frame indifference for an elastic energy functional $\psi$ formulation is described as follows:

\begin{equation}
\psi(\tensor{Q}\tensor{F}) = \psi(\tensor{F}) \: \text{for all} \:  \tensor{F} \in GL^{+}(3,\mathbb{R}), \tensor{Q} \in SO(3),
\label{eq:invariance_formulation}
\end{equation}
where $\tensor{Q}$ is a rotation tensor. The above definition can be proven to expand for the equivalent stress and stiffness measures:

\begin{equation}
P_{iJ}(\tensor{Q}\tensor{F}) = Q_{ij} P_{jJ}(\tensor{F}) \: \text{for all} \:  \tensor{F} \in GL^{+}(3,\mathbb{R}), \tensor{Q} \in SO(3),
\label{eq:invariance_formulation_stress}
\end{equation} 

\begin{equation}
c_{iJkL}(\tensor{Q}\tensor{F}) = Q_{ij} Q_{kl} c_{jJlL}(\tensor{F}) \: \text{for all} \:  \tensor{F} \in GL^{+}(3,\mathbb{R}), \tensor{Q} \in SO(3).
\label{eq:invariance_formulation_stiffness}
\end{equation} 

Thus, a constitutive law is frame-indifferent, if the responses for the energy, the stress and stiffness predictions are left rotationally invariant. Frame invariance requires that \citep{borja_plasticity_2013, kirchdoerfer2016data} , 

\begin{equation}
\label{eq:left_rot_inv_a}
\psi(\tensor{F}) = \psi(\tensor{F^{+}}), \quad \quad \tensor{F^{+}} = \tensor{QF}.
\end{equation}
 
The above is automatically satisfied when the response is modeled as an equivalent function of the right Cauchy-Green deformation tensor $\tensor{C}$, since:

\begin{equation}
\label{eq:left_rot_inv_b}
\tensor{C^{+}} = (\tensor{F^{+}})^{T} \tensor{F^{+}} =  \tensor{F}^{T} \tensor{Q}^{T} \tensor{Q} \tensor{F} = \tensor{F}^{T} \tensor{F} \equiv  \tensor{C}.
\end{equation}

By training all the models in this work as a function of the right Cauchy-Green deformation tensor $\tensor{C}$, this condition is automatically satisfied.

\subsection{Isotropy}
\label{sec:isotropy_theory}
The material response described by a constitutive law is expected to be isotropic, if the following is valid:

\begin{equation}
\psi(\tensor{F}\tensor{Q}) = \psi(\tensor{F}) \: \text{for all} \:  \tensor{F} \in GL^{+}(3,\mathbb{R}), \tensor{Q} \in SO(3).
\label{eq:isotropy_formulation}
\end{equation}

This expands to the stress and stiffness response of the material:

\begin{equation}
\tensor{P}_{iJ}(\tensor{F}\tensor{Q}) = P_{iI}(\tensor{F}) Q_{IJ} \: \text{for all} \:  \tensor{F} \in GL^{+}(3,\mathbb{R}), \tensor{Q} \in SO(3),
\label{eq:isotropy_formulation_stress}
\end{equation} 

\begin{equation}
c_{iJkL}(\tensor{F}\tensor{Q}) = c_{iIkK}(\tensor{F}) Q_{IJ} Q_{KL}  \: \text{for all} \:  \tensor{F} \in GL^{+}(3,\mathbb{R}), \tensor{Q} \in SO(3).
\label{eq:isotropy_formulation_stiffness}
\end{equation} 

Thus, for a material to be isotropic, its response must be right rotationally invariant. In the case that the response is anisotropic, as in the inherently anisotropic material studied in this work, the above should no t be valid. In Section~\ref{sec:isotropy_check}, it is shown that the behavior of the polycrystals predicted by the hybrid architecture is, indeed, anisotropic.

\subsection{Convexity}

To ensure the thermodynamical consistency of the trained neural network models, the predicted energy functional must be convex. Testing the convexity of a black box data-driven function without an explicitly stated equation is not necessarily a straight-forward process. There have been developed certain algorithms to estimate the convexity of black box functions \citep{tamura2019quantitative}, however, it is outside the scope of this work and will be considered in the future. While convexity would be straight-forward to visually check for a low-dimensional function, this is not necessarily true for a high-dimensional function described by the hybrid models.

A function $f:\mathbb{R}^n	\rightarrow \mathbb{R}$ is convex over a compact domain $D$ if for all $x,y \in D$ and all $\lambda \in \left[ 0,1 \right]$, if:

\begin{equation}
\label{eq:convexity}
f(\lambda x + (1- \lambda) y) \geq \lambda f(x) + (1 - \lambda)f(y).
\end{equation}

For a twice differentiable function $f:\mathbb{R}^n	\rightarrow \mathbb{R}$ over a compact domain $D$, the definition of convexity can be proven to be equivalent with the following statement:

\begin{equation}
\label{eq:gradient_inequality}
f(y)\geq f(x) + \nabla f(x)^T(y-x), \quad \text{for all} \quad x,y \in D .
\end{equation} 

The above can be interpreted as the first order Taylor expansion at any point of the domain being a global under-estimator of the function $f$. In terms of the approximated black-box function $\hat{\psi}(\tensor{C},\mathbb{G})$ used in the current work, the inequality~\ref{eq:gradient_inequality} can be rewritten as:

\begin{equation}
\label{eq:gradient_inequality_blackbox}
\hat{\psi}(\tensor{C}_{\alpha},\mathbb{G})\geq \hat{\psi}(\tensor{C}_{\beta},\mathbb{G}) + \frac{\partial \hat{\psi}}{\partial \tensor{C}}(\tensor{C}_{\beta},\mathbb{G}) 
:(\tensor{C}_{\alpha}-\tensor{C}_{\beta}), \quad \text{for all} \quad \tensor{C}_{\alpha},\tensor{C}_{\beta} \in D.
\end{equation} 

The above constitutes a necessary condition for the approximated energy functional for a specific polycrystal  (represented by the connectivity graph $\mathbb{G}$)  to be convex, if it is valid for any pair of right Cauchy deformation tensors $\tensor{C}_{\alpha}$ and $\tensor{C}_{\beta}$ in a compact domain $D$. This check is shown to be satisfied in Section~\ref{sec:convexity_check}.

\remark{
The trained neural network models in this work will be shown in Section~\ref{sec:verification_tests} to satisfy the checks and necessary conditions for being consistent with the expected objectivity, anisotropy, and convexity principles. However, in the case where one or more of these properties appears to be absent, it is noted that it can be enforced during the optimization procedure by modifying the loss function. Additional weighted penalty terms could be added to the loss function to promote consistency to required mechanical principles. For example, in the case of objectivity, the additional training objective, parallel to those expressed in Eq.~\ref{eq:l2_loss} and \ref{eq:h1_loss}, could be expressed as:

\begin{equation}
\tensor{W}',\tensor{b}' = \argmin_{\tensor{W},\tensor{b}}\left( \frac{1}{N} \sum_{i=1}^{N}  \lambda \left\lVert \hat{\psi}(\mathbb{G},\tensor{Q}\tensor{F}|\tensor{W},\tensor{b}) - \hat{\psi}(\mathbb{G},\tensor{F}|\tensor{W},\tensor{b})\right\rVert^2_2  \right), \tensor{Q} \in SO(3),
\label{eq:objectivity_loss}
\end{equation} 

where $\lambda$ is a weight variable, chosen between $[0,1]$, setting the importance of this objective in the now multi-objective loss function, and $\tensor{Q}$ are randomly sampled rigid rotations from the $SO(3)$ group. Constraints of this kind where not deemed necessary in the current paper and will be investigated in future work.
}

\section{FFT offline database generation}
This section firstly introduces the  fast Fourier transform
(FFT) based method for the mesoscale homogenization problem,
which was chosen to efficiently provide the database of graph structures and material responses to be used in geometric learning.
Following that, the anisotropic Fung hyperelastic model is briefly summarized as the constitutive relation at the basis of the simulations.
Finally, the numerical setup is introduced focusing on the numerical discretization,
grain structure generation, and initial orientation of the structures in question.

\subsection{FFT based method with periodic boundary condition}
This section deals with solving mesoscale homogenization problem using an FFT-based method.
Supposing that the mesoscale problem is defined in a 3D periodic domain,
where the displacement field is periodic while the surface traction is anti-periodic,
the homogenized deformation gradient $\overline{\tensor{F}}$ and first P-K stress $\overline{\tensor{P}}$ can be defined as:
\begin{equation}
\overline{\tensor{F}} = \langle \tensor{F} \rangle, \overline{\tensor{P}} = \langle \tensor{P} \rangle,
\end{equation}
where $\langle \cdot \rangle$ denotes the volume average operation.

Within a time step, when the average deformation gradient increment $\Delta \overline{\tensor{F}}$ is prescribed,
the local stress $\tensor{P}$ within the periodic domain can be computed by solving the Lippman-Schwinger equation:
\begin{equation}
\tensor{F} + \tensor{\Gamma}^0 \ast \left( \tensor{P}( \tensor{F} ) - \tensor{C}^0 : \tensor{F} \right ) = \overline{\tensor{F}},
\end{equation}
where $\ast$ denotes a convolution operation, $\tensor{\Gamma}^0$ is Green's operator,
and $\tensor{C}^0$ is the homogeneous stiffness of the reference material.
The convolution operation can be conveniently performed in the Fourier domain,
so the Lippman-Schwinger equation is usually solved by the FFT based spectral method \citep{ma_fft_2019}.
Note that due to the periodicity of the trigonometric basis functions,
the displacement field and the strain field are always periodic.

\subsection{Anisotropic Fung elasticity}
An anisotropic elasticisity model at the mesoscale level is utilized to generate the homogenized response database for then training graph-based model in the macroscale.
In this section, a generalized Fung elasticity model is utilized as the mesoscale constitutive relation
due to its frame-invariance and convenient implementation \citep{fung1965foundations}.

In the generalized Fung elasticity model, the strain energy density function $W$ is written as:
\begin{equation}
W = \frac{1}{2}c\left[ \exp \left( Q \right) - 1 \right], \quad
Q = \frac{1}{2} \tensor{E} : \tensor{a} : \tensor{E},
\end{equation}
where $c$ is a scalar material constant, $\tensor{E}$ is the Green strain tensor,
and $\tensor{a}$ is the fourth order stiffness tensor.
The material anisotropy is reflected in the stiffness tensor $\tensor{a}$,
which is a function of the spatial orientation and the material symmetry type.

For a material with orthotropic symmetry, the strain energy density can be written in a simpler form as:
\begin{equation}
Q = c^{-1} \sum_{a=1}^3 \left[ 2 \mu_a \tensor{A}_a^0 : \tensor{E}^2 + \sum_{b=1}^3 \lambda_{ab}
\left( \tensor{A}_a^0:\tensor{E} \right) \left( \tensor{A}_b^0:\tensor{E} \right) \right], \quad
\tensor{A}_a^0 = \vec{a}_a^0 \otimes \vec{a}_a^0,
\end{equation}
where $\mu_a$ and $\lambda_{ab}$ are anisotropic Lam\'{e} constants,
and $\vec{a}_a^0$ is the unit vector of the orthotropic plane normal,
which represents the orientation of the material point in the reference configuration.
Note that $\lambda_{ab}$ is a symmetric second order tensor,
and the material symmetry type becomes cubic symmetry when certain values of
$\tensor{\lambda}$ and $\tensor{\mu}$ are adopted.

The elastic constants take the value:
\begin{equation}
c = 2 \textrm{ (MPa)},
\lambda = \begin{bmatrix}
0.6 & 0.7 & 0.6 \\ 
0.7 & 1.4 & 0.7\\ 
0.6 & 0.7 & 0.5
\end{bmatrix}
\textrm{ (MPa)},
\mu = \begin{bmatrix}
0.1\\ 
0.7\\ 
0.5
\end{bmatrix}
\textrm{ (MPa),}
\end{equation}
and remain constant across all the mesoscale simulations.
The only changing variable is the grain structure and the initial orientation of the representative volume element (RVE),
which is introduced in the following section.

\subsection{Numerical aspects for database generation}
The grain structures and initial orientations of the mesoscale simulations
are randomly generated in the parameter space to generate the database.
The mesoscale RVE is equally divided into $49 \times 49 \times 49$ grid points
to maintain a high enough resolution at an acceptable computational cost.
The grain structures are generated by the open source software NEPER \citep{quey_large-scale_2011}.
An equiaxed grain structure is randomly generated with 40 to 50 grains.
A sample RVE is shown in Figure \ref{fig:RVE_RM}.


\begin{figure}[!htb]
\begin{center}
\subfigure[Sample polycrystal microstructure.]{
\includegraphics[width=0.3\textwidth]{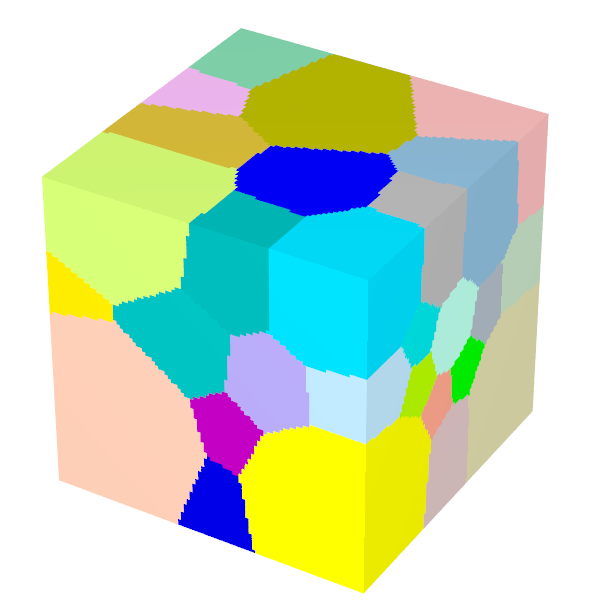}
}%
\subfigure[Sample initial orientation.]{
\includegraphics[width=0.6\textwidth]{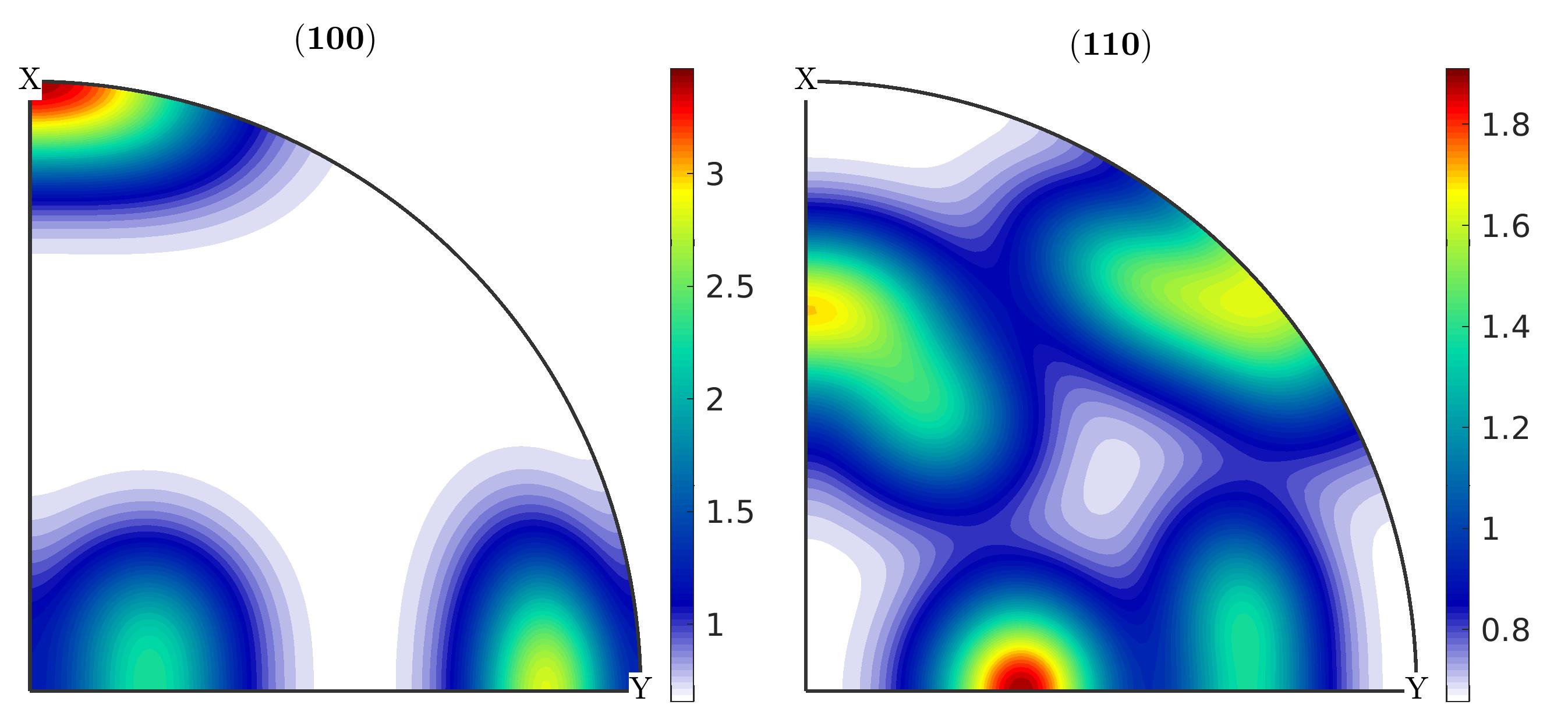}
}
\end{center}
\caption{
Sample of the randomly generated initial microstructure:
(a) Initial RVE with 50 equiaxed grains, which is equally discretized by $49 \times 49 \times 49$ grid points;
(b) Pole figure plot of initial orientation distribution function (ODF) combining uniform and unimodal ODF.
The Euler angles of the unimodal direction are $\left(207.1^{\circ}, 17.8^{\circ}, 159.0^{\circ}\right)$ in Bunge notation,
and the half width of the unimodal ODF is $10^{\circ}$.
The weight value is 0.66 for uniform ODF and 0.34 for unimodal ODF.
}
\label{fig:RVE_RM}
\end{figure}

The initial orientations are generated using the open source software MTEX \citep{bachmann_texture_2010}.
The orientation distribution function (ODF) is randomly generated by combining
uniform orientation and unimodal orientation:
\begin{equation}
f(\vec{x}; \vec{g}) = w + (1-w) \psi \left( \vec{x}, \vec{g} \right),\quad \vec{x} \in SO(3),
\end{equation}
where $w \in [0,1]$ is a random weight value, $\vec{g} \in SO(3)$ is a random modal orientation,
and $\psi \left( \vec{x}, \vec{g} \right)$ is the von Mises--Fisher distribution function
considering cubic symmetry.
The half width of the unimodal texture $\psi \left( \vec{x}, \vec{g} \right)$ is $10^{\circ}$,
and the preferential orientation $g$ of the unimodal texture is also randomly generated.
A sample initial ODF is shown in Figure \ref{fig:RVE_RM} (b).

The average strain is randomly generated from an acceptable strain space,
and simulations are performed for each RVE with 200 average strains.
Note that the constitutive relation is hyperelastic, so the simulation result is path independent.
To avoid any numerical convergence issues, the range of each strain component $\left( \tensor{F} - \tensor{I} \right)$ is
between $0.0$ and $0.1$ in the RVE coordinate.

\section{Numerical Examples}

The predictive capabilities of the hybrid 
 geometric learning have been tested  on a set of anisotropic hyperelastic behavior data  inferred from simulations conducted on 150 polycrystal RVEs, as described in the previous section. While a supervised learning using 
energy-conjugate pair as training data may infer a surrogate model 
for one particular RVE, this black-box approach cannot further generate 
a single surrogate to other RVE without comprising accuracy \citep{wang2019meta}. This problem can also be interpreted as attempting to describe a behavior with an incomplete basis. If the basis of the model is missing an axis, the prediction of the model is nothing but a projection of the true prediction on the missing axis. Thus, model of the ${\cal M}^{H1}_{mlp}$ type may demonstrate a decent accuracy score, since the deviation of the dataset values from the mean is not rather large, but the model itself lacks any significant mechanical meaning, as it cannot fully describe the anisotropic response - only a projection of the true behavior on the missing bases that describe the anisotropy.
 
One major advantage of the hybrid \citep{frankel_predicting_2019} or grah-based training \citep{wang_multiscale_2018} is the ability to generalize the 
forward prediction ranges by introducing RVE data as an additional initial input. 
Consequently, one hybrid learning may generate a \textit{constitutive
law for a family of RVE} instead of a surrogate model specified for one RVE. 
This important distinction is demonstrated in the first numerical example.  
Comparison results for different combinations of architecture and training procedures are showcased. It is also verified that the hybrid material model is innately frame invariant and anisotropic.

Following this first example, the hybrid neural network is utilized as a material model in a brittle fracture finite element parametric study. The problem formulation is briefly discussed and then the neural network model's anisotropic behavior is showcased through a series of dynamic fracture  numerical experiments.

To facilitate the qualitative visualization of the training and testing results, the scaled MSE performances of different models are represented using non-parametric, empirical cumulative distribution functions (eCDFs), as in \citep*{kendall1946advanced, Gentle2009}. The results are plotted in scaled MSE vs eCDF curves in semilogarithmic scale for the training and testing partitions of the dataset. The distance between these curves can be a qualitative metric for the performance of the various models on various datasets - e.g. the distance of the eCDF curves of a model for the training and testing datasets is a qualitative metric of the overfitting phenomenon. For a dataset $N$ with $\text{MSE}_i$ sorted in ascending order, the eCDF can be computed as follows: 

\begin{equation}
F_{N}(\text{MSE}) = \left \{
\begin{aligned}
&0, &\text{MSE} < \text{MSE}_1,\\
&\frac{r}{N}, &\text{MSE}_{r} \leq \text{MSE} < \text{MSE}_{r+1},\ r = 1,...,N-1,\\
&1, &\text{MSE}_{N} \leq \text{MSE}\,.
\end{aligned}
\right .
\label{eq:eCDFmse}
\end{equation}

In the following sections and applications, we use abbreviations related to each of the model architectures and training algorithms as summarized in Table \ref{tab:AbbrevModels}.
\begin{table}[h]
\centering
\caption{Summary of the considered model and training algorithm combinations.}
\label{tab:AbbrevModels}       
{
\begin{tabular}{p{0.8cm}p{10.5cm}}
\hline\noalign{\smallskip}
Model & Description\\
\noalign{\smallskip}\hline\\[-3mm]
${\cal M}^{L2}_{mlp}$ & Multilayer perceptron feed-forward architecture. Loss function used is the $L_2$ norm (Eq.~\ref{eq:l2_loss})  \\[2mm]
${\cal M}^{H1}_{mlp}$ & Multilayer perceptron feed-forward architecture. Loss function used is the $H_1$ norm (Eq.~\ref{eq:h1_loss})\\[3mm]
${\cal M}^{H1}_{hybrid} $ & Hybrid architecture described in Fig.~\ref{fig:hybrid_architecture}. Loss function used is the $H_1$ norm (Eq.~\ref{eq:h1_loss})\\[3mm]
${\cal M}^{H1}_{reg} $ & Hybrid architecture described in Fig.~\ref{fig:hybrid_architecture}. Loss function used is the $H_1$ norm (Eq.~\ref{eq:h1_loss}). The geometric learning branch of the network is regularized against overfitting. \\
\noalign{\smallskip}\hline
\end{tabular}
}
\end{table}

It is noted that, in order to compare all the models in consideration in equal terms, the neural network training hyperparameters throughout all the experiments were kept identical wherever it was possible. This includes hyperparameters such as the number of training epochs, the type of optimizer, and learning rates as well as training techniques such as reduction of the learning rate when the loss would stop decreasing. The learning capacity of the models (i.e. layer depth and layer dimensions) for the multilayer perceptrons for ${\cal M}^{L2}_{mlp}$  and ${\cal M}^{H1}_{mlp}$, as well as the multilayer perceptron branch of the ${\cal M}^{H1}_{hybrid} $ and ${\cal M}^{H1}_{reg} $ was kept identical. In this current work, the values used for the hyperparameters were deemed adequate to provide as accurate results as possible for all methods while maintaining fair comparison terms. The optimization of these hyperparameters to achieve the maximum possible accuracy will be the objective of future work. The node weights - features that were used as inputs for the feature matrix $\tensor{X}$ of geometric learning branch were the crystal volumes and the three Euler angles for each crystal. 

\subsection{Training constitutive models for polycrystals with non-Euclidean data}
\label{sec:graph_based_model}

The ability to capture the elastic stored energy functional of a single polycrystal is initially tested with a simple experiment. 

To determine whether the incorporation of graph data improves the 
accuracy and robustness of the forward prediction, we both conduct 
the hybrid learning and the classical supervised machine learning.
The latter is used as a control experiment. 
First, a two-hidden-layer feed-forward neural network is trained and tested on 200 sample points - 200 different, randomly generated deformation tensors with their equivalent elastic stored energy and stress measure. Sobolev training  described in Section ~\ref{sec:sob_training} (model ${\cal M}^{H1}_{mlp}$) was utilized. 
Then, this architecture are incorporated into the hybrid learning where it constitutes the multilayer perceptron branch of the hybrid network described previously in Fig.~\ref{fig:hybrid_architecture}. To eliminate as much as possible any objectivity on the dataset of the experiment, the networks capability is tested with a K-fold cross validation algorithm (cf. \citep{bengio2004no}). The 200 sample points are separated into 10 different groups - folds of 20 sample points each and, recursively, a fold is selected as a testing set and the rest are selected as training set for the network. 

The K-Fold testing results can be seen in Fig.~\ref{fig:single_surrogate_sample} where the model can predict the data for a single RVE formation adequately, as well as interpolate smoothly between the data points to generate the response surface estimations for the energy and the stress field (Fig.~\ref{fig:single_surrogate}). A good performance for both training and testing on a single polycrystal was expected as no additional input is necessary, other than the strain tensor. Any additional input - i.e. structural information - would be redundant in the training since it would be constant for the specific RVE.

\begin{figure}[t]
\centering
\begin{minipage}{.5\linewidth}
  \centering
\includegraphics[width=8.0cm,angle=0]{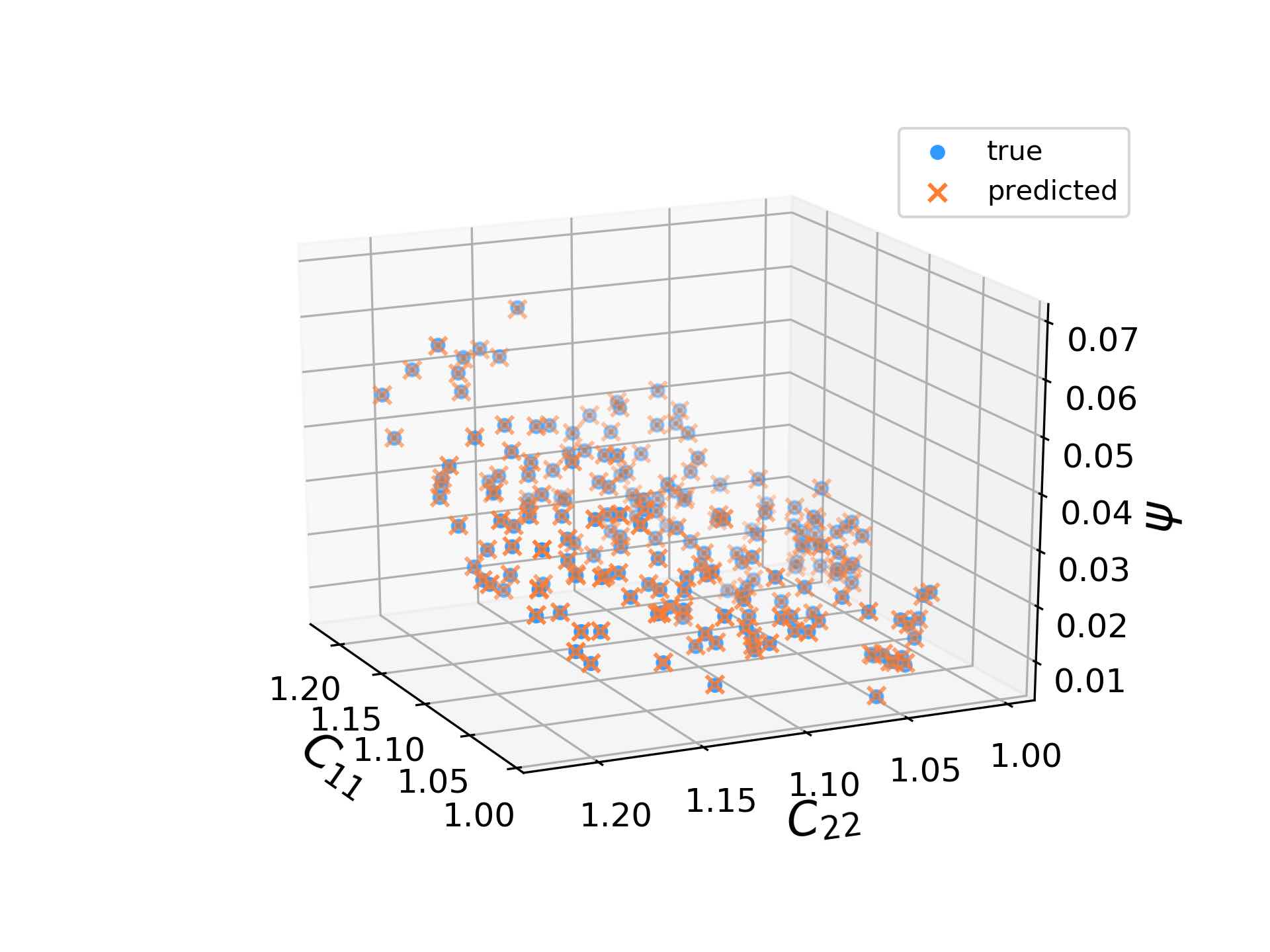}
\end{minipage}%
\begin{minipage}{.5\linewidth}
  \centering
\vspace*{0.2cm}\includegraphics[width=8.0cm,angle=0]{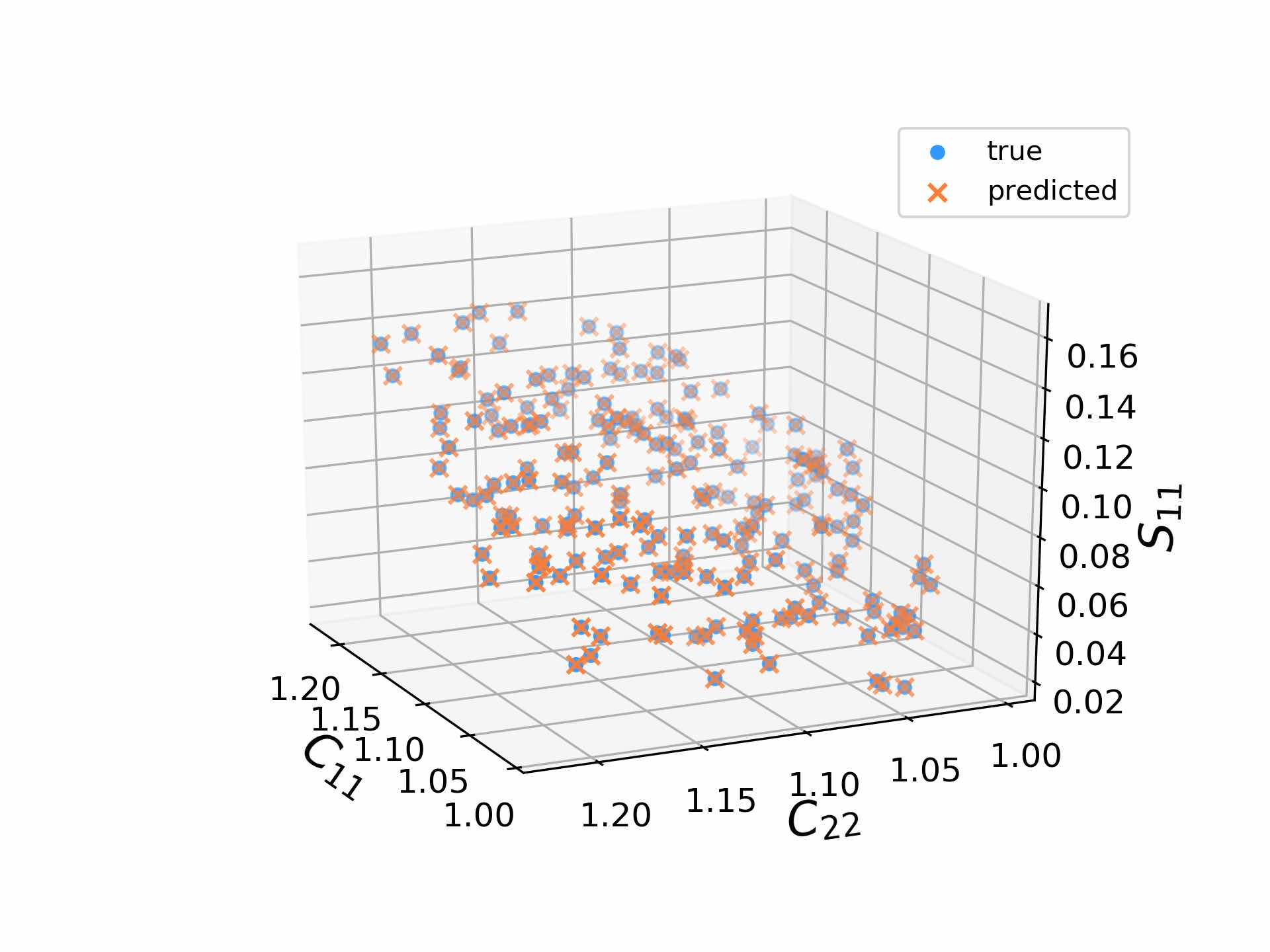}
\end{minipage}
\caption{K-fold testing results for the energy functional $\psi$ (left) and the first component of the 2nd Piola-Kirchhoff stress tensor (right) by a surrogate neural network model ${\cal M}^{H1}_{mlp}$ trained on data for a single RVE. The tensor components of the right Cauchy–Green deformation tensor $\tensor{C}$ are randomly generated for each polycrystal training dataset. To illustrate the multidimensional data, a projection of all the sample points on the $C_{11}$ and $C_{22}$ axes is demonstrated. } 
\label{fig:single_surrogate_sample}
\end{figure}

\begin{figure}[h!]
\centering
\begin{minipage}{.5\linewidth}
  \centering
\includegraphics[width=8.0cm,angle=0]{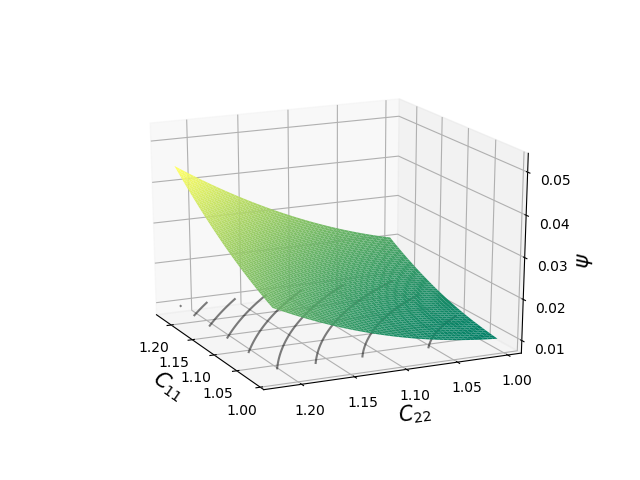}
  \label{fig:psi_estimation_surrogate}
\end{minipage}%
\begin{minipage}{.5\linewidth}
  \centering
\includegraphics[width=8.0cm,angle=0]{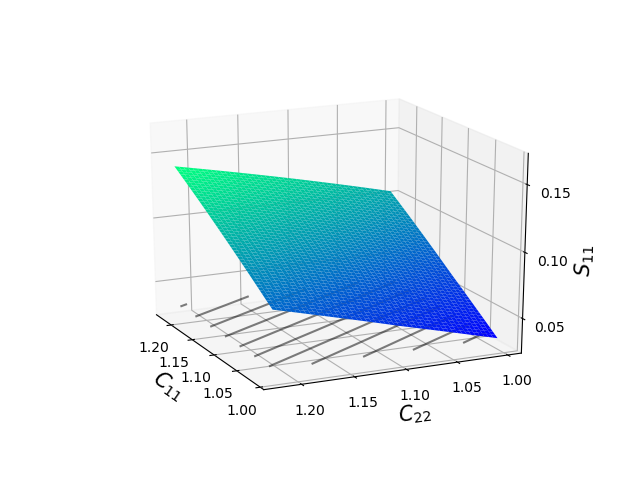}
  \label{fig:P11_estimation_surrogate}
\end{minipage}
\caption{Estimated $\psi$ energy functional surface (left) and the first component of the 2nd Piola-Kirchhoff stress tensor (right) generated by a surrogate neural network model (${\cal M}^{H1}_{mlp}$) trained on data for a single RVE.} 
\label{fig:single_surrogate}
\end{figure}

In this current work, we generalize the learning problem by introducing the polycrystal weighted connectivity graph as the additional input data. 
This connectivity graph is inferred directly from the micro-structure by assigning each grain in the poly-crystal 
as a vertex (node) and assigning edge on each grain contact pair.

\begin{figure}
\centering
\includegraphics[width=15.0cm,angle=0]{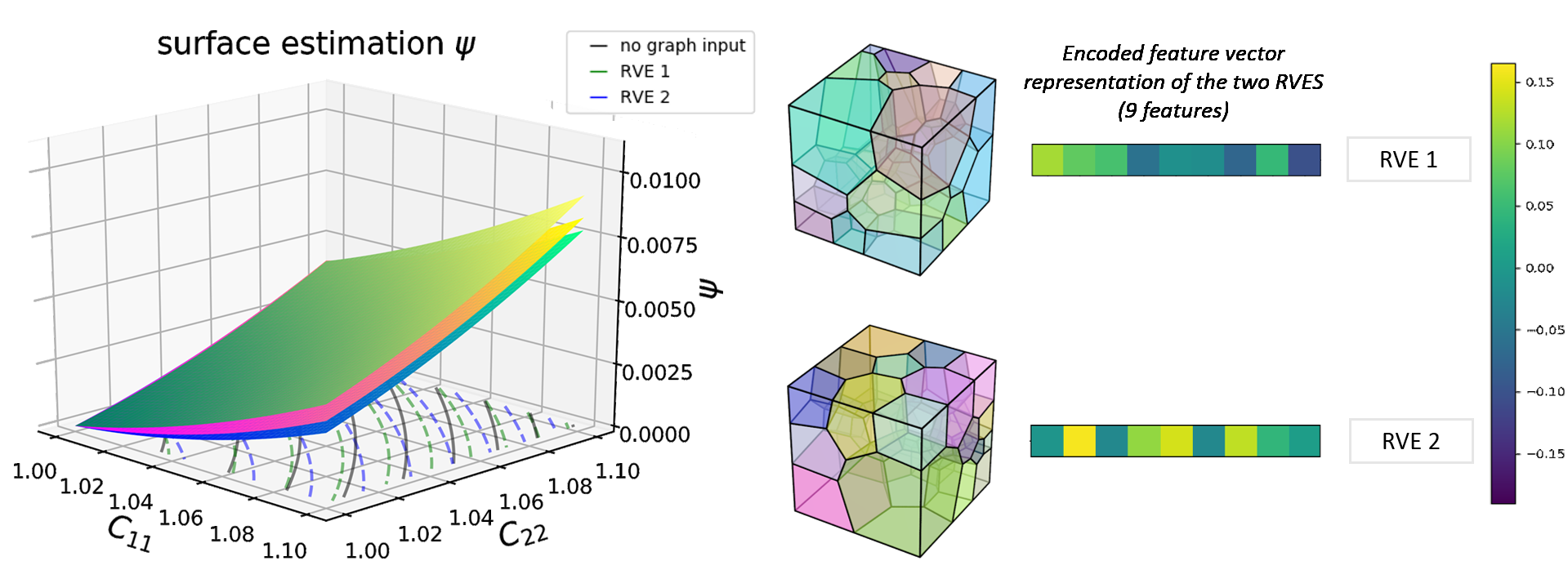}
\caption{Without any additional input (other than the strain tensor), the neural network cannot differentiate between the behavior of the two polycrystals. The two anisotropic behaviors can be distinguished when the weighted connectivity graph is also provided as input. Through the unsupervised encoding branch of the hybrid architecture, each polycrystal is mapped on an encoded feature vector. The feature vector is fed to the multilayer perceptron branch and procures a unique energy prediction.} 
\label{fig:surface_estimation_2RVES}
\end{figure}

It is shown that the hybrid architecture proposed in Fig.~\ref{fig:hybrid_architecture} can leverage the information from a weighted connectivity graph to perform this task. The next experimental setup expands to learning over multiple polycrystals. As previously mentioned, the graph convolutions work better in statistically similar graph structures, thus we consider a family of 100 polycrystals with similar number of crystals, ranging from 40 to 50 crystals. A K-fold validation algorithm is performed on these 100 randomly generated polycrystal RVEs. The 100 RVEs are separated into 5 folds of 20 RVEs each. In doing so, every polycrystal RVE will be considered as blind data for the model at least once. The K-fold cross validation algorithm is repeated for the model architectures and training algorithms ${\cal M}^{L2}_{mlp}$, ${\cal M}^{H1}_{mlp}$, and ${\cal M}^{H1}_{reg}$. The results are presented as scaled MSE vs eCDF curves for the energy functional $\psi$ and  second Piola-Kirchhoff stress $\tensor{\tau}$ tensor principal values and principal direction predictions in Fig.~\ref{fig:kfold_ecdf}. It can be seen that using the Sobolev training method greatly reduces the blind prediction errors - both the ${\cal M}^{L2}_{mlp}$ energy and stress prediction errors are higher than those of the ${\cal M}^{H1}_{mlp}$ and ${\cal M}^{H1}_{reg}$ models. The ${\cal M}^{H1}_{reg}$ model demonstrates superior predictive results than the ${\cal M}^{H1}_{mlp}$ model, as it can distinguish between different RVE behaviors.

\begin{figure}[h!]

\centering
\begin{minipage}{.5\linewidth}
  \centering
\includegraphics[height=5cm,angle=0]{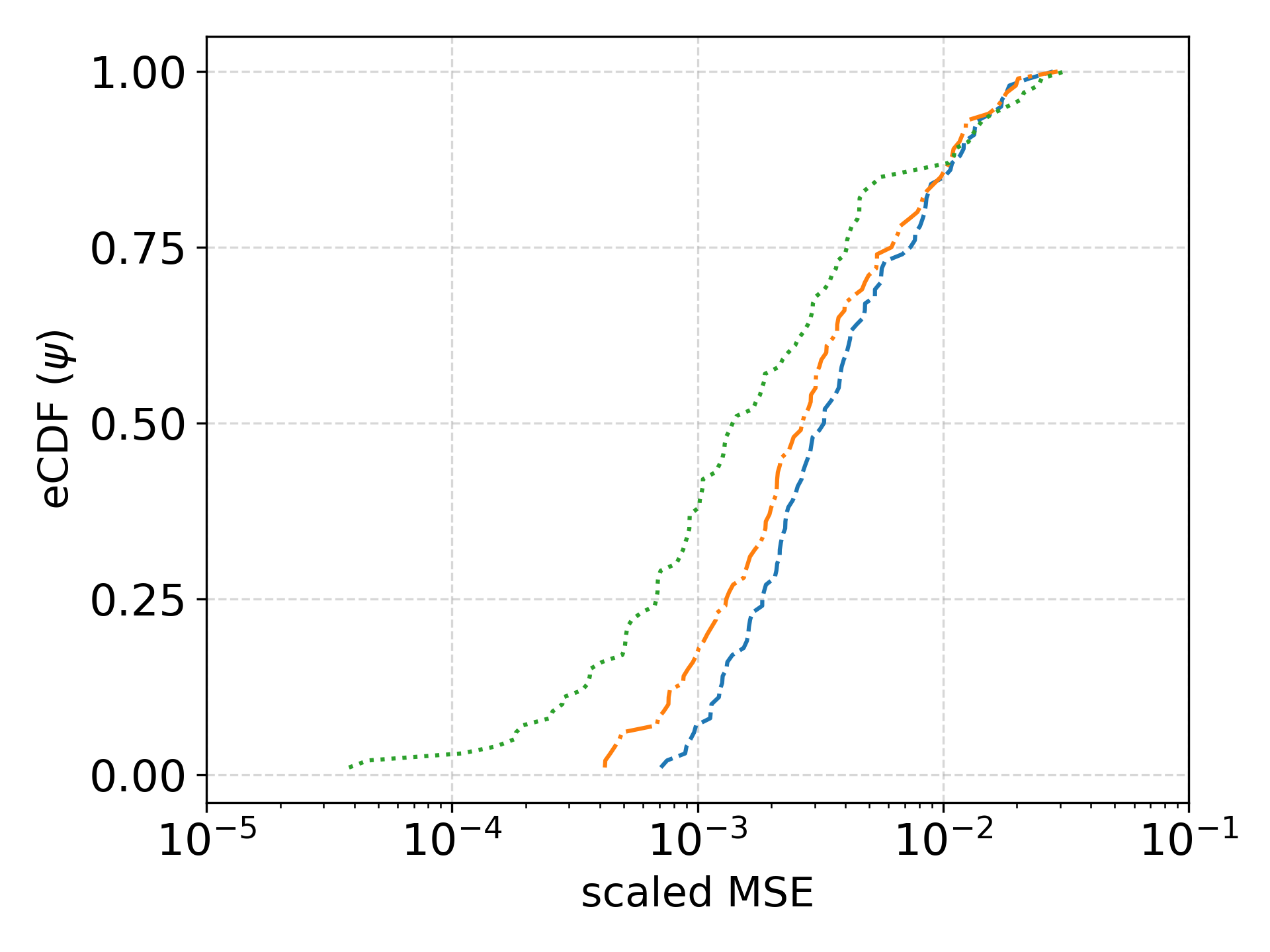}
\end{minipage}%
\begin{minipage}{.5\linewidth}
  \centering
\includegraphics[height=5cm,angle=0]{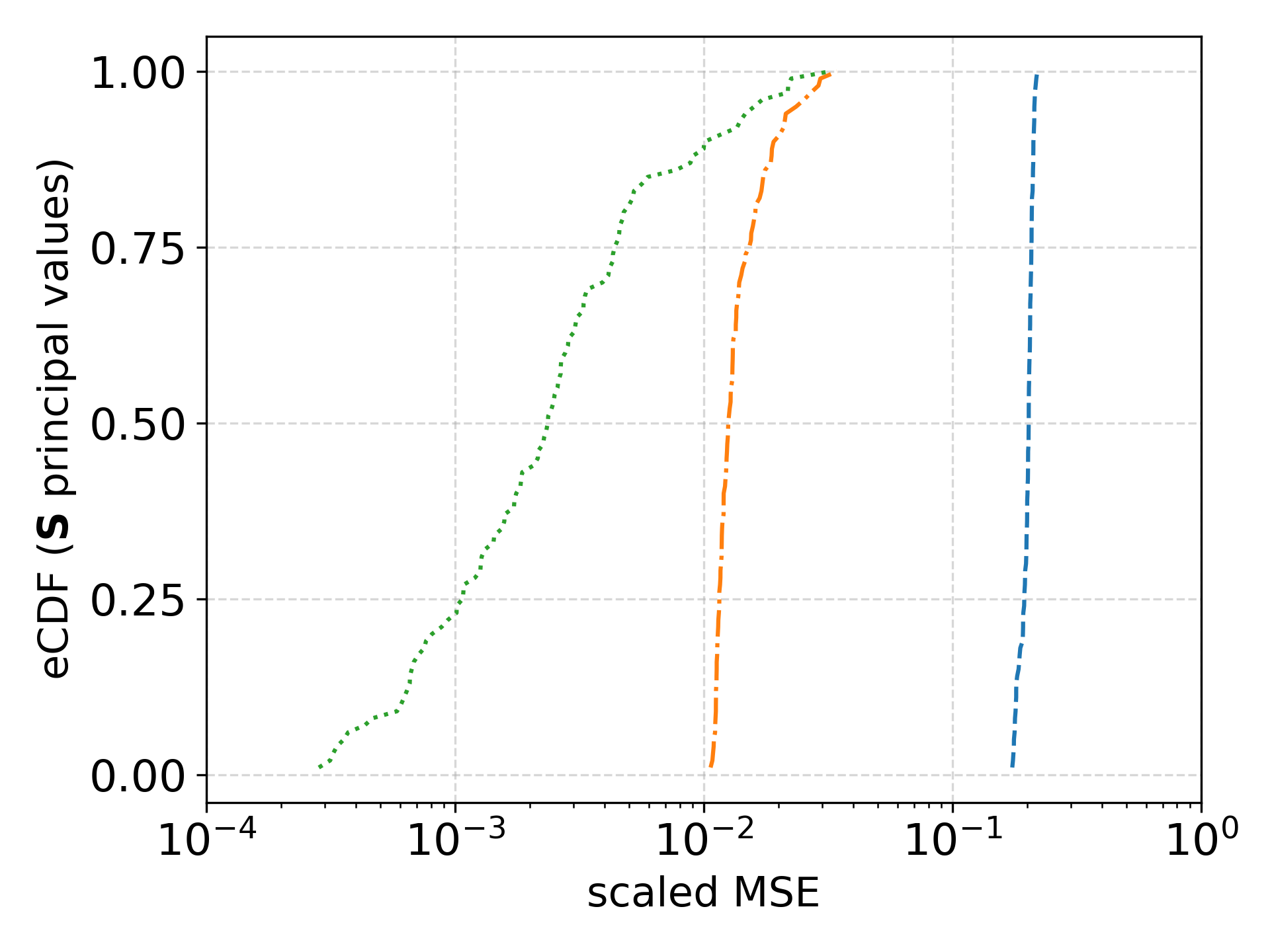}
\end{minipage}
\hspace*{3cm}\includegraphics[height=5cm,angle=0]{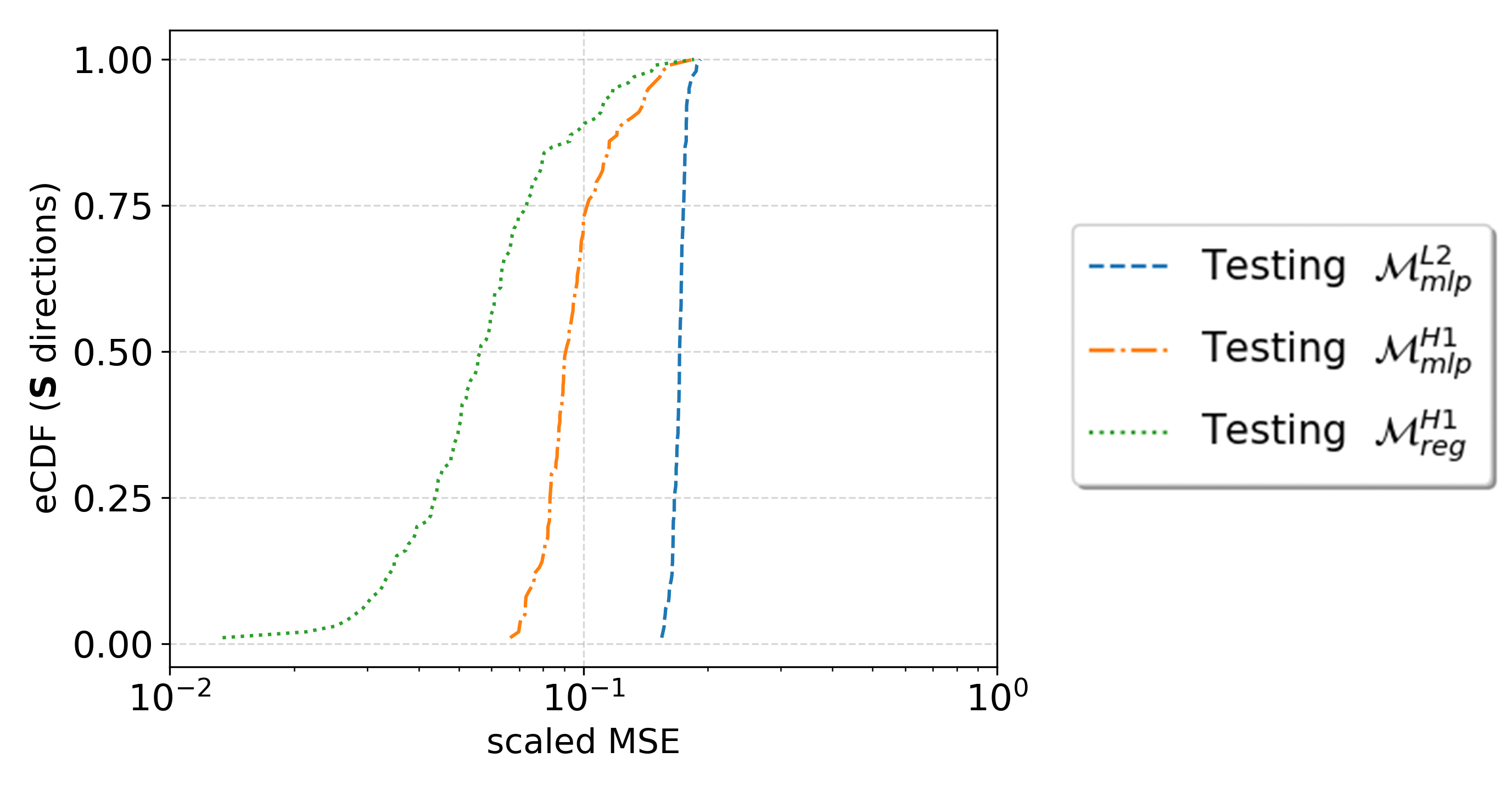}

\caption{Scaled MSE vs eCDF curves for $\psi$ energy functional (top left), second Piola-Kirchhoff stress $\tensor{S}$ tensor principal values (top right), and  second Piola-Kirchhoff stress $\tensor{S}$ tensor principal direction predictions (bottom) for the models ${\cal M}^{L2}_{mlp}$, ${\cal M}^{H1}_{mlp}$, and ${\cal M}^{H1}_{reg}$. The dataset consists of 100 polycrystal RVEs with number of crystals ranging from 40 to 50. The models' performance is tested with a K-fold algorithm - only the blind prediction results are shown.} 
\label{fig:kfold_ecdf}

\end{figure}

\begin{figure}[h!]
\centering
\includegraphics[width=7.5 cm,angle=0]{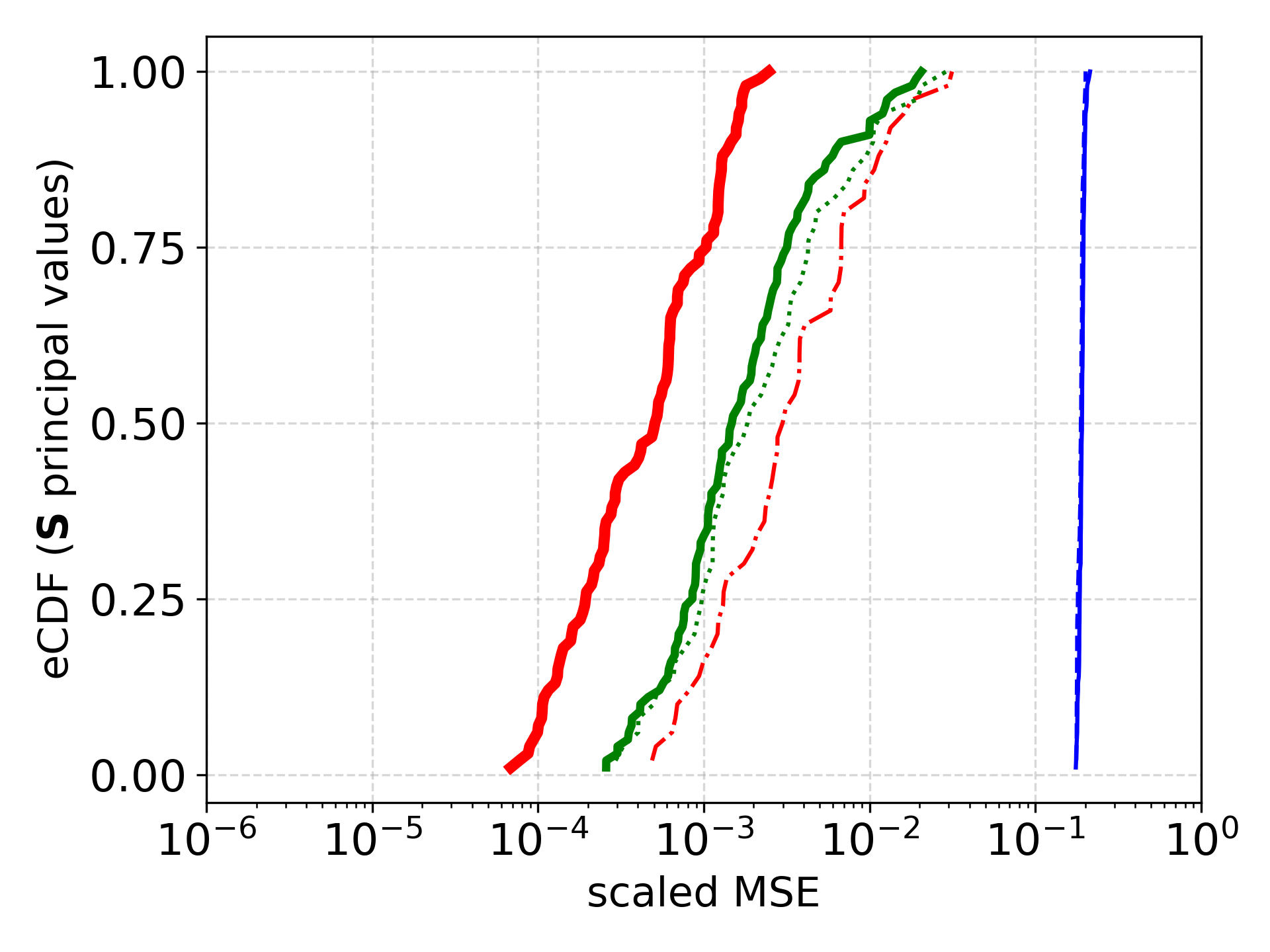}
\includegraphics[width=7.5 cm,angle=0]{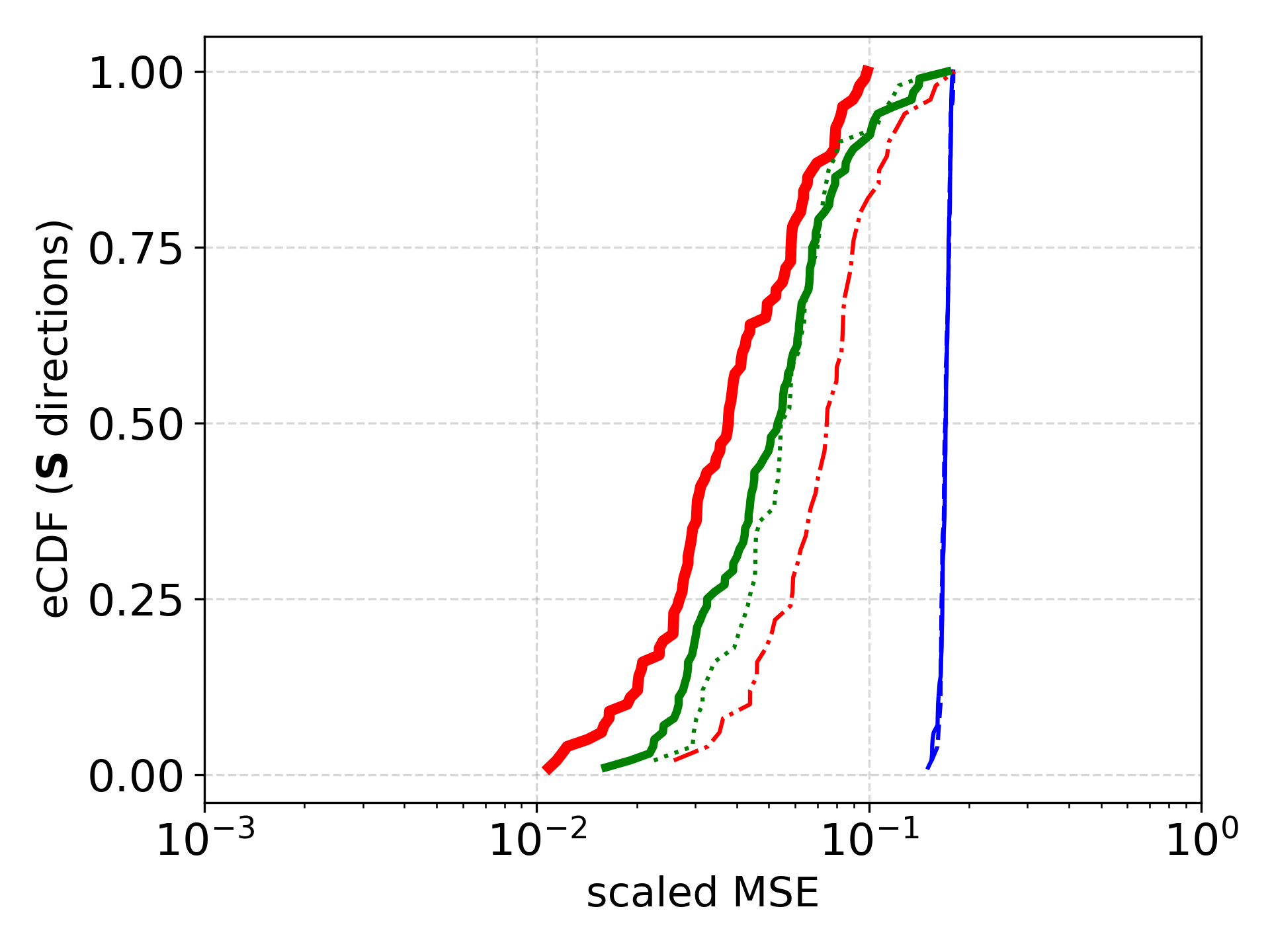}
\includegraphics[width=8 cm,angle=0]{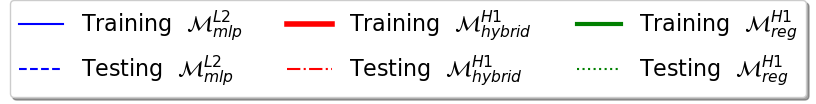}

\caption{Scaled mean squared error comparison for the models ${\cal M}^{H1}_{hybrid}$, ${\cal M}^{H1}_{reg}$, and ${\cal M}^{L2}_{mlp}$ for the second Piola - Kirchhoff stress $\tensor{S}$ tensor principal value and direction predictions. The training and testing was performed on 150 polycrystals - 100 RVEs in the training set and 50 RVEs in the testing set. While ${\cal M}^{H1}_{hybrid}$ outperforms the simple MLP model, it appears to be prone to overfitting - the training error is much lower than the blind prediction error. This issue is alleviated with regularization techniques that promote the model's robustness. This can be qualitatively seen on the scaled MSE vs eCDF plot - the distance between training and testing curves closes. } 
\label{fig:data_reg}
\end{figure}

Other than this quantitative metric, the hybrid network also appears to procure superior results qualitatively. In figure~\ref{fig:surface_estimation_2RVES}, the energy potential surface estimations are shown for the simple multilayer perceptron and the hybrid architecture for two different polycrystals. Without the graph as input the network cannot distinguish behaviors, while the hybrid architecture estimates two different energy surfaces and, thus, distinctive stress behaviors too. The weighted connectivity graph of each polycrystalline formation is encoded in a perceivably different feature vector that aids the downstream multilayer perceptron to identify and interpolate between different behaviors. For the experiment show in Fig.~\ref{fig:surface_estimation_2RVES}, the selected encoded feature vector dimension was 9. The size of the feature was chosen for providing the better results. It was seen that a small encoded feature vector (less than three features) did not procure as good results. This could be possibly interpreted as the inability to compress the connectivity graph information for the anisotropic structure in such low dimensions. It is common practice in engineering mechanics to express anisotropy with higher order measures (e.g texture tensor) and not a single feature - scalar. It was also seen than increasing the encoded feature vector dimension substantially over nine features did not procure any improvement in the prediction results.

\subsection{Verification Tests: anisotropy and convexity of the trained models}
\label{sec:verification_tests}
To ensure that the constitutive response predicted by the trained neural network are consistent
with the known mechanics principles, we subject the trained models to two numerical tests, i.e. the isotropy test and the convexity tests. A material frame indifference test was not deemed necessary, since the objectivity condition was shown to automatically be fulfilled in Section~\ref{sec:mfi}. 


\subsubsection{Isotropy}
\label{sec:isotropy_check}
Since the polycrystals we used for training are inherently anisotropic, the hyperelasticity model 
generated from the hybrid learning should not exhibit isotropic behavior. Nevertheless, 
an isotropy test is still recommended to test if the training process itself induces artificial bias and 
therefore anisotropy that is not physical. The test followed the definitions described in Section~\ref{sec:isotropy_theory}.
Indeed, rotating an RVE yields different behaviors. The different energy and stress response surfaces for an RVE, rotating along the z-axis, can be seen in Fig.~\ref{fig:failed_isotropy}.

\begin{figure}[h!]
\centering
\begin{minipage}{.5\linewidth}
  \centering
\includegraphics[width=8.0cm,angle=0]{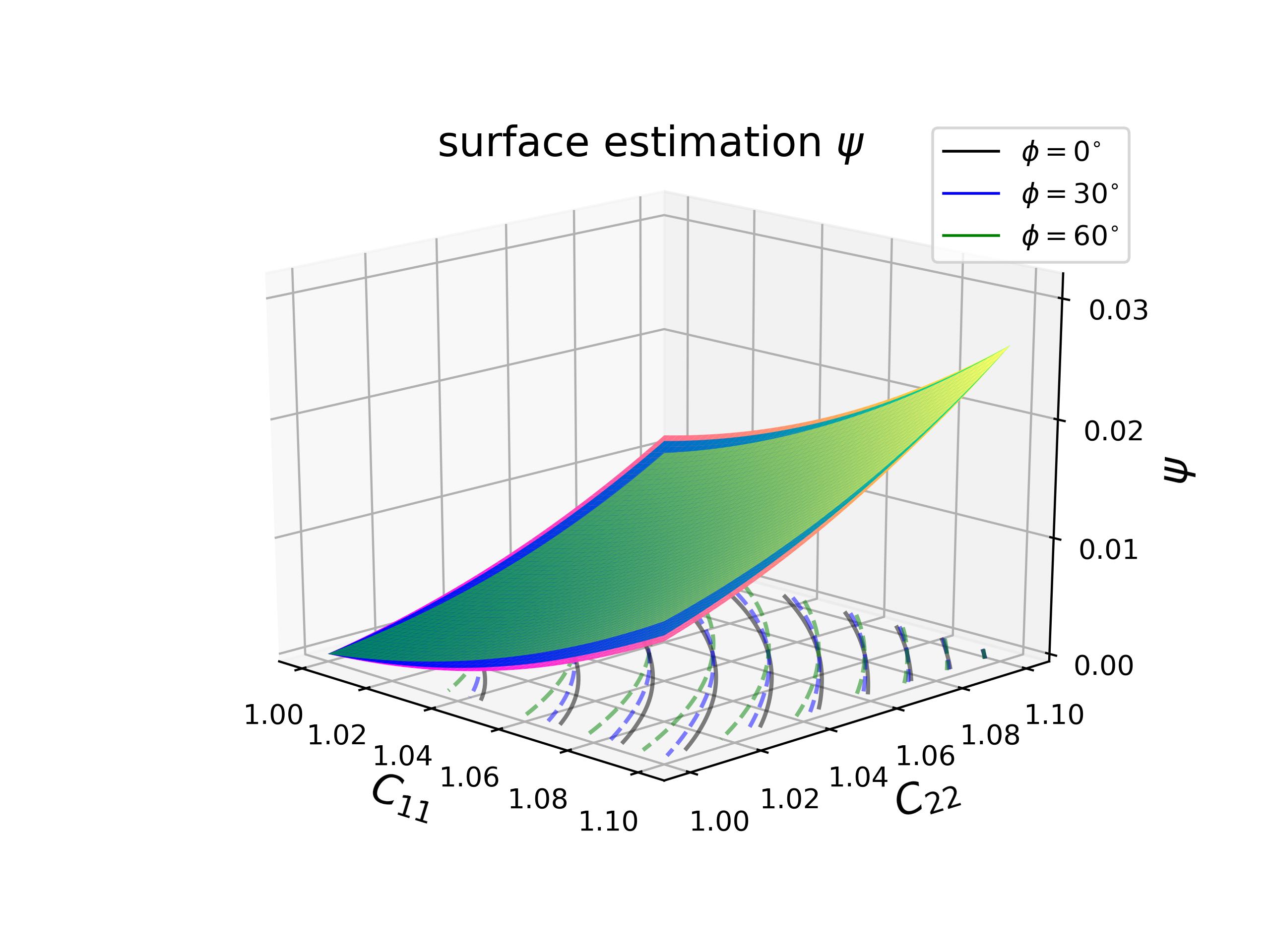}
\end{minipage}%
\begin{minipage}{.5\linewidth}
  \centering
\includegraphics[width=6.0cm,angle=0]{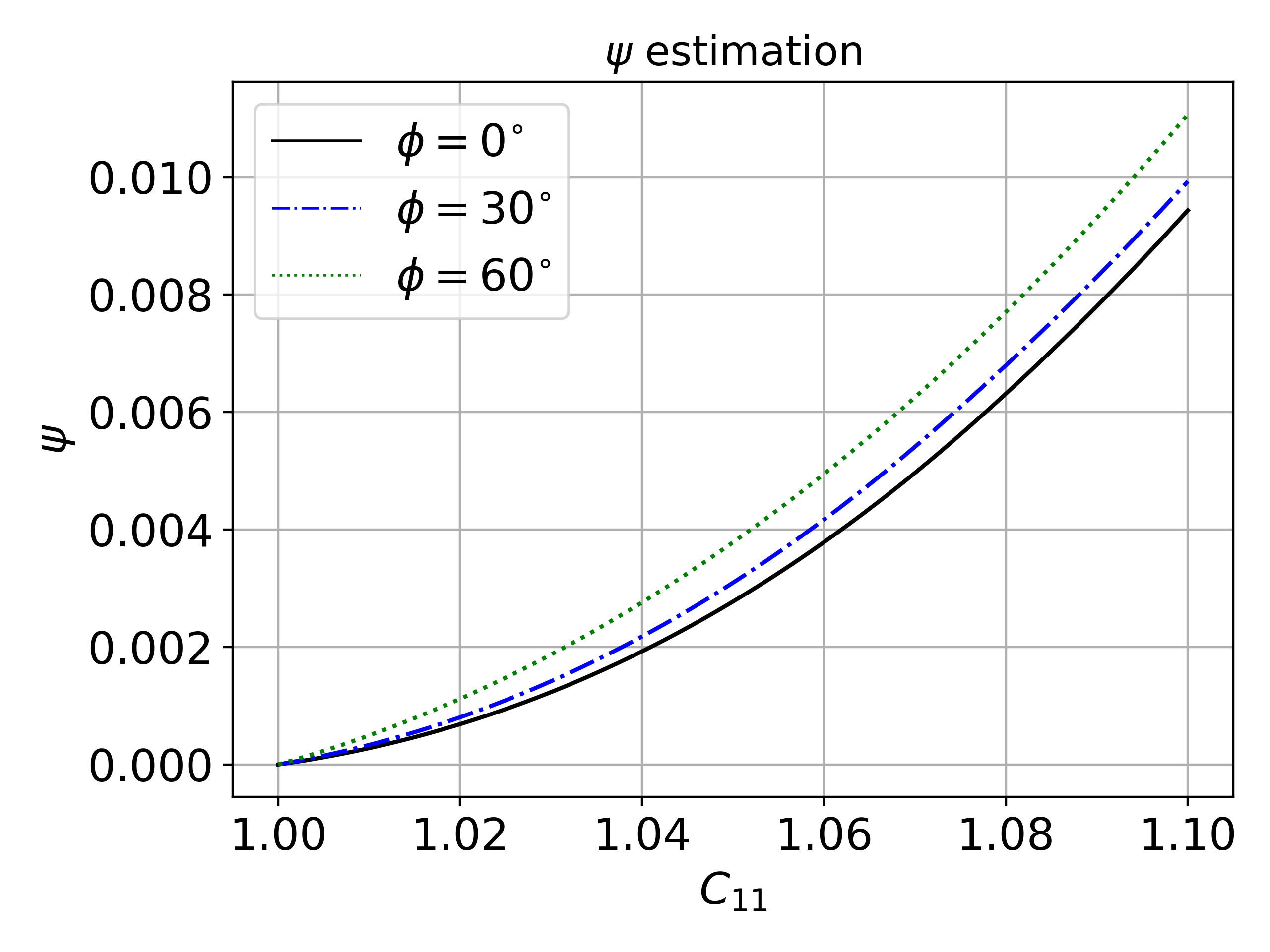}
\end{minipage}
\begin{minipage}{.5\linewidth}
  \centering
\includegraphics[width=6.0cm,angle=0]{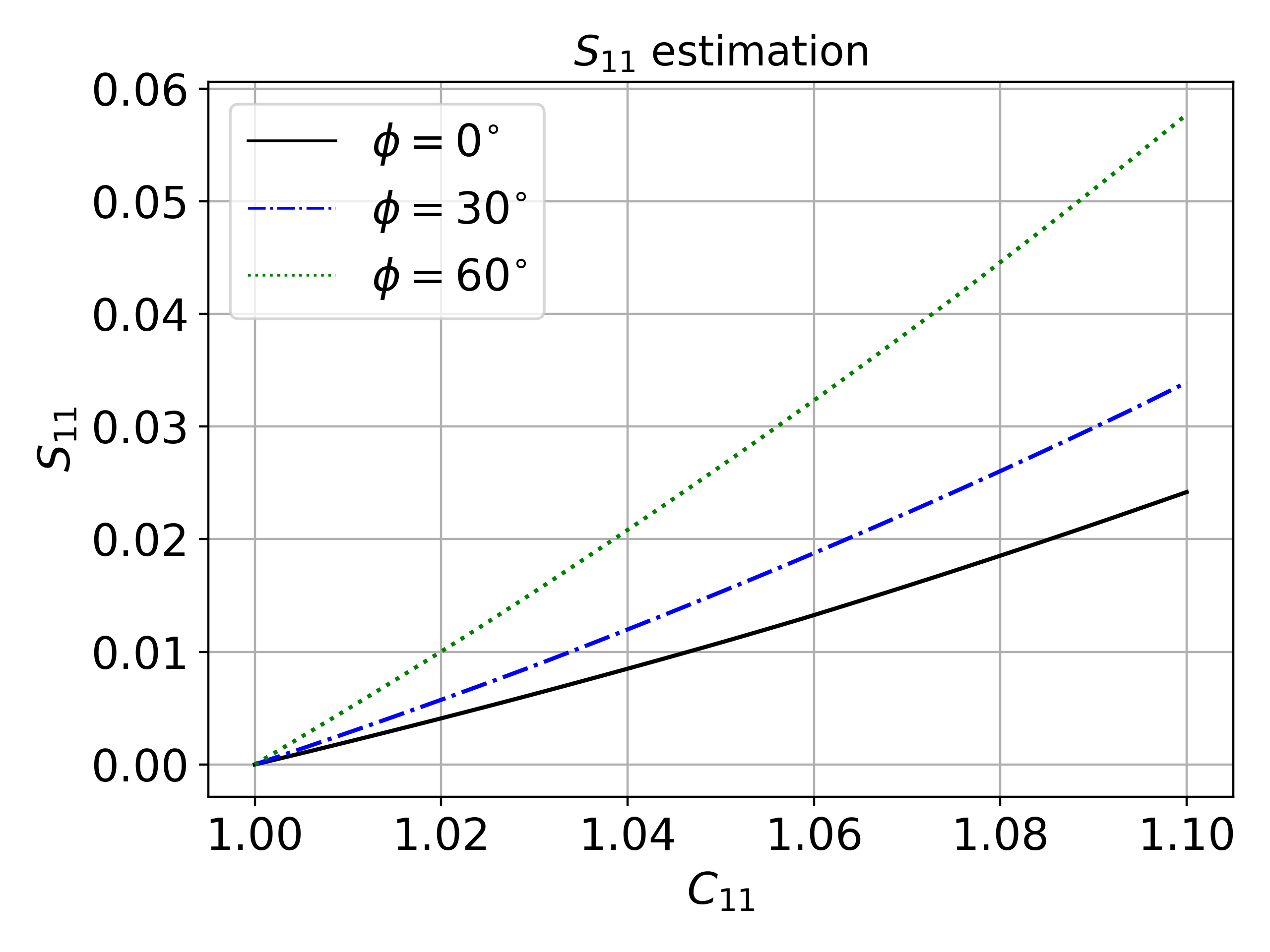}
\end{minipage}%
\begin{minipage}{.5\linewidth}
  \centering
\includegraphics[width=6.0cm,angle=0]{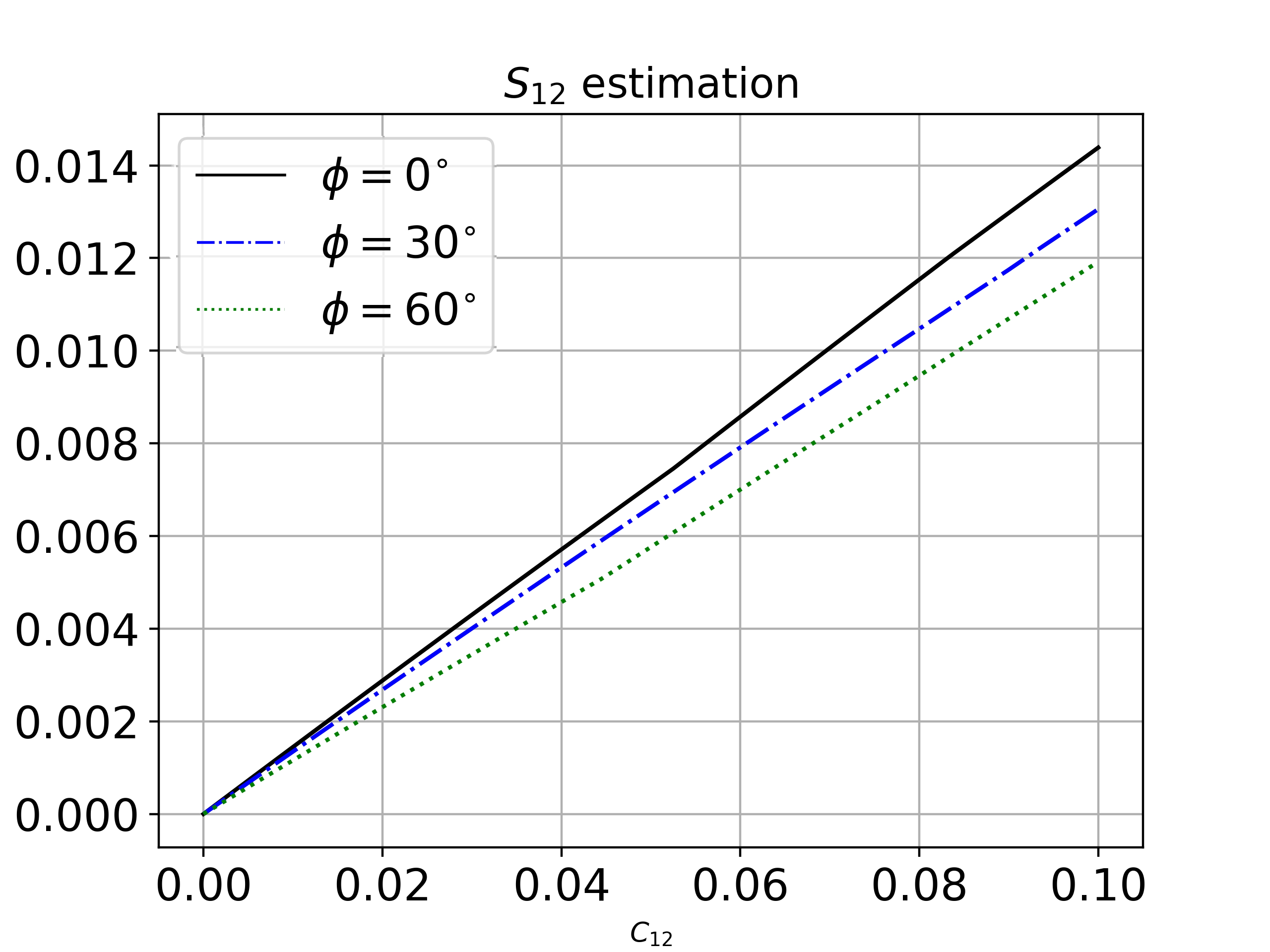}
\end{minipage}
\caption{Estimated $\psi$, $S_{11}$, and $S_{12}$ responses  for $0 ^{\circ}$, $30 ^{\circ}$, and $60 ^{\circ}$ rotations of the RVE.} 
\label{fig:failed_isotropy}
\end{figure}

\subsubsection{Convexity}
\label{sec:convexity_check}
To check the convexity for the trained hybrid models, a numerical check was conducted on the trained hybrid architecture models. The models where tested for the check described in Eq.~\ref{eq:gradient_inequality_blackbox}. The $\tensor{C}_{\alpha}$ and $\tensor{C}_{\beta}$ were chosen to be right Cauchy deformation tensors sampled from the training and testing sets of deformations. The input $\mathbb{G}$ was checked for all the 150 RVEs trained and tested on number of RVEs. For every graph input, the approximated energy functional must be convex. Thus, to verify that for all the poly-crystal formations, the convexity check is repeated for every RVE in the dataset. It is noted that, while these checks describe a necessary condition for convexity, they do not describe a sufficient condition and more robust methods of checking convexity will be considered in the future. For a specific poly-crystal formation - graph input, the network has six independent variables - deformation tensor $\tensor{C}$ components. To check the convexity, for every RVE in the dataset, deformation tensors $\tensor{C}$ are sampled in a grid and are checked pairwise (approximately 265,000 combinations of points / checks per RVE) and are found to satisfy the inequality \ref{eq:gradient_inequality_blackbox}. In Figure~\ref{fig:convexity}, a sub-sample of 100 convexity checks for three RVEs is demonstrated. 

\begin{figure}[h!]
\centering
\includegraphics[width=8.5 cm,angle=0]{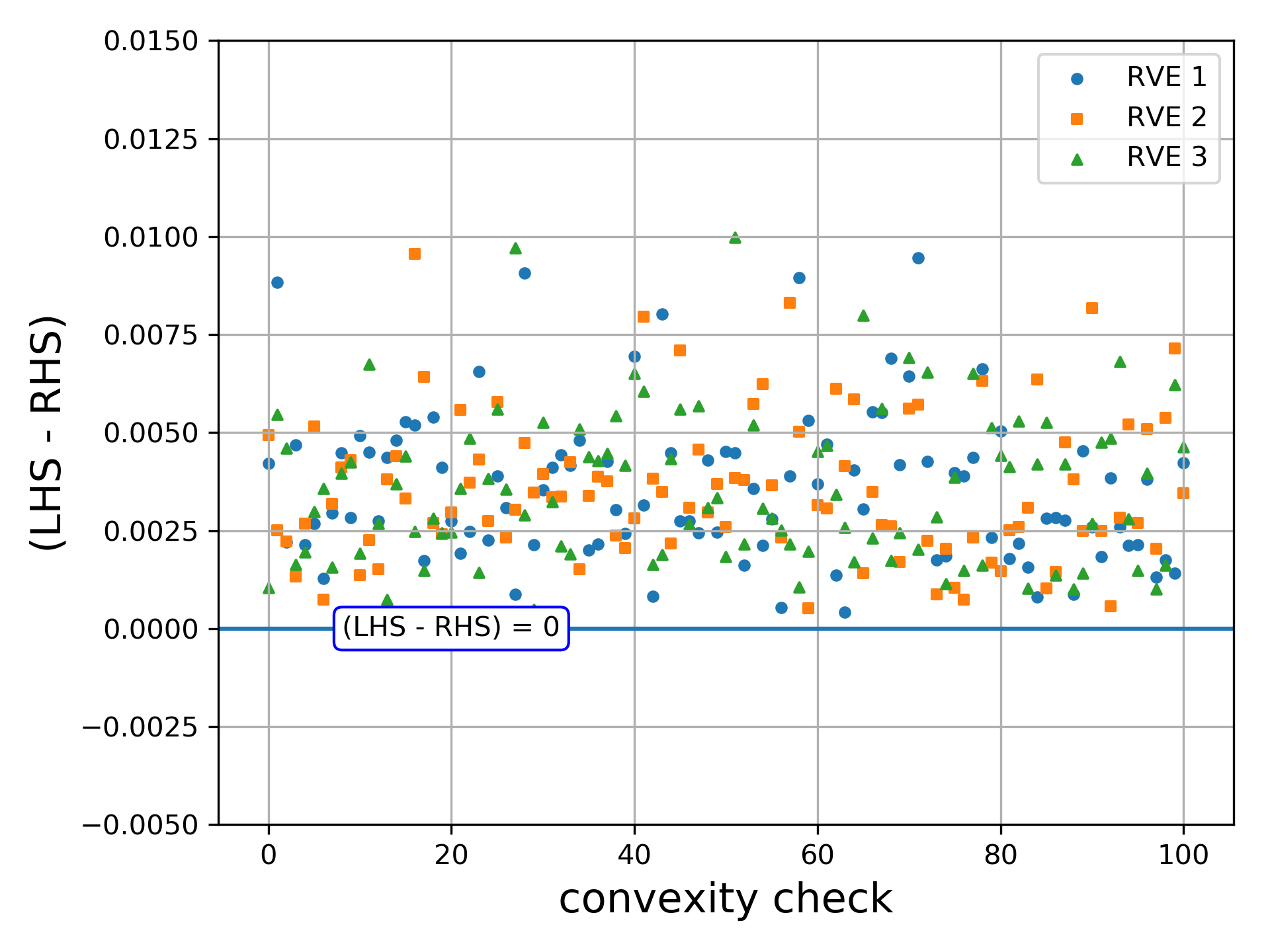}
\caption{Approximated energy functional convexity check results for three different polycrystals. Each point represents a convexity check and must be above the $\left[\text{LHS} - \text{RHS} = 0 \right]$ line so that the inequality \ref{eq:gradient_inequality_blackbox} is satisfied. } 
\label{fig:convexity}
\end{figure}

\subsection{Parametric study: Anisotropic responses of polycrystals in phase-field fracture}

The anisotropic elastic responses predicted using the hybrid neural network trained by
both non-Euclidean descriptors and FFT simulations performed on polycrystals 
are further examined in the phase field fracture simulations in which the stored energy functional generated from the hybrid learning is degraded according to a driving force.
In this series of paramatric studies, the Kalthoff-Wikler experiment is numerically simulated via a phase field model in which the elasticity is predicted by the hybrid neural network  \citep*{kalthoff1988failure,kalthoff_modes_2000}.
We adopt the effective stress theory \citep{simo1987strain} is valid such that the stored energy can be written in terms of the product of a degradation function and the stored elastic energy. The degradation function and the driving force are both pre-defined in this study. The training of incremental functional for the path-dependent constitutive responses will be considered in the second part of this series of work. 

In the first numerical experiment, we conduct a parametric study by varying the orientation of the RVE 
to analyze how the elastic anisotropy predicted by the graph-dependent energy functional affects the nucleation 
and propagation of cracks. In the second numerical experiment, the hybrid neural network is given new microstructures. Forward predictions of the elasticity of the two new RVEs are made by the hybrid neural network without further calibration. We then compare the crack patterns for the two RVEs and compare the predictions made without the graph input 
 to analyze the impact of the incorporation of non-Euclidean descriptors on the quality of predictions od crack growths. 
 
While previous work, such as \citet{kochmann2018efficient}, has utilized FFT simulations to
generate incremental constitutive updates, the efficiency of the FFT-FEM model may 
highly depends on the complexity of the microstructures and the existence of sharp gradient 
of material properties of the RVEs. 
In this work, the FFT simulations are not performed during the multiscale simulations. Instead, 
they are used as the training and validation data to generate a ML surrogate model following the treatment in  \citet{wang_multiscale_2018} and \citet{wang2019updated}.

For brevity, we omit the detailed description of the phase field model for brittle fracture. Interested readers please refer to, for instance,  \citet{bourdin2008variational} and \citet{borden2012phase}. 
In this work, we adopt the viscous regularized version of phase field brittle fracture model in \citet{miehe_phase_2010} in which the degradation function and the critical energy release rate pre-defined. 
The equations solved are the balance of linear momentum and the rate-dependent phase-field governing equation:

\begin{equation}
\label{eq:momentum_eq}
\Diver \tensor{P} + \tensor{B} = \rho \ddot{\tensor{U}},
\end{equation}

\begin{equation}
\label{eq:phase_field_equation}
\frac{g_c}{l_0} (d - l_0^2 \Diver  [ \partial_{\nabla d} \gamma  ]) + \eta \dot{d} = 2 (1 - d) \mathcal{H} ,
\end{equation}
where $\gamma$ is the crack density function that represents the diffusive fracture, i.e., 
\begin{equation}
\gamma(d, \nabla d) = \frac{1}{2 l} d^{2} + \frac{l}{2} | \nabla d |^{2}  .
\end{equation} 

The problem is solved following a standard staggered time discretization \citep{borden_phase-field_2012-1} such that
the balance of linear momentum and the phase field governing equations are updated sequentially. 
In the above Eq.~\eqref{eq:momentum_eq}, $\tensor{P}$ is the first Piola-Kirchhoff stress tensor, $\tensor{B}$ is the body force and $\ddot{\tensor{U}}$ is the second time derivative of the displacement $\tensor{U}$. In Eq.~\eqref{eq:phase_field_equation}, following \citep{miehe_phase_2010-1}, $d$ refers to the phase-field variable, with $d = 0$ signifying the undamaged and $d = 1$ the fully damaged material, while $\Delta d$ refers to the Laplacian of the phase-field. The variable $l_0$ refers to the length scale parameter used to approximate the sharp crack topology as a diffusive crack profile, such that as $l_0 \rightarrow 0$ the sharp crack is recovered. The parameter $g_c$ is the critical energy release rate from the Griffith crack theory. The parameter $\eta$ refers to an artificial viscosity term used to regularize the crack propagation by giving it a viscous resistance. The term $\mathcal{H}$ is the force driving the crack propagation and, in order to have an irreversible crack propagation in tension, it is defined as the maximum tensile ("positive") elastic energy that a material point has experienced up to the current time step $t_n$, formulated as:

\begin{equation}
\label{eq:history_variable}
\mathcal{H}(\tensor{F}_{t_n}, \mathbb{G})= \max_{t_n \leq t}\psi^{+}(\tensor{F}_{t_n}, \mathbb{G}). 
\end{equation}

The degradation of the energy due to fracture should take place only under tension and can be linked to the that of the undamaged elastic solid as:

\begin{equation}
\label{eq:energy_degradation}
\psi(\tensor{F},d, \mathbb{G}) = (g(d) + r) \psi^{+}(\tensor{F}, \mathbb{G}) + \psi^{-}(\tensor{F}, \mathbb{G}).
\end{equation}

The parameter $r$ refers to a residual energy remaining even in the full damaged material and it is set $r\approx 0$ for these experiments. For these numerical experiments, the degradation function that was used was the commonly used quadratic \citep*{miehe_phase_2010-1}:

\begin{equation}
\label{eq:degradation_function}
g(d) = (1 - d)^2 \quad \text{with} \quad g(0) = 1 \quad \text{and} \quad g(1) = 0.
\end{equation}

In order to perform a tensile-compressive split, the deformation gradient is split into a volumetric and an isochoric part. The energy and the stress response of the material should not be degraded under compression. The split of the deformation gradient, following \citep{de_souza_neto_computational_2011}, is performed as follows:

\begin{equation}
\label{eq:vol_iso_split}
\tensor{F} = \tensor{F_{\text{iso}}} \tensor{F_{\text{vol}}} = \tensor{F_{\text{vol}}} \tensor{F_{\text{iso}}},
\end{equation}

where the volumetric component of $\tensor{F}$ is defined as

\begin{equation}
\label{eq:F_vol}
\tensor{F_{\text{vol}}} = (\det\tensor{F})^{1/3} \tensor{I} ,
\end{equation}
and the volume-preserving isochoric component as

\begin{equation}
\label{eq:F_iso}
\tensor{F_{\text{iso}}} = (\det\tensor{F})^{-1/3} \tensor{F} .
\end{equation}
The strain energy is, thus, split in a "tensile" and "compressive" part, such that:

\begin{equation}
\label{eq:energy_split_pos}
\psi^{+}= \left\{
\begin{array}{ll}
      \psi(\tensor{F}, \mathbb{G}) & \quad J\geq 1 \\
      \psi(\tensor{F}, \mathbb{G}) - \psi(\tensor{F_{\text{vol}}}, \mathbb{G}) & \quad J < 1 ,\\
\end{array} 
\right. 
\end{equation}

\begin{equation}
\label{eq:energy_split_neg}
\psi^{-}= \left\{
\begin{array}{ll}
      0 & \quad J\geq 1 \\
      \psi(\tensor{F_{\text{vol}}}, \mathbb{G}) &  \quad J < 1 . \\
\end{array} 
\right. 
\end{equation}

where $J = \det(\tensor{F})$. In these examples, the energy values are calculated using the hybrid architecture neural network model, whose derivatives with respect to the strain input will be the stress. Since the model's input is in terms of the right Cauchy-Green deformation tensor, the degraded stress is calculated as:

\begin{equation}
\label{eq:degraded_stress}
\tensor{P}(\tensor{F},d,\mathbb{G}) = 2 \tensor{F}\left[ g(d)\frac{\partial \hat{\psi}^{+}(\tensor{C},\mathbb{G})}{\partial \tensor{C}} + \frac{\partial \hat{\psi}^{-}(\tensor{C},\mathbb{G})}{\partial \tensor{C}}\right].
\end{equation}

The experiment in question studies the crack propagation due to the high velocity impact of a projectile. The geometry and boundary conditions of the domain, as well as the configuration of the pre-existing crack is shown in Fig.~\ref{fig:winkler_setup}. Since the problem is symmetric, only half of the domain is studied.  \citep*{kalthoff1988failure,kalthoff_modes_2000} have observed the crack to propagate at $70 ^{\circ} $ for an isotropic material, results that have previously been reproduced with numerical simulations in other studies \citep{belytschko_dynamic_2003,song_comparative_2008,borden_phase-field_2012-1}. The experiment is conducted for two impact velocities ($v_0 = 16.5\: m /s$ and $v_0 = 33.0 \:m/s$) to test the crack branching phenomenon expected for higher impact velocities. 

\begin{figure}[h!]
\centering
\includegraphics[width=7.0cm,angle=0]{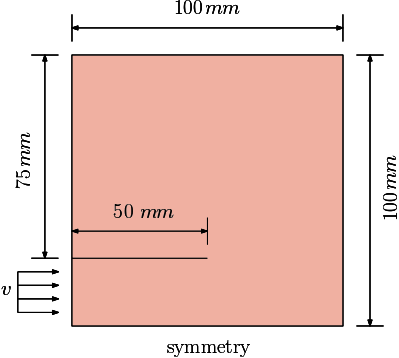}
\caption{The geometry and boundary conditions of the domain for the dynamic shear loading experiment. The velocity is prescribed progressively at the bottom left corner of the domain. The mesh is designed to have a pre-existing crack of $50.0 \: mm$. Only half the domain can be modelled due to symmetry.} 
\label{fig:winkler_setup}
\end{figure}

The experiment lasts for $80 \:\mu s$ and the prescribed velocity is applied progressively following the scheme below for $t_0 = 1 \:\mu s$:

\begin{equation}
\label{eq:velocity_application}
v= \left\{
\begin{array}{ll}
      \frac{t}{t_0} v_0 & t \leq t_0 \\
      v_0 & t > t_0. \\
\end{array} 
\right. 
\end{equation}

The domain is meshed uniformly with 20,000 triangular elements and the length scale is chosen to be $l_0 = 1.2 \times 10^{-3} \: m$. While this mesh is rather coarse compared to previous studies of the same problem, it was deemed adequate to simulate the problem at hand with acceptable accuracy to qualitatively demonstrate the anisotropic model behavior. The time-step used for the explicit method was chosen to be slightly smaller than the critical one ($\Delta t = 5 \times 10 ^{-8} \: s$). Changing the time step of the explicit method did not appear to affect the phase-field solution, as long as the explicit solver for the momentum equation was stable. 

\begin{figure}[h!]
\newcommand\siz{3cm}
\centering
\begin{tabular}{M{3.5cm}M{3.5cm}M{3.5cm}}
\includegraphics[width=2.5cm ,angle=0]{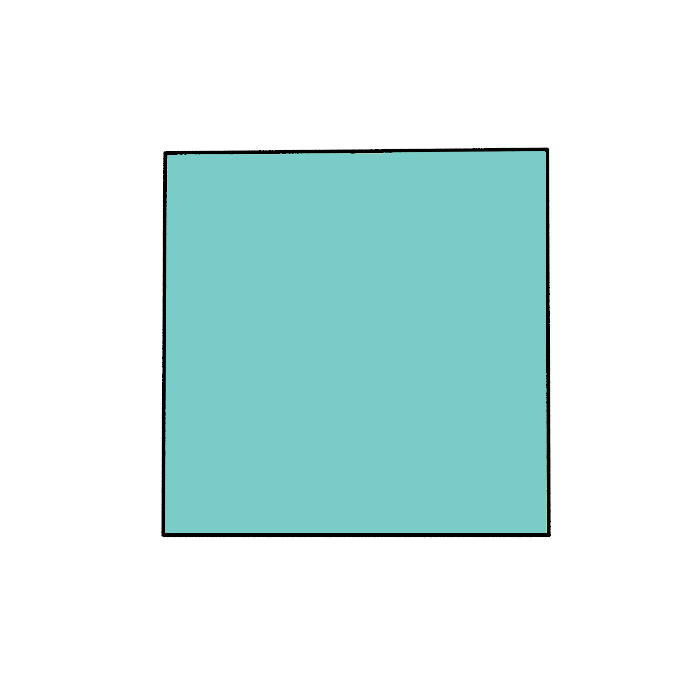} &
\includegraphics[width=3cm ,angle=0]{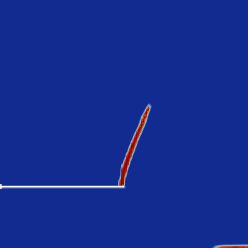} &   
\includegraphics[width=3cm,angle=0]{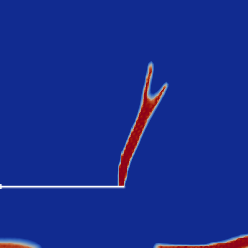} \\

isotropic & (a) & (b)  \\[6pt]
\includegraphics[width=2.5cm ,angle=0]{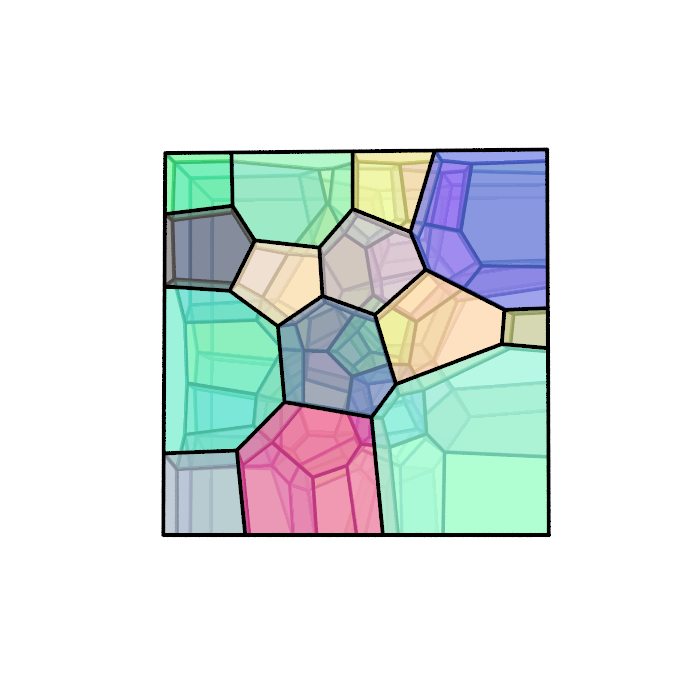} &
\includegraphics[width=3cm,angle=0]{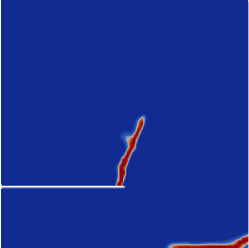} &   
\includegraphics[width=3cm,angle=0]{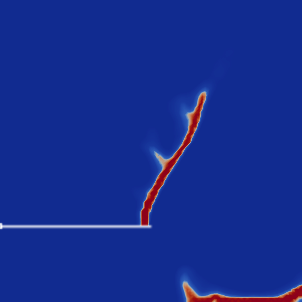} \\

$\phi =0 ^{\circ}$& (c)  & (d) \\[6pt]
\includegraphics[width=2.5cm ,angle=0]{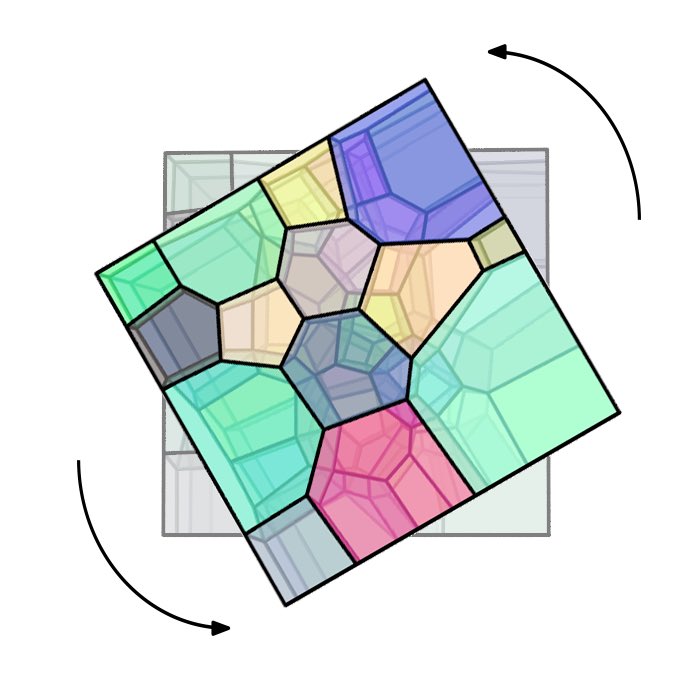} &
\includegraphics[width=3cm ,angle=0]{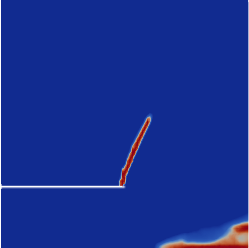} &   
\includegraphics[width=3cm,angle=0]{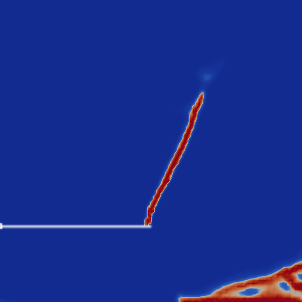} \\

$\phi =30 ^{\circ}$ & (e)  & (f)  \\[6pt]
\includegraphics[width=2.5cm ,angle=0]{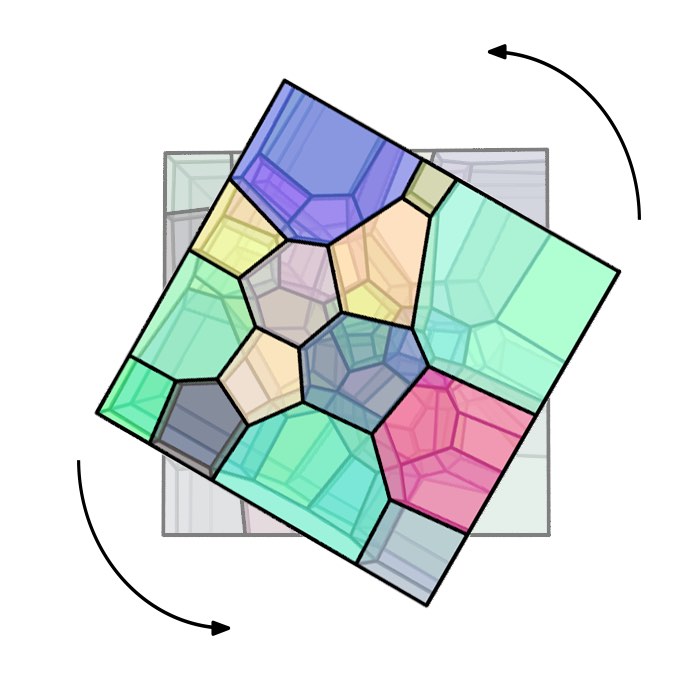} &
\includegraphics[width=3cm ,angle=0]{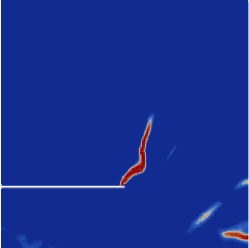} &   
\includegraphics[width=3cm,angle=0]{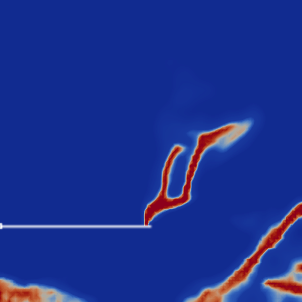} \\

$\phi =60 ^{\circ}$ & (g)  & (h)   \\[6pt]

\end{tabular}

\caption{Crack patterns at $65 \:\mu s$ for the dynamic shear loading experiment for the isotropic material and the anisotropic material for a constant graph, rotated at various angles. The right column shows the experiments for $v = 16.5\: m /s$ and the left column for $v = 33.0\: m /s$.}
\label{fig:anisotropy_1RVE}

\end{figure}

For the first numerical experiment, the behavior of a single polycrystal was tested. In other words, the graph input of the hybrid architecture material model remained constant for all the simulations. The material model is a trained neural network of type ${\cal M}^{H1}_{reg}$ with the graph input set constant. The purpose of this experiment is to show that by rotating the highly anisotropic RVE, under the same boundary conditions, different wave propagation and crack nucleation patterns can be observed. This experiment could be paralleled to rotating a transversely isotropic material - different fiber orienations should procure different results under identical boundary conditions. 

The experiment is first conducted on an isotropic material with equivalent parameters to verify the formulation and compare with the anisotropic results. It can be seen that with the current formulation, the isotropic model can recover the approximately $70 ^{\circ} $ angle previously reported in the experiments and numerical simulations. In Fig.~\ref{fig:anisotropy_1RVE}, it is demonstrated that the neural network material model is indeed anisotropic, showing varying behaviors while rotating the RVE for  $0 ^{\circ} $, $30 ^{\circ} $, and $60 ^{\circ} $. The nature of the anisotropy becomes more apparent when the impact velocity is doubled and the crack branching is more prevalent. While the behavior of the material under rotation is not clear to interpret, it could be justified by observing how the enrgy response surface changes while rotating the RVE. The different energy responses for the RVE in question can be seen in Fig.~\ref{fig:failed_isotropy}. As the RVE is increasingly rotated, the incline of the response surface increases - for the same deformation, more energy is stored for larger rotations. Thus, for the same impact velocity, more energy is gathered along the crack for larger rotations and the branching happens sooner - as it appears to be happening for the case of the $60 ^{\circ}$ rotation. It is noted that this is an intuitive interpretation valid for the current polycrystal - other formations could behave differently under rotation.

\begin{figure}[h!]

\newcommand\siz{2.3cm}
\centering
\begin{tabular}{M{2.5cm}M{2.5cm}M{2.5cm}M{2.5cm}M{2.5cm}}

\includegraphics[width=1.cm ,angle=0]{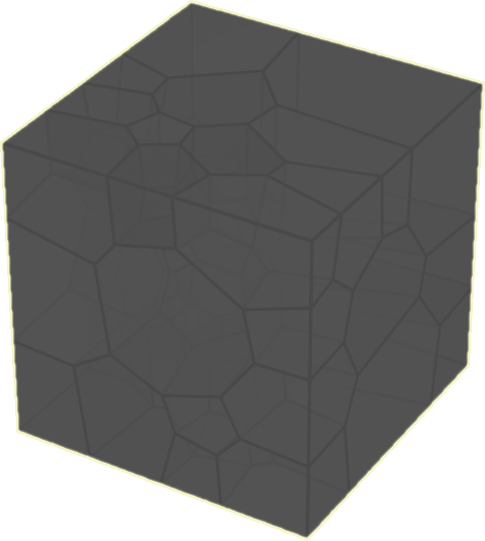} &
\includegraphics[width=2.3cm,angle=0]{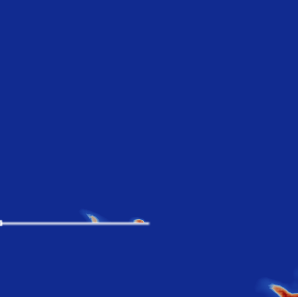} &
\includegraphics[width=2.3cm ,angle=0]{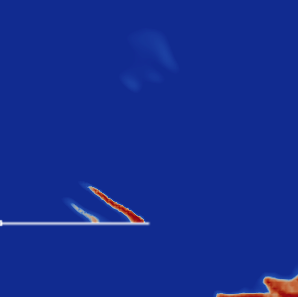} &
\includegraphics[width=2.3cm,angle=0]{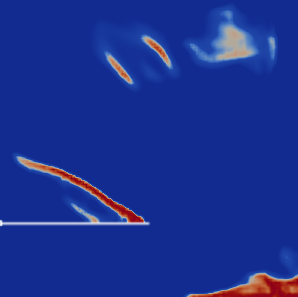} &
\includegraphics[width=2.3cm ,angle=0]{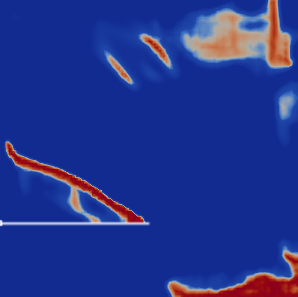} \\

no graph & (a) & (b) & (c) & (d)\\[6pt]

\includegraphics[width=1.cm ,angle=0]{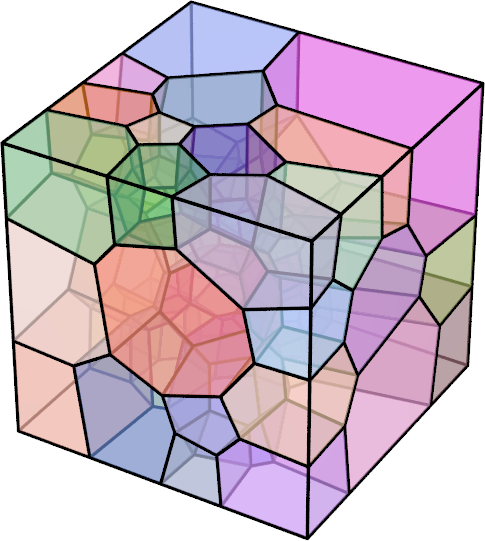} &
\includegraphics[width=2.3cm ,angle=0]{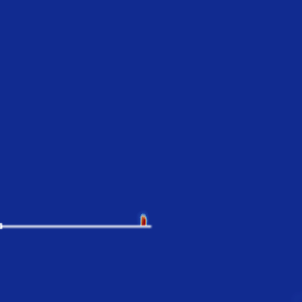} &
\includegraphics[width=2.3cm,angle=0]{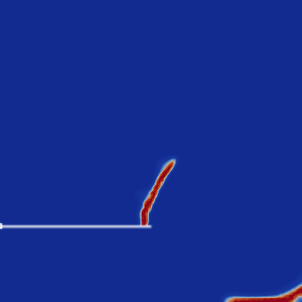} &
\includegraphics[width=2.3cm ,angle=0]{./figure/0_2x_f.png} &
\includegraphics[width=2.3cm,angle=0]{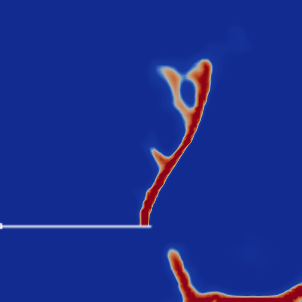} \\

RVE A & (e) & (f) & (g) & (h) \\[6pt]
\includegraphics[width=1.cm ,angle=0]{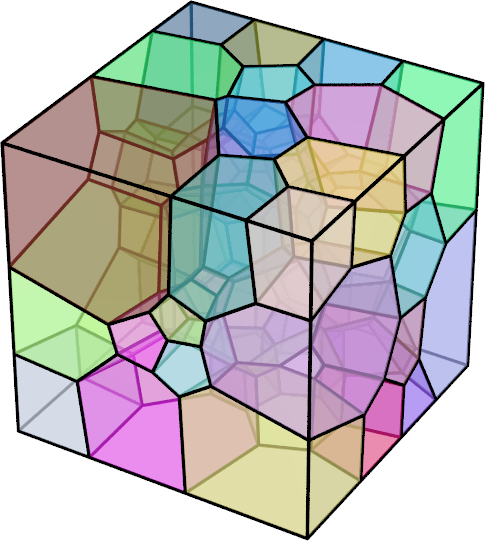} &
\includegraphics[width=2.3cm,angle=0]{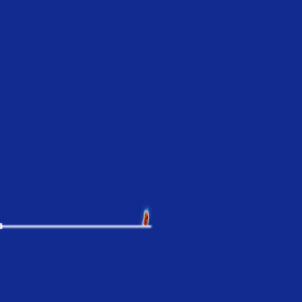} &
\includegraphics[width=2.3cm,angle=0]{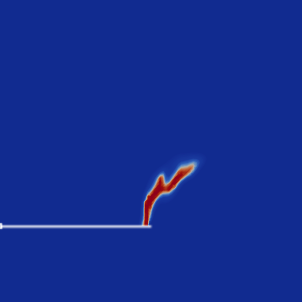} &
\includegraphics[width=2.3cm,angle=0]{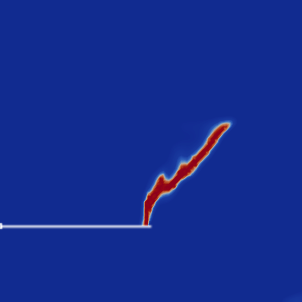} &
\includegraphics[width=2.3cm,angle=0]{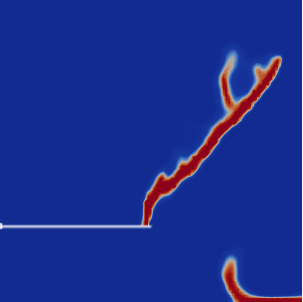} \\

RVE B & (i)  & (j) & (k) & (l)\\[6pt]

\end{tabular}

\caption{Crack patterns at $30 \:\mu s$, $50 \:\mu s$, $65 \:\mu s$, $85 \:\mu s$ for the dynamic shear loading experiment with an impact velocity of $v = 33.0\: m /s$ for a model without a graph input (a, b, c, d) and two different polycrystals (e, f, g, h and i, j, k, l). It is noted that all the parameters are identical for both simulations but the graph input.}
\label{fig:anisotropy_2RVEs}

\end{figure}

For the second numerical experiment, the material response was tested for different polycrystals (model type ${\cal M}^{H1}_{reg}$) as well as for a model without any graph inputs (type ${\cal M}^{H1}_{mlp}$) . The aim of this experiment was to verify that the hybrid architecture and the graph input can capture the anisotropy of the polycrystal material that is originating from the interactions between crystals, as expressed by the connectivity graph. The above experiment was repeated for different graph inputs and the results are demonstrated in Fig~\ref{fig:anisotropy_2RVEs}. In the absence of a graph input, while there is crack propagation the results look noisy and the direction of the propagation is not similar to that of specific RVEs, something that could be potentially attributed to the model being trained on multiple polycrystal behaviors. For the model with the graph input, the difference in behaviors appears to become more apparent in the areas where branching is more prevalent, with the polycrystal affecting the crack branching phenomena. No additional anisotropy measures or crack branching criteria were utilized for these simulations. The sole additional information in the input of the material model would be the weighted connectivity graph.

\section{Conclusion}

We have designed a hybrid neural network that predicts the elastic stored energy functional for Green-elastic materials. 
By utilizing non-Euclidean data represented by weighted graphs, we introduce these graphs as 
new descriptors for geometric learning such that the hybrid deep learning can produce an energy functional that leverages the rich 
micro-structural information not describable by the classical Euclidean descriptors such as porosity and density. 
To overcome the spurious oscillations of the derivative often occurs due to lack of regularization or overfitting, we adopt the 
Sobolev tranning such that the resultant hyperelastic energy functional does not exhibit spurious oscillations in the parametric space. 
This work opens new doors for creating new constitutive models with non-Euclidean data. Compared to the hybrid neural network 
that combined unsupervised learning of images and supervised learning, the graph-based approach requires less data while the added value of the graph input on the forward predictions is clearly shown in the k-fold validation.

\section{Acknowledgments}
The authors are supported by 
by the NSF CAREER grant from Mechanics of Materials and Structures program
at National Science Foundation under grant contract CMMI-1846875, the Dynamic Materials 
and Interactions Program from the Air Force Office of Scientific 
Research under grant contracts FA9550-17-1-0169 and FA9550-19-1-0318.
These supports are gratefully acknowledged. 
The views and conclusions contained in this document are those of the authors, 
and should not be interpreted as representing the official policies, either expressed or implied, 
of the sponsors, including the Army Research Laboratory or the U.S. Government. 
The U.S. Government is authorized to reproduce and distribute reprints for 
Government purposes notwithstanding any copyright notation herein.

\bibliographystyle{plainnat}
\bibliography{main}

\begin{thebibliography}{89}
\providecommand{\natexlab}[1]{#1}
\providecommand{\url}[1]{\texttt{#1}}
\expandafter\ifx\csname urlstyle\endcsname\relax
  \providecommand{\doi}[1]{doi: #1}\else
  \providecommand{\doi}{doi: \begingroup \urlstyle{rm}\Url}\fi

\bibitem[Anand and Kothari(1996)]{anand_computational_1996}
L~Anand and M~Kothari.
\newblock A computational procedure for rate-independent crystal plasticity.
\newblock \emph{Journal of the Mechanics and Physics of Solids}, 44\penalty0
  (4):\penalty0 525--558, 1996.

\bibitem[Bachmann et~al.(2010)Bachmann, Hielscher, and
  Schaeben]{bachmann_texture_2010}
F.~Bachmann, Ralf Hielscher, and Helmut Schaeben.
\newblock \emph{Texture {Analysis} with {MTEX} -- {Free} and {Open} {Source}
  {Software} {Toolbox}}.
\newblock 2010.
\newblock \doi{10.4028/www.scientific.net/SSP.160.63}.

\bibitem[Bang-Jensen and Gutin(2008)]{bang2008digraphs}
J{\o}rgen Bang-Jensen and Gregory~Z Gutin.
\newblock \emph{Digraphs: theory, algorithms and applications}.
\newblock Springer Science \& Business Media, 2008.

\bibitem[Belytschko et~al.(2003)Belytschko, Chen, Xu, and
  Zi]{belytschko_dynamic_2003}
Ted Belytschko, Hao Chen, Jingxiao Xu, and Goangseup Zi.
\newblock Dynamic crack propagation based on loss of hyperbolicity and a new
  discontinuous enrichment.
\newblock \emph{International Journal for Numerical Methods in Engineering},
  58\penalty0 (12):\penalty0 1873--1905, 2003.
\newblock ISSN 1097-0207.
\newblock \doi{10.1002/nme.941}.

\bibitem[Bengio and Grandvalet(2004)]{bengio2004no}
Yoshua Bengio and Yves Grandvalet.
\newblock No unbiased estimator of the variance of k-fold cross-validation.
\newblock \emph{Journal of machine learning research}, 5\penalty0
  (Sep):\penalty0 1089--1105, 2004.

\bibitem[Bengio et~al.(2013)Bengio, Courville, and
  Vincent]{bengio2013representation}
Yoshua Bengio, Aaron Courville, and Pascal Vincent.
\newblock Representation learning: A review and new perspectives.
\newblock \emph{IEEE transactions on pattern analysis and machine
  intelligence}, 35\penalty0 (8):\penalty0 1798--1828, 2013.

\bibitem[Bessa et~al.(2017)Bessa, Bostanabad, Liu, Hu, Apley, Brinson, Chen,
  and Liu]{bessa2017framework}
MA~Bessa, R~Bostanabad, Z~Liu, A~Hu, Daniel~W Apley, C~Brinson, Wei Chen, and
  Wing~Kam Liu.
\newblock A framework for data-driven analysis of materials under uncertainty:
  Countering the curse of dimensionality.
\newblock \emph{Computer Methods in Applied Mechanics and Engineering},
  320:\penalty0 633--667, 2017.

\bibitem[Borden et~al.(2012{\natexlab{a}})Borden, Verhoosel, Scott, Hughes, and
  Landis]{borden2012phase}
Michael~J Borden, Clemens~V Verhoosel, Michael~A Scott, Thomas~JR Hughes, and
  Chad~M Landis.
\newblock A phase-field description of dynamic brittle fracture.
\newblock \emph{Computer Methods in Applied Mechanics and Engineering},
  217:\penalty0 77--95, 2012{\natexlab{a}}.

\bibitem[Borden et~al.(2012{\natexlab{b}})Borden, Verhoosel, Scott, Hughes, and
  Landis]{borden_phase-field_2012-1}
M.J. Borden, C.V. Verhoosel, M.A. Scott, T.J.R. Hughes, and C.M. Landis.
\newblock A phase-field description of dynamic brittle fracture.
\newblock \emph{Computer Methods in Applied Mechanics and Engineering},
  217-220:\penalty0 77--95, 2012{\natexlab{b}}.
\newblock ISSN 00457825.
\newblock \doi{10.1016/j.cma.2012.01.008}.

\bibitem[Borja(2013)]{borja_plasticity_2013}
Ronaldo~I Borja.
\newblock \emph{Plasticity}.
\newblock Springer Berlin Heidelberg, Berlin, Heidelberg, 2013.
\newblock ISBN 978-3-642-38546-9.
\newblock \doi{10.1007/978-3-642-38547-6}.

\bibitem[Borja and Lee(1990)]{borja_cam-clay_1990}
Ronaldo~I Borja and Seung~R Lee.
\newblock Cam-clay plasticity, part 1: implicit integration of elasto-plastic
  constitutive relations.
\newblock \emph{Computer Methods in Applied Mechanics and Engineering},
  78\penalty0 (1):\penalty0 49--72, 1990.

\bibitem[Borja et~al.(1997)Borja, Tamagnini, and Amorosi]{borja1997coupling}
Ronaldo~I Borja, Claudio Tamagnini, and Angelo Amorosi.
\newblock Coupling plasticity and energy-conserving elasticity models for
  clays.
\newblock \emph{Journal of geotechnical and geoenvironmental engineering},
  123\penalty0 (10):\penalty0 948--957, 1997.

\bibitem[Bourdin et~al.(2008)Bourdin, Francfort, and
  Marigo]{bourdin2008variational}
Blaise Bourdin, Gilles~A Francfort, and Jean-Jacques Marigo.
\newblock The variational approach to fracture.
\newblock \emph{Journal of elasticity}, 91\penalty0 (1-3):\penalty0 5--148,
  2008.

\bibitem[Chollet et~al.(2015)]{chollet2015keras}
Fran\c{c}ois Chollet et~al.
\newblock Keras.
\newblock \url{https://keras.io}, 2015.

\bibitem[Czarnecki et~al.(2017)Czarnecki, Osindero, Jaderberg, Swirszcz, and
  Pascanu]{czarnecki_sobolev_2017}
Wojciech~M Czarnecki, Simon Osindero, Max Jaderberg, Grzegorz Swirszcz, and
  Razvan Pascanu.
\newblock Sobolev training for neural networks.
\newblock In \emph{Advances in {Neural} {Information} {Processing} {Systems}},
  pages 4278--4287, 2017.

\bibitem[de~Souza~Neto et~al.(2011)de~Souza~Neto, Peric, and
  Owen]{de_souza_neto_computational_2011}
Eduardo~A de~Souza~Neto, Djordje Peric, and David~RJ Owen.
\newblock \emph{Computational methods for plasticity: theory and applications}.
\newblock John Wiley \& Sons, 2011.

\bibitem[Defferrard et~al.(2016)Defferrard, Bresson, and
  Vandergheynst]{defferrard_convolutional_2016}
Micha{\"e}l Defferrard, Xavier Bresson, and Pierre Vandergheynst.
\newblock Convolutional {Neural} {Networks} on {Graphs} with {Fast} {Localized}
  {Spectral} {Filtering}.
\newblock In D.~D. Lee, M.~Sugiyama, U.~V. Luxburg, I.~Guyon, and R.~Garnett,
  editors, \emph{Advances in {Neural} {Information} {Processing} {Systems} 29},
  pages 3844--3852. Curran Associates, Inc., 2016.

\bibitem[Eggersmann et~al.(2019)Eggersmann, Kirchdoerfer, Reese, Stainier, and
  Ortiz]{eggersmann2019model}
Robert Eggersmann, Trenton Kirchdoerfer, Stefanie Reese, Laurent Stainier, and
  Michael Ortiz.
\newblock Model-free data-driven inelasticity.
\newblock \emph{Computer Methods in Applied Mechanics and Engineering},
  350:\penalty0 81--99, 2019.

\bibitem[Frankel et~al.(2019)Frankel, Jones, Alleman, and
  Templeton]{frankel_predicting_2019}
A.~L. Frankel, R.~E. Jones, C.~Alleman, and J.~A. Templeton.
\newblock Predicting the mechanical response of oligocrystals with deep
  learning.
\newblock \emph{Computational Materials Science}, 169:\penalty0 109099,
  November 2019.
\newblock ISSN 0927-0256.
\newblock \doi{10.1016/j.commatsci.2019.109099}.

\bibitem[Fung(1965)]{fung1965foundations}
Yuan-cheng Fung.
\newblock Foundations of solid mechanics.
\newblock 1965.

\bibitem[Gentle(2009)]{Gentle2009}
J.E. Gentle.
\newblock \emph{Computational Statistics}.
\newblock Springer. ISBN 978-0-387-98145-1, 2009.

\bibitem[Ghaboussi et~al.(1991)Ghaboussi, Garrett~Jr, and
  Wu]{ghaboussi_knowledge-based_1991}
J~Ghaboussi, JH~Garrett~Jr, and Xiping Wu.
\newblock Knowledge-based modeling of material behavior with neural networks.
\newblock \emph{Journal of engineering mechanics}, 117\penalty0 (1):\penalty0
  132--153, 1991.

\bibitem[Goodfellow et~al.(2016)Goodfellow, Bengio, and
  Courville]{Goodfellow-et-al-2016}
Ian Goodfellow, Yoshua Bengio, and Aaron Courville.
\newblock \emph{Deep Learning}.
\newblock MIT Press, 2016.

\bibitem[Graham et~al.(1989)Graham, Knuth, Patashnik, and
  Liu]{graham1989concrete}
Ronald~L Graham, Donald~E Knuth, Oren Patashnik, and Stanley Liu.
\newblock Concrete mathematics: a foundation for computer science.
\newblock \emph{Computers in Physics}, 3\penalty0 (5):\penalty0 106--107, 1989.

\bibitem[Grattarola(2019)]{noauthor_spektral_nodate}
Daniele Grattarola.
\newblock \emph{Spektral}.
\newblock 2019.
\newblock URL \url{https://danielegrattarola.github.io/spektral/}.

\bibitem[Grover and Leskovec(2016)]{grover_node2vec:_2016}
Aditya Grover and Jure Leskovec.
\newblock node2vec: {Scalable} {Feature} {Learning} for {Networks}.
\newblock \emph{arXiv:1607.00653 [cs, stat]}, July 2016.

\bibitem[Gurson(1977)]{gurson_continuum_1977}
Arthur~L Gurson.
\newblock Continuum theory of ductile rupture by void nucleation and growth:
  {Part} {I}---{Yield} criteria and flow rules for porous ductile media.
\newblock \emph{Journal of engineering materials and technology}, 99\penalty0
  (1):\penalty0 2--15, 1977.

\bibitem[He and Chen(2019)]{he2019physics}
Qizhi He and Jiun-Shyan Chen.
\newblock A physics-constrained data-driven approach based on locally convex
  reconstruction for noisy database.
\newblock \emph{arXiv preprint arXiv:1907.12651}, 2019.

\bibitem[Heider and Sun(2019)]{heider_invariance_2019}
Yousef Heider and WaiChing Sun.
\newblock So(3)-invariance of graph-based deep neural network for anisotropic
  elastoplastic materials.
\newblock \emph{Computer Methods in Applied Mechanics and Engineering}, 2019.
\newblock tentatively accepted.

\bibitem[Holzapfel et~al.(2000)Holzapfel, Gasser, and
  Ogden]{holzapfel_new_2000}
Gerhard~A. Holzapfel, Thomas~C. Gasser, and Ray~W. Ogden.
\newblock A {New} {Constitutive} {Framework} for {Arterial} {Wall} {Mechanics}
  and a {Comparative} {Study} of {Material} {Models}.
\newblock \emph{Journal of elasticity and the physical science of solids},
  61\penalty0 (1):\penalty0 1--48, July 2000.
\newblock ISSN 1573-2681.
\newblock \doi{10.1023/A:1010835316564}.

\bibitem[Huang et~al.(2019)Huang, Xu, Farhat, and Darve]{huang_predictive_2019}
Daniel~Z Huang, Kailai Xu, Charbel Farhat, and Eric Darve.
\newblock Predictive {Modeling} with {Learned} {Constitutive} {Laws} from
  {Indirect} {Observations}.
\newblock \emph{arXiv preprint arXiv:1905.12530}, 2019.

\bibitem[Jaquet et~al.(2013)Jaquet, And{\'o}, Viggiani, and
  Talbot]{jaquet2013estimation}
Clara Jaquet, Edward And{\'o}, Gioacchino Viggiani, and Hugues Talbot.
\newblock Estimation of separating planes between touching 3d objects using
  power watershed.
\newblock In \emph{International Symposium on Mathematical Morphology and Its
  Applications to Signal and Image Processing}, pages 452--463. Springer, 2013.

\bibitem[Jerphagnon et~al.(1978)Jerphagnon, Chemla, and
  Bonneville]{jerphagnon_description_1978}
Jean Jerphagnon, Daniel Chemla, and R~Bonneville.
\newblock The description of the physical properties of condensed matter using
  irreducible tensors.
\newblock \emph{Advances in Physics}, 27\penalty0 (4):\penalty0 609--650, 1978.

\bibitem[Jones et~al.(2018)Jones, Templeton, Sanders, and
  Ostien]{jones2018machine}
Reese~E Jones, Jeremy~A Templeton, Clay~M Sanders, and Jakob~T Ostien.
\newblock Machine learning models of plastic flow based on representation
  theory.
\newblock \emph{arXiv preprint arXiv:1809.00267}, 2018.

\bibitem[Kalthoff and Winkler(1988)]{kalthoff1988failure}
JF~Kalthoff and S~Winkler.
\newblock Failure mode transition at high rates of shear loading.
\newblock \emph{DGM Informationsgesellschaft mbH, Impact Loading and Dynamic
  Behavior of Materials}, 1:\penalty0 185--195, 1988.

\bibitem[Kalthoff(2000)]{kalthoff_modes_2000}
Joerg~F. Kalthoff.
\newblock Modes of dynamic shear failure in solids.
\newblock \emph{International Journal of Fracture}, 101\penalty0 (1):\penalty0
  1--31, January 2000.
\newblock ISSN 1573-2673.
\newblock \doi{10.1023/A:1007647800529}.

\bibitem[Kendall et~al.(1946)]{kendall1946advanced}
Maurice~George Kendall et~al.
\newblock The advanced theory of statistics.
\newblock \emph{The advanced theory of statistics.}, \penalty0 (2nd Ed), 1946.

\bibitem[Kipf and Welling(2017)]{kipf_semi-supervised_2017}
Thomas~N. Kipf and Max Welling.
\newblock Semi-{Supervised} {Classification} with {Graph} {Convolutional}
  {Networks}.
\newblock \emph{arXiv:1609.02907 [cs, stat]}, February 2017.

\bibitem[Kirchdoerfer and Ortiz(2016)]{kirchdoerfer2016data}
Trenton Kirchdoerfer and Michael Ortiz.
\newblock Data-driven computational mechanics.
\newblock \emph{Computer Methods in Applied Mechanics and Engineering},
  304:\penalty0 81--101, 2016.

\bibitem[Kochmann et~al.(2018)Kochmann, Ehle, Wulfinghoff, Mayer, Svendsen, and
  Reese]{kochmann2018efficient}
Julian Kochmann, Lisa Ehle, Stephan Wulfinghoff, Joachim Mayer, Bob Svendsen,
  and Stefanie Reese.
\newblock Efficient multiscale fe-fft-based modeling and simulation of
  macroscopic deformation processes with non-linear heterogeneous
  microstructures.
\newblock In \emph{Multiscale Modeling of Heterogeneous Structures}, pages
  129--146. Springer, 2018.

\bibitem[Krizhevsky et~al.(2012)Krizhevsky, Sutskever, and
  Hinton]{krizhevsky_imagenet_2012}
Alex Krizhevsky, Ilya Sutskever, and Geoffrey~E Hinton.
\newblock {ImageNet} {Classification} with {Deep} {Convolutional} {Neural}
  {Networks}.
\newblock In F.~Pereira, C.~J.~C. Burges, L.~Bottou, and K.~Q. Weinberger,
  editors, \emph{Advances in {Neural} {Information} {Processing} {Systems} 25},
  pages 1097--1105. Curran Associates, Inc., 2012.

\bibitem[Kuhn et~al.(2015)Kuhn, Sun, and Wang]{kuhn_stress-induced_2015}
Matthew~R Kuhn, WaiChing Sun, and Qi~Wang.
\newblock Stress-induced anisotropy in granular materials: fabric, stiffness,
  and permeability.
\newblock \emph{Acta Geotechnica}, 10\penalty0 (4):\penalty0 399--419, 2015.

\bibitem[Le et~al.(2015)Le, Yvonnet, and He]{le2015computational}
BA~Le, Julien Yvonnet, and Q-C He.
\newblock Computational homogenization of nonlinear elastic materials using
  neural networks.
\newblock \emph{International Journal for Numerical Methods in Engineering},
  104\penalty0 (12):\penalty0 1061--1084, 2015.

\bibitem[Lecun et~al.(1998)Lecun, Bottou, Bengio, and
  Haffner]{lecun_gradient-based_1998}
Y.~Lecun, L.~Bottou, Y.~Bengio, and P.~Haffner.
\newblock Gradient-based learning applied to document recognition.
\newblock \emph{Proceedings of the IEEE}, 86\penalty0 (11):\penalty0
  2278--2324, November 1998.
\newblock ISSN 0018-9219, 1558-2256.
\newblock \doi{10.1109/5.726791}.

\bibitem[Lefik et~al.(2009)Lefik, Boso, and Schrefler]{lefik_artificial_2009}
M~Lefik, DP~Boso, and BA~Schrefler.
\newblock Artificial neural networks in numerical modelling of composites.
\newblock \emph{Computer Methods in Applied Mechanics and Engineering},
  198\penalty0 (21-26):\penalty0 1785--1804, 2009.

\bibitem[Liu et~al.(2016)Liu, Sun, and Fish]{liu_determining_2016}
Yang Liu, WaiChing Sun, and Jacob Fish.
\newblock Determining material parameters for critical state plasticity models
  based on multilevel extended digital database.
\newblock \emph{Journal of Applied Mechanics}, 83\penalty0 (1):\penalty0
  011003, 2016.

\bibitem[Liu et~al.(2018)Liu, Kafka, Yu, and Liu]{liu2018data}
Zeliang Liu, Orion~L Kafka, Cheng Yu, and Wing~Kam Liu.
\newblock Data-driven self-consistent clustering analysis of heterogeneous
  materials with crystal plasticity.
\newblock In \emph{Advances in computational plasticity}, pages 221--242.
  Springer, 2018.

\bibitem[Liu et~al.(2019)Liu, Wu, and Koishi]{liu_deep_2019}
Zeliang Liu, CT~Wu, and M~Koishi.
\newblock A deep material network for multiscale topology learning and
  accelerated nonlinear modeling of heterogeneous materials.
\newblock \emph{Computer Methods in Applied Mechanics and Engineering},
  345:\penalty0 1138--1168, 2019.

\bibitem[Lu et~al.(2019)Lu, Giovanis, Yvonnet, Papadopoulos, Detrez, and
  Bai]{lu_data-driven_2019}
Xiaoxin Lu, Dimitris~G Giovanis, Julien Yvonnet, Vissarion Papadopoulos,
  Fabrice Detrez, and Jinbo Bai.
\newblock A data-driven computational homogenization method based on neural
  networks for the nonlinear anisotropic electrical response of
  graphene/polymer nanocomposites.
\newblock \emph{Computational Mechanics}, 64\penalty0 (2):\penalty0 307--321,
  2019.

\bibitem[Lubbers et~al.(2017)Lubbers, Lookman, and
  Barros]{lubbers_inferring_2017}
Nicholas Lubbers, Turab Lookman, and Kipton Barros.
\newblock Inferring low-dimensional microstructure representations using
  convolutional neural networks.
\newblock \emph{Physical Review E}, 96\penalty0 (5):\penalty0 052111, November
  2017.
\newblock ISSN 2470-0045, 2470-0053.
\newblock \doi{10.1103/PhysRevE.96.052111}.

\bibitem[Ma and Sun(2019)]{ma_fft_2019}
Ran Ma and WaiChing Sun.
\newblock Fft-based solver for higher-order and multi-phase-field fracture
  models applied to strongly anisotropic brittle materials and poly-crystals.
\newblock \emph{Computer Methods in Applied Mechanics and Engineering}, 2019.
\newblock tentatively accepted.

\bibitem[Ma et~al.(2018)Ma, Truster, Puplampu, and
  Penumadu]{ma_investigating_2018}
Ran Ma, Timothy~J Truster, Stephen~B Puplampu, and Dayakar Penumadu.
\newblock Investigating mechanical degradation due to fire exposure of aluminum
  alloy 5083 using crystal plasticity finite element method.
\newblock \emph{International Journal of Solids and Structures}, 134:\penalty0
  151--160, 2018.

\bibitem[Manzari and Dafalias(1997)]{manzari_critical_1997}
Majid~T Manzari and Yannis~F Dafalias.
\newblock A critical state two-surface plasticity model for sands.
\newblock \emph{Geotechnique}, 47\penalty0 (2):\penalty0 255--272, 1997.

\bibitem[Miehe et~al.(2010{\natexlab{a}})Miehe, Hofacker, and
  Welschinger]{miehe_phase_2010-1}
C.~Miehe, M.~Hofacker, and F.~Welschinger.
\newblock A phase field model for rate-independent crack propagation: {Robust}
  algorithmic implementation based on operator splits.
\newblock \emph{Computer Methods in Applied Mechanics and Engineering},
  199\penalty0 (45-48):\penalty0 2765--2778, 2010{\natexlab{a}}.
\newblock ISSN 00457825.
\newblock \doi{10.1016/j.cma.2010.04.011}.

\bibitem[Miehe et~al.(2010{\natexlab{b}})Miehe, Hofacker, and
  Welschinger]{miehe_phase_2010}
Christian Miehe, Martina Hofacker, and Fabian Welschinger.
\newblock A phase field model for rate-independent crack propagation: {Robust}
  algorithmic implementation based on operator splits.
\newblock \emph{Computer Methods in Applied Mechanics and Engineering},
  199\penalty0 (45):\penalty0 2765--2778, 2010{\natexlab{b}}.

\bibitem[Mikolov et~al.(2013)Mikolov, Sutskever, Chen, Corrado, and
  Dean]{mikolov_distributed_2013}
Tomas Mikolov, Ilya Sutskever, Kai Chen, Greg~S Corrado, and Jeff Dean.
\newblock Distributed {Representations} of {Words} and {Phrases} and their
  {Compositionality}.
\newblock In C.~J.~C. Burges, L.~Bottou, M.~Welling, Z.~Ghahramani, and K.~Q.
  Weinberger, editors, \emph{Advances in {Neural} {Information} {Processing}
  {Systems} 26}, pages 3111--3119. Curran Associates, Inc., 2013.

\bibitem[Na and Sun(2018)]{na_computational_2018}
SeonHong Na and WaiChing Sun.
\newblock Computational thermomechanics of crystalline rock, {Part} {I}: {A}
  combined multi-phase-field/crystal plasticity approach for single crystal
  simulations.
\newblock \emph{Computer Methods in Applied Mechanics and Engineering},
  338:\penalty0 657--691, 2018.

\bibitem[Nahshon and Hutchinson(2008)]{nahshon_modification_2008}
Ken Nahshon and JW~Hutchinson.
\newblock Modification of the {Gurson} model for shear failure.
\newblock \emph{European Journal of Mechanics-A/Solids}, 27\penalty0
  (1):\penalty0 1--17, 2008.

\bibitem[Narayanan et~al.(2017)Narayanan, Chandramohan, Venkatesan, Chen, Liu,
  and Jaiswal]{narayanan_graph2vec:_2017}
Annamalai Narayanan, Mahinthan Chandramohan, Rajasekar Venkatesan, Lihui Chen,
  Yang Liu, and Shantanu Jaiswal.
\newblock graph2vec: {Learning} {Distributed} {Representations} of {Graphs}.
\newblock \emph{arXiv:1707.05005 [cs]}, July 2017.

\bibitem[Needleman(1987)]{needleman_continuum_1987}
Alan Needleman.
\newblock A continuum model for void nucleation by inclusion debonding.
\newblock \emph{Journal of applied mechanics}, 54\penalty0 (3):\penalty0
  525--531, 1987.

\bibitem[Nielsen and Tvergaard(2010)]{nielsen_ductile_2010}
Kim~Lau Nielsen and Viggo Tvergaard.
\newblock Ductile shear failure or plug failure of spot welds modelled by
  modified {Gurson} model.
\newblock \emph{Engineering Fracture Mechanics}, 77\penalty0 (7):\penalty0
  1031--1047, 2010.

\bibitem[Perozzi et~al.(2014)Perozzi, Al-Rfou, and
  Skiena]{perozzi_deepwalk:_2014}
Bryan Perozzi, Rami Al-Rfou, and Steven Skiena.
\newblock {DeepWalk}: {Online} {Learning} of {Social} {Representations}.
\newblock \emph{Proceedings of the 20th ACM SIGKDD international conference on
  Knowledge discovery and data mining - KDD '14}, pages 701--710, 2014.
\newblock \doi{10.1145/2623330.2623732}.

\bibitem[Quey et~al.(2011)Quey, Dawson, and Barbe]{quey_large-scale_2011}
R.~Quey, P.~R. Dawson, and F.~Barbe.
\newblock Large-scale 3d random polycrystals for the finite element method:
  {Generation}, meshing and remeshing.
\newblock \emph{Computer Methods in Applied Mechanics and Engineering},
  200\penalty0 (17):\penalty0 1729--1745, April 2011.
\newblock ISSN 0045-7825.
\newblock \doi{10.1016/j.cma.2011.01.002}.

\bibitem[Salinger et~al.(2016)Salinger, Bartlett, Bradley, Chen, Demeshko, Gao,
  Hansen, Mota, Muller, Nielsen, et~al.]{salinger2016albany}
Andrew~G Salinger, Roscoe~A Bartlett, Andrew~M Bradley, Qiushi Chen, Irina~P
  Demeshko, Xujiao Gao, Glen~A Hansen, Alejandro Mota, Richard~P Muller, Erik
  Nielsen, et~al.
\newblock Albany: using component-based design to develop a flexible, generic
  multiphysics analysis code.
\newblock \emph{International Journal for Multiscale Computational
  Engineering}, 14\penalty0 (4), 2016.

\bibitem[Satake(1992)]{satake1992discrete}
Masao Satake.
\newblock A discrete-mechanical approach to granular materials.
\newblock \emph{International journal of engineering science}, 30\penalty0
  (10):\penalty0 1525--1533, 1992.

\bibitem[Scarselli et~al.(2008)Scarselli, Gori, Tsoi, Hagenbuchner, and
  Monfardini]{scarselli2008graph}
Franco Scarselli, Marco Gori, Ah~Chung Tsoi, Markus Hagenbuchner, and Gabriele
  Monfardini.
\newblock The graph neural network model.
\newblock \emph{IEEE Transactions on Neural Networks}, 20\penalty0
  (1):\penalty0 61--80, 2008.

\bibitem[Schofield and Wroth(1968)]{schofield_critical_1968}
Andrew Schofield and Peter Wroth.
\newblock \emph{Critical state soil mechanics}, volume 310.
\newblock McGraw-Hill London, 1968.

\bibitem[Simo and Ju(1987)]{simo1987strain}
Juan~C Simo and JW~Ju.
\newblock Strain-and stress-based continuum damage models---i. formulation.
\newblock \emph{International journal of solids and structures}, 23\penalty0
  (7):\penalty0 821--840, 1987.

\bibitem[Simonovsky and Komodakis(2017)]{simonovsky2017dynamic}
Martin Simonovsky and Nikos Komodakis.
\newblock Dynamic edge-conditioned filters in convolutional neural networks on
  graphs, 2017.

\bibitem[Song et~al.(2008)Song, Wang, and Belytschko]{song_comparative_2008}
Jeong-Hoon Song, Hongwu Wang, and Ted Belytschko.
\newblock A comparative study on finite element methods for dynamic fracture.
\newblock \emph{Computational Mechanics}, 42\penalty0 (2):\penalty0 239--250,
  July 2008.
\newblock ISSN 1432-0924.
\newblock \doi{10.1007/s00466-007-0210-x}.

\bibitem[Sonoda and Murata(2017)]{sonoda2017neural}
Sho Sonoda and Noboru Murata.
\newblock Neural network with unbounded activation functions is universal
  approximator.
\newblock \emph{Applied and Computational Harmonic Analysis}, 43\penalty0
  (2):\penalty0 233--268, 2017.

\bibitem[Srivastava et~al.(2014)Srivastava, Hinton, Krizhevsky, Sutskever, and
  Salakhutdinov]{srivastava_dropout:_2014}
Nitish Srivastava, Geoffrey Hinton, Alex Krizhevsky, Ilya Sutskever, and Ruslan
  Salakhutdinov.
\newblock Dropout: {A} {Simple} {Way} to {Prevent} {Neural} {Networks} from
  {Overfitting}.
\newblock \emph{Journal of Machine Learning Research}, 15:\penalty0 1929--1958,
  2014.

\bibitem[Stoffel et~al.(2019)Stoffel, Bamer, and Markert]{stoffel2019stability}
M~Stoffel, F~Bamer, and B~Markert.
\newblock Stability of feed forward artificial neural networks versus nonlinear
  structural models in high speed deformations: A critical comparison.
\newblock \emph{Archives of Mechanics}, 71\penalty0 (2), 2019.

\bibitem[Sun(2013)]{sun_unified_2013}
WaiChing Sun.
\newblock A unified method to predict diffuse and localized instabilities in
  sands.
\newblock \emph{Geomechanics and Geoengineering}, 8\penalty0 (2):\penalty0
  65--75, 2013.

\bibitem[Sun and Mota(2014)]{sun_multiscale_2014}
Waiching Sun and Alejandro Mota.
\newblock A multiscale overlapped coupling formulation for large-deformation
  strain localization.
\newblock pages 803--820, 2014.
\newblock \doi{10.1007/s00466-014-1034-0}.

\bibitem[Sun et~al.(2013)Sun, Kuhn, and Rudnicki]{sun2013multiscale}
WaiChing Sun, Matthew~R Kuhn, and John~W Rudnicki.
\newblock A multiscale dem-lbm analysis on permeability evolutions inside a
  dilatant shear band.
\newblock \emph{Acta Geotechnica}, 8\penalty0 (5):\penalty0 465--480, 2013.

\bibitem[Tamura and Gallagher(2019)]{tamura2019quantitative}
Kenichi Tamura and Marcus Gallagher.
\newblock Quantitative measure of nonconvexity for black-box continuous
  functions.
\newblock \emph{Information Sciences}, 476:\penalty0 64--82, 2019.

\bibitem[Teichert et~al.(2019)Teichert, Natarajan, Van~der Ven, and
  Garikipati]{teichert2019machine_a}
GH~Teichert, AR~Natarajan, A~Van~der Ven, and K~Garikipati.
\newblock Machine learning materials physics: Integrable deep neural networks
  enable scale bridging by learning free energy functions.
\newblock \emph{Computer Methods in Applied Mechanics and Engineering},
  353:\penalty0 201--216, 2019.

\bibitem[Teichert and Garikipati(2019)]{teichert2019machine_b}
Gregory~H Teichert and Krishna Garikipati.
\newblock Machine learning materials physics: Surrogate optimization and
  multi-fidelity algorithms predict precipitate morphology in an alternative to
  phase field dynamics.
\newblock \emph{Computer Methods in Applied Mechanics and Engineering},
  344:\penalty0 666--693, 2019.

\bibitem[Tordesillas et~al.(2014)Tordesillas, Pucilowski, Walker, Peters, and
  Walizer]{tordesillas2014micromechanics}
Antoinette Tordesillas, Sebastian Pucilowski, David~M Walker, John~F Peters,
  and Laura~E Walizer.
\newblock Micromechanics of vortices in granular media: connection to shear
  bands and implications for continuum modelling of failure in geomaterials.
\newblock \emph{International Journal for Numerical and Analytical Methods in
  Geomechanics}, 38\penalty0 (12):\penalty0 1247--1275, 2014.

\bibitem[Vincent et~al.(2008)Vincent, Larochelle, Bengio, and
  Manzagol]{vincent_extracting_2008}
Pascal Vincent, Hugo Larochelle, Yoshua Bengio, and Pierre-Antoine Manzagol.
\newblock Extracting and {Composing} {Robust} {Features} with {Denoising}
  {Autoencoders}.
\newblock In \emph{Proceedings of the 25th {International} {Conference} on
  {Machine} {Learning}}, {ICML} '08, pages 1096--1103, New York, NY, USA, 2008.
  ACM.
\newblock ISBN 978-1-60558-205-4.
\newblock \doi{10.1145/1390156.1390294}.

\bibitem[Wang and Sun(2018)]{wang_multiscale_2018}
Kun Wang and WaiChing Sun.
\newblock A multiscale multi-permeability poroplasticity model linked by
  recursive homogenizations and deep learning.
\newblock \emph{Computer Methods in Applied Mechanics and Engineering},
  334:\penalty0 337--380, 2018.

\bibitem[Wang and Sun(2019{\natexlab{a}})]{wang2019meta}
Kun Wang and WaiChing Sun.
\newblock Meta-modeling game for deriving theory-consistent,
  microstructure-based traction--separation laws via deep reinforcement
  learning.
\newblock \emph{Computer Methods in Applied Mechanics and Engineering},
  346:\penalty0 216--241, 2019{\natexlab{a}}.

\bibitem[Wang and Sun(2019{\natexlab{b}})]{wang2019updated}
Kun Wang and WaiChing Sun.
\newblock An updated lagrangian lbm--dem--fem coupling model for
  dual-permeability fissured porous media with embedded discontinuities.
\newblock \emph{Computer Methods in Applied Mechanics and Engineering},
  344:\penalty0 276--305, 2019{\natexlab{b}}.

\bibitem[Wang et~al.(2016{\natexlab{a}})Wang, Sun, Salager, Na, and
  Khaddour]{wang2016identifying}
Kun Wang, Waiching Sun, Simon Salager, SeonHong Na, and Ghonwa Khaddour.
\newblock Identifying material parameters for a micro-polar plasticity model
  via x-ray micro-ct images: lessons learned from the curve-fitting exercises.
\newblock \emph{International Journal for Multiscale Computational
  Engineering}, 2016{\natexlab{a}}.

\bibitem[Wang et~al.(2016{\natexlab{b}})Wang, Sun, Salager, Na, and
  Khaddour]{wang_identifying_2016}
Kun Wang, WaiChing Sun, Simon Salager, SeonHong Na, and Ghonwa Khaddour.
\newblock Identifying material parameters for a micro-polar plasticity model
  via {X}-ray micro-computed tomographic ({CT}) images: lessons learned from
  the curve-fitting exercises.
\newblock \emph{International Journal for Multiscale Computational
  Engineering}, 14\penalty0 (4), 2016{\natexlab{b}}.

\bibitem[West et~al.(2001)]{west2001introduction}
Douglas~Brent West et~al.
\newblock \emph{Introduction to graph theory}, volume~2.
\newblock Prentice hall Upper Saddle River, 2001.

\bibitem[Wu et~al.(2019)Wu, Pan, Chen, Long, Zhang, and
  Yu]{wu_comprehensive_2019}
Zonghan Wu, Shirui Pan, Fengwen Chen, Guodong Long, Chengqi Zhang, and
  Philip~S. Yu.
\newblock A {Comprehensive} {Survey} on {Graph} {Neural} {Networks}.
\newblock \emph{arXiv:1901.00596 [cs, stat]}, August 2019.
\newblock arXiv: 1901.00596.

\bibitem[Zhang et~al.(2000)Zhang, Thaulow, and {\O}degard]{zhang_complete_2000}
ZL~Zhang, C~Thaulow, and J~{\O}degard.
\newblock A complete {Gurson} model approach for ductile fracture.
\newblock \emph{Engineering Fracture Mechanics}, 67\penalty0 (2):\penalty0
  155--168, 2000.

\end{thebibliography}

\end{document}